\documentclass[10pt,journal,compsoc]{IEEEtran}
\usepackage{cite}
\usepackage{algorithm}
\usepackage{algorithmicx} 
\usepackage[noend]{algcompatible}
\usepackage{amsmath,bm}
\usepackage[pdftex]{graphicx}
\usepackage{multirow}
\usepackage{array}
\usepackage[justification=centering]{caption}
\usepackage{epstopdf}
\usepackage{comment}
\usepackage[table,xcdraw]{xcolor}
\usepackage{booktabs} 
\usepackage{subfigure}
\usepackage{graphicx}
\usepackage{pifont}
\newcommand{\cmark}{\ding{51}}%
\newcommand{\xmark}{\ding{55}}%
\newcommand\norm[1]{\left\lVert#1\right\rVert}
\makeatletter
\long\def\@IEEEtitleabstractindextextbox#1{\parbox{0.922\textwidth}{#1}}
\makeatother
\graphicspath{{./Figures/}}
\begin{document}
\title{RL-DistPrivacy: Privacy-Aware Distributed Deep Inference for low latency IoT systems}
\author{
    \IEEEauthorblockN{Emna~Baccour\IEEEauthorrefmark{1}, Aiman~Erbad\IEEEauthorrefmark{1}, Amr~Mohamed\IEEEauthorrefmark{2}, Mounir~Hamdi\IEEEauthorrefmark{1}, and
    Mohsen~Guizani\IEEEauthorrefmark{3}.}\\
    \IEEEauthorblockA{\IEEEauthorrefmark{1}College of Science and Engineering, Hamad Bin Khalifa University, Doha, Qatar.}\\
    \IEEEauthorblockA{\IEEEauthorrefmark{2}CS department, College of Engineering, Qatar University.}\\
   \IEEEauthorblockA{\IEEEauthorrefmark{3} Mohamed Bin Zayed University of Artificial Intelligence (MBZUAI), Abu Dhabi, UAE}
}
\IEEEtitleabstractindextext{%
\begin{abstract}
Although Deep Neural Networks (DNN) have become the backbone technology of several ubiquitous applications, their deployment in resource-constrained machines, e.g., Internet of Things (IoT) devices, is still challenging. To satisfy the resource requirements of such a paradigm, collaborative deep inference with IoT synergy was introduced. However, the distribution of DNN networks suffers from severe data leakage. Various threats have been presented, including black-box attacks, where malicious participants can recover arbitrary inputs fed into their devices. Although many countermeasures were designed to achieve privacy-preserving DNN, most of them result in additional computation and lower accuracy. In this paper, we present an approach that targets the security of collaborative deep inference via re-thinking the distribution strategy, without sacrificing the model performance. Particularly, we examine different DNN partitions that make the model susceptible to black-box threats and we derive the amount of data that should be allocated per device to hide proprieties of the original input. We formulate this methodology, as an optimization, where we establish a trade-off between the latency of co-inference and the privacy-level of data. Next, to relax the optimal solution, we shape our approach as a Reinforcement Learning (RL) design that supports heterogeneous devices as well as multiple DNNs/datasets.
\end{abstract}
}
\maketitle
\IEEEdisplaynontitleabstractindextext

\IEEEpeerreviewmaketitle
\IEEEraisesectionheading{\section{Introduction}\label{sec:introduction}}
\IEEEPARstart{D}{riven} by the recent advancement of big data and Artificial Intelligence (AI), deep neural networks have presented substantial breakthroughs in a broad range of applications, spanning from speech recognition \cite{speech} and smart home appliances to image recognition \cite{image}, video surveillance and robotics \cite{robots}. Such high performance of deep networks is related to the ability of the model to obtain distinctive high-level features from raw data, owing to its extensive mathematical operations, its complex structure, the large number of parameters and hidden layers, and the significant amount of processed data. A representative example of a state-of-the-art Deep Neural Network that has demonstrated unprecedented performance in vision recognition tasks, is VGG. VGG \cite{VGG} includes more than 144 million parameters, 15 million neurons and  3 billion connections and can only be executed on powerful devices as deploying it on mobile devices presents intolerable classification latency, reaching over 16 seconds \cite{MoDNN}.

As DNN-based applications use a tremendous amount of data, they require accordingly large memory and computation capacities. However, pervasive data-generating devices fail to deploy large-scale computation with reasonable energy consumption and latency. The current progress to adapt DNNs to mobile devices presents only unsatisfactory solutions: either to execute the classification on resource-constrained units using compressed models \cite{squeeze} resulting in low accuracy, or to offload the data to be identified by larger models hosted in powerful servers.
The traditional wisdom resorts to remote
cloud servers to execute complex inference tasks \cite{remote1}. 
However, the remote inference limits the  ubiquitous deployment of deep learning. Indeed, such a server-centric approach cannot ensure the real-time requirements of services that need prompt intervention, such as car plates or accidents recognition. 
To cater for the latency and bandwidth issues, DNN computation should be moved at the proximity of data-generating devices to make the system autonomous and reduce the required communication bandwidth.

\begin{figure*}[!h]
\centering
	\includegraphics[scale=0.63]{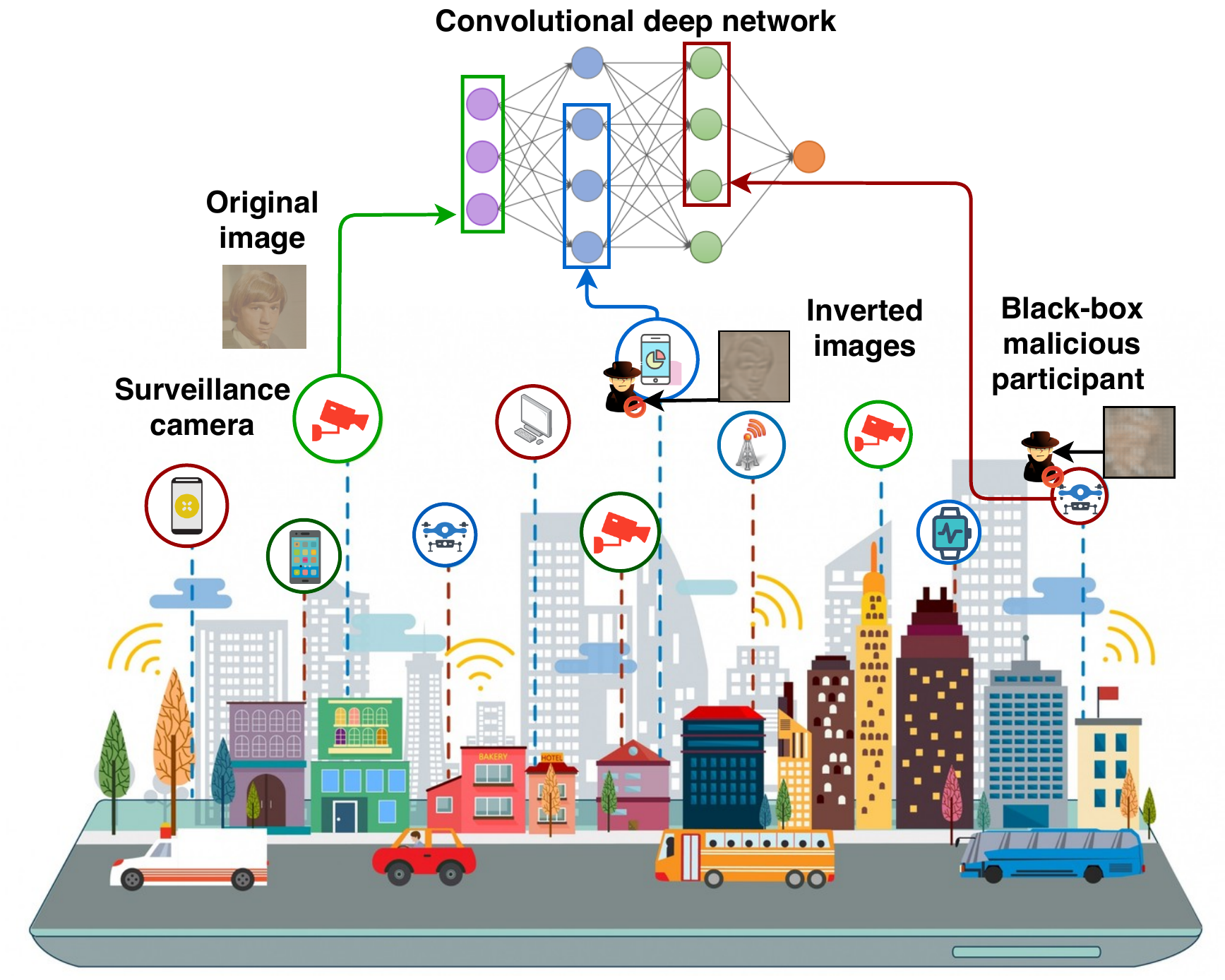}
	\caption{Illustration of the privacy-aware distributed CNN with black-box malicious attackers.}
	\label{journal_model}
	\vspace{- 3 mm}
\end{figure*}
To enable the revolutionary
shift of data computation from remote cloud servers to pervasive end-devices (e.g., mobile and IoT devices),
 collaborative deep learning strategies have been proposed in the literature. Indeed, two collaborative modes have been designed: The first mode, namely federated learning, consists of distributing the training tasks over multiple helpers while keeping the data locally in devices, which improves the efficiency and the speed of the learning \cite{federated2}. In the second mode, studied in this paper, the inference phase is distributed. Particularly, the trained DNN model is partitioned into segments, which are assigned to different participants. Each participant computes a segment and shares the output to the subsequent device until generating the final prediction. The distributed inference has attracted the focus of academia, particularly the mobile-cloud collaboration scenarios \cite{neurosurgeon,Boomerang}. A step forward, the researchers studied the feasibility to use only the resource-constrained IoT devices to execute the entire inference at the proximity of the data source by jointly compute different segments of the neural network \cite{MoDNN,DeepThings}. These efforts examined different DNN splitting scenarios and mainly focused on finding the optimal partitioning that reduces the dependency and data sharing between participants. However, designing a joint strategy where participants are scheduled to compute different segments of online inference requests while being constrained by their limited resources was not considered by previous works.  Therefore, in order to support deep neural processing in IoT systems existing at the proximity of the data-generating device, the design of DNN distribution must be completely rethought by considering hardware and physical constraints, which will be done in our work.

However, this multi-party processing introduces new vulnerabilities and attacks because of the untrusted environment. More specifically, when the trained model is split and distributed among different helpers, an untrusted device can recover the input fed by the previous helper, even if it does not have complete information about the model structure. In this context, 
authors in \cite{security} introduced different possible attacks against collaborative inference, including black-box attacks. In black-box settings, the malicious participant only has knowledge about his segment and attempts to design an inverse DNN network to map the received features to the targeted input and recover the original data. Authors showed that reversing the original data is possible, when the neural system is split at shallow layers as proposed in \cite{dis5}. Hence, an intuitive technique is to split the DNN after the fully connected layers, owing to their ability to mix the input and hide the sample features. Yet, this 
may not be adequate to all IoT nodes, where devices have limited resources (e.g., sensors) and cannot execute a high number of complex layers. Many other countermeasures have been considered to strengthen the robustness of deep networks, including the differential privacy, which suggests adding noise to the intermediate outputs to blur or obscure it. However, performing classification on corrupted data  may decrease the accuracy of the system. A similar technique  to secure the distributed environment is to encrypt the data before inference in order to guarantee the non-leakage of the private information. An obvious drawback of such approach is that it suffers from learning inefficiency, requires additional computation, and cannot be applicable to all DNN networks nor supported by all IoT devices.

In our work, we start by testing black-box threats on different partition scenarios to derive facts that make the model susceptible to black-box attacks. We demonstrate that splitting the neural network into layers is highly fragile against inversion threats. Meanwhile, distributing different feature maps resulting from each layer into multiple IoT devices guarantees the non-exposure of all important features and properties to a single device and hence preventing to reconstruct the initial input, even in shallow layers,  as illustrated in Figure \ref{journal_model}.

In this paper, we study the deployment of Convolutional Neural Networks (CNNs) within a surveillance IoT system. We chose CNNs as they are widely used for image classification tasks. We build a collaboration strategy that conducts online inference tasks at the proximity of  data-generating devices through distributing CNN feature maps into multiple heterogeneous IoT devices, aiming to minimize the latency while taking into consideration the privacy of the data that become highly vulnerable to attacks. To the best of our knowledge,  we  believe that  we  are  the  first  to  exploit the CNN distribution to enhance the privacy of the system, while being adequate to different types of devices or DNN tasks and without reducing the accuracy of results or increasing the computational load. The contributions of our paper include the following:
\begin{itemize}
    \item  We conduct  deep  empirical  experiments  to  test  the  efficiency of  black-box  attacks  on  different  partition  scenarios.  In  our simulation,  we  tested  four  different  datasets  and  four  state-of-the-art CNN networks.
    \item We formulate our privacy-aware collaborative inference as an optimization problem aiming at minimizing the latency of the classification, while taking into consideration the required privacy and the limited  resources  of participating IoT devices.
    \item To relax the optimization problem, we design a Reinforcement Learning approach for Distributed and Privacy-aware CNN networks, namely RL-DistPrivacy. This RL technique learns the temporal inference tasks to come up with the best policy for allocation and offloading of computational CNN segments by taking real-time actions. On this basis, the sets of states and actions are defined along with the reward function. Furthermore, the RL approach is solved using an efficient algorithm, namely Deep Q-Learning (DQN).
    \item To evaluate the performance of our novel RL-DistPrivacy system, we conduct extensive simulations on different CNN types, datasets, and network configurations. We show that our predictive approach can reduce the inference latency  and enhance the system privacy compared  to  recent  approaches.  Moreover, we show that the proposed RL-approach achieves a performance exceeding 70 \% compared to the optimal results, which justifies its capacity to handle CNN allocation tasks.
\end{itemize}
Our paper is organized as follows: In section \ref{related_works}, we explore the related literature. In section \ref{deep_privacy}, we present our novel framework, the empirical study, and the problem formulation, and we introduce our RL-DistPrivacy approach. In section \ref{simulation}, the evaluation of the system is conducted, using different state-of-the-art CNN networks. Then, in section \ref{conclusion}, we draw the conclusions.
\section{Related works}\label{related_works}
Deep neural networks have presented substantial breakthroughs in a wide range of applications including robotics, speech recognition and computer vision. 
A typical DNN network may be built of multiple layers and neurons and these components may scale from tens to reach millions. 
Such high resource-demanding DNNs can only be computed on powerful machines (e.g., cloud.) having large memory capacity, which is incompatible with low cost and limited energy IoT devices.
However, the remote inference limits the  ubiquitous deployment of deep learning as it cannot ensure the real-time requirements of services that need instant intervention.
\begin{table*}[!h]
\centering
\caption{Comparison between privacy-aware  strategies.}
\label{tab:privacy}
\begin{tabular}{|c|c|c|c|c|c|c|c|}
\hline
\begin{tabular}[c]{@{}c@{}}\textbf{Privacy-aware}\\ \textbf{strategy}\end{tabular} & \begin{tabular}[c]{@{}c@{}}\textbf{Privacy}\\ \textbf{level}\end{tabular} & \begin{tabular}[c]{@{}c@{}}\textbf{Accuracy}\\ \textbf{preserving}\end{tabular} & \begin{tabular}[c]{@{}c@{}}\textbf{DNN}\\ \textbf{re-training}\end{tabular} & \begin{tabular}[c]{@{}c@{}}\textbf{Compatibility with}\\ \textbf{all IoT and DNNs}\end{tabular} & \begin{tabular}[c]{@{}c@{}}\textbf{Partitioning}\\ \textbf{strategy}\end{tabular} & \begin{tabular}[c]{@{}c@{}}\textbf{Communication}\\ \textbf{overhead}\end{tabular} & \begin{tabular}[c]{@{}c@{}}\textbf{Computation}\\ \textbf{overhead on}\\ \textbf{source-device}\end{tabular} \\ \hline
\begin{tabular}[c]{@{}c@{}}Deep split \end{tabular} & +++ & \cmark & \xmark & \cmark & per-layer & + & +++ \\ \hline
\begin{tabular}[c]{@{}c@{}}Feature extraction \end{tabular} & + & \xmark & \xmark & \cmark & per-layer & + & ++ \\ \hline
\begin{tabular}[c]{@{}c@{}}Noise addition \end{tabular} & ++ & \xmark & \cmark & \xmark & per-layer & ++ & +++ \\ \hline
\begin{tabular}[c]{@{}c@{}}Cryptography \end{tabular} & +++ & \xmark & \cmark & \xmark & per-layer & ++ & +++ \\ \hline
\begin{tabular}[c]{@{}c@{}}TEE \end{tabular} & +++ & \cmark & \xmark & \xmark & on-device & None & +++ \\ \hline
\begin{tabular}[c]{@{}c@{}}RL-DistPrivacy \end{tabular} & ++ & \cmark & \xmark & \cmark & \begin{tabular}[c]{@{}c@{}}filter\\ splitting \end{tabular} & +++ & + \\ \hline
\end{tabular}
\end{table*}

The nexus between IoT devices motivates the research community to push the DNN computation in close proximity of data source, in order to reduce the inference latency, minimize the network cost, and avoid bandwidth bottlenecks  \cite{survey2,survey21}.
Particularly, the trained DNN network is split into segments that are assigned to different helpers. 
The DNN partitioning and distribution approaches can be categorized into three classes:  The main idea of the first approach is to split the model into two segments, where the shallow part is computed in the data-generating device and the deep layers are offloaded to remote servers, i.e., edge or cloud servers. Therefore, the objective will be to decide the split point depending on the size of intermediate data  that is relatively small for deep layers and the latency to transmit it, as done in Neurosurgeon \cite{neurosurgeon}. 
The hierarchical partition is the second strategy proposed in the literature, where the DNN network is split and distributed across cloud, edge servers, and  end-devices. The work in \cite{DDNN} adopted this approach combined with exiting the model at early points so that the computation does not always reach cloud servers.

The above-mentioned approaches cover mainly the offloading of part of DNN tasks from data-generating devices to high-performance servers, in order to speed up the inference while minimizing the traffic load. Still, in case of high classification load, network congestion and bottleneck can occur. To cater with remote transmission, researchers have started to investigate the feasibility of distributing CNN networks among resource-limited IoT devices in order to jointly compute the inferences locally.
The works in \cite{dis5,dis6,dis7} presented per-layer CNN partitioning strategy and formulated the allocation approach to distribute different segments into resource constrained devices, as an optimization problem. In addition to per layer partitioning, the DNN can also be divided along the input dimension (e.g., rows or columns of feature maps). Such input wise splitting enables more flexible and fine grained allocation as the data size  of each segment can be chosen arbitrarily to be adequate to lightweight devices (e.g., sensors), compared to the fixed partition sizes defined by the layers of the DNN. 
However, a major shortcoming of this fine-grained partition is the high dependency between participants. Among the contributions that led the research to examine the feasibility of input-wise partitioning, we can cite DeepThings \cite{DeepThings} and MoDNN \cite{MoDNN}. These efforts studied different DNN splitting scenarios and mainly focused on finding the optimal partitioning that reduces the dependency and data sharing between participants. Still, designing a joint strategy where participants are scheduled to compute different segments of online inference requests while being constrained by their limited resources was not considered by previous works.

The data generated by source devices and distributed to multiple nodes may contain sensitive information that should not be shared with untrusted parties (e.g., images, GPS coordinates, and audio.). 
However, the IoT system is composed of uncontrolled devices that may attack the input fed by the previous helpers and expose the system to privacy issues. For example, authors in \cite{security} presented some potential attacks and vulnerabilities menacing the collaborative inference, including black-box attacks. In black-box settings, the attacker trains a DNN network that maps the received features to the targeted input in order to obtain the original sensitive data. Attacking the inference through an inverse network is feasible and more efficient, when the neural system is distributed into layers as done in \cite{dis5}.

Many countermeasures have been designed to enhance the privacy of distributed deep networks. One of these measures is the deep split \cite{security}, where the model is split and distributed after fully-connected (fc) layers. Particularly, owing to their ability to mix the input and hide the sample features, the outputs of fully-connected layers are difficult to be recovered and the initial data cannot be generated. Thus, the idea is to compute the first segment in the data-generating device, while delegating the second part to another participant. However, following this approach, all convolutional layers should be hosted in the source device, which is not always possible and compatible with all IoT devices (e.g., sensors.) due to the complexity and resource requirements of these tasks. 
A second approach \cite{encoding} proposes to extract the features 
from the original image or from one of the layers' outputs using an encoder and transmit the data to the centralized server for inference. This approach prevents the exposure of irrelevant information to the untrusted party that may use it for unwanted inferences. However, in some cases, the amount  of  extracted  features  can be  sufficient  for  adversary to recover the original sample. Whereas, less  may  also  result  in  low  classification  accuracy. 
We highlight also that the large amount of extracted data offloaded to remote servers may become challenging, especially for systems with unstable bandwidth availability. 

Another mainstream direction to preserve the privacy of the inference is the differential privacy \cite{diffPriv}, which consists of adding noise to the input data in order to obscure it. A similar technique to secure the inference process is the encryption, which is introduced in \cite{crypto} and \cite{crypto2}. This approach proposes to infer encoded data, in order to guarantee the non-leakage of the private information. The main drawbacks of the two aforementioned techniques are the potential accuracy loss, the computation overhead, and the incompatibility of these tasks with some DNN operations \cite{security}.
Finally, many efforts focused on building a Trusted Execution Environment (TEE) on-device to run the DNN inference privately. This approach consists of establishing an isolated region implemented in the main processor, in order to ensure the confidentiality of the data and the model. In this way, untrusted devices cannot access the TEE content. However, this requires special architecture support on the device, and careful crypto key management, which may not be adequate to all IoT data-generating devices.
Moreover, the TEE approach incurs a computation overhead and a potential accuracy loss, as shown in \cite{TEE1} and \cite{TEE2}.


Scheduling pervasive IoT devices and designing a collaboration strategy to conduct online inference tasks  while being constrained by computation and memory resources in addition to privacy issues, is a complex resource allocation problem. Recently, a lot of research works have used the RL for a variety of applications and systems, spanning from wireless networks \cite{RL1,RL11} and mobile edge computing \cite{RL3} to ride-sourcing services \cite{RL2} and crowdsourcing systems \cite{RL4}, owing to the ability of reinforcement learning to efficiently solve real-time and dynamic problems. The power of the RL technique is that it is independent from any static knowledge about the system and it relies on the try-and-error interaction with the environment to learn the most effective actions and adapt to any system fluctuation. On these basis, reinforcement learning will be used as a tool to solve our resource allocation environment.

Compared to previous works, our contribution is distinctive as it uses the distribution of the CNN networks, first to minimize the classification latency and make it possible on lightweight devices existing at the proximity of data-generating device and second to mitigate the privacy issues incoming from multi-party processing. In fact, we propose to partition the private data into small features and distribute it to multiple participants, in such a way any malicious device does not own enough information to recover the original output with high accuracy. Our work outperforms other approaches by its ability to obfuscate the data, while using only resource constrained IoT devices and without affecting the accuracy of the classification or adding any computation overhead.The contributions and the novelty of this paper can be summarized as follows: 
(1) Testing black-box attacks on different partition scenarios using four different datasets and state-of-the-art CNN networks, to derive privacy vulnerabilities and estimate the amount of data per device that lowers the similarity with the original sample in case of attacks, (2) Proposing a strategy that reduces the inference delay while respecting the privacy requirements of the system, (3) Designing an  RL-based solution, characterized by its online decisions and its performance towards the optimal framework.

\section{Privacy-aware Distributed CNN for IoT devices}\label{deep_privacy}
In this section, we introduce our privacy-aware distributed deep inference approach for low latency IoT systems. For that, we start by conducting our deep empirical experiments to test the vulnerability of different partitions against black-box attacks. Understanding the privacy threats is followed by designing the system model and the formulation of the problem as an optimization. To relax the NP-hard optimal solution, we shaped our problem as a reinforcement learning system that is appropriate for online CNN distribution and resource allocation.
\subsection{Black-box Inversion Attack Against Distributed CNNs}\label{emperical_results}
In this paper, we will study only the CNN vulnerability against black-box attacks. The threat model and the adversary capabilities are described as follows:
\newline
\textbf{Threat model:}
Let $\mathcal{I}=\{I_1,...,I_D\}$ present the set of $D$ ubiquitous and heterogeneous IoT devices collaborating to compute different feature maps of the CNN and let $f_{\theta}(x)$ denote the CNN network to be distributed on this pervasive system. We consider that the data-generating device is trusted and correctly processes the first layer of the model without leaking the original sample to other parties. The rest of the CNN network is composed of segments $f^i_{\theta}(x)$, each one can be executed by an untrusted device, as a part of the inference. The output of each segment is potentially fed to one or multiple participants. We assume that all participants are malicious and may attempt to attack the received input, which is reasonable in an IoT system that is not controlled by an operator.
\newline
\textbf{Adversary capabilities:} We assume that any untrusted device $I_i$ strictly follows the collaborative inference protocols. Particularly, it receives $in=f^j_{\theta}(y)$ from $I_j$ and generates $out=f^i_{\theta}(in)$, without being able to compromise the task processed by $I_j$. Moreover, in black-box settings, the adversary has no knowledge about the weights, parameters or structure of the segments owned by previous participants except $f^i_{\theta}$, the input $in$, and the distribution of the data, which makes it more challenging to inverse the data. Meaning, without knowing the model parameters, the malicious participant cannot apply a gradient descent optimization to map the received input to the original image collected by the source device. Yet, after receiving the output feature maps from a previous participant, key insights and information can be extracted to reconstruct the sensitive data \cite{reverse}. Authors in \cite{security} proved that even without collecting any insight, an inverse network can be trained to identify the inversion mapping from the received input to the original image.  Conceptually, an inverse network, created from a segment $f^{i+1}_{\theta}$, can be defined as $g=(f^{i+1}_{\theta})^{-1}$ and trained with the inputs received from a previous device $in=f^i_{\theta}(x)$, where $x$ denotes the original image and the desired inversion output. The loss function is expressed as follows:
\begin{equation}\label{loss}
\footnotesize
    \begin{aligned}
L(g^{(t)})=\frac{1}{k}\sum_{i=1}^k \norm{g(f_{\theta}(x_i))-x_i}^2_2
    \end{aligned}
\end{equation}
 However, unlike \cite{security}, where authors studied possible attacks to recover the original data based on the output of full layers, we trained  various  networks  when  having only  few feature  maps of the intermediate data. The black-box attack is performed in three steps: (1) collect the training set of the inverse network by participating in the inference process: $(f^i_{\theta}(x_1),f^i_{\theta}(x_2),...,f^i_{\theta}(x_n))$; (2) train the network; and (3) recover the original image $x$ by inputting $in$ received from a previous participant.  Note that the quality of the inversion is higher when the adversary has access to the original training dataset or the distribution of the samples as done in this work. Still, it is also possible to recover the data by assuming that the images have standard Gaussian distribution, as shown in \cite{security}. 
\begin{figure}[h]
	\centering
	\mbox{
 \subfigure[\label{black-box1}]{\includegraphics[scale=0.51]{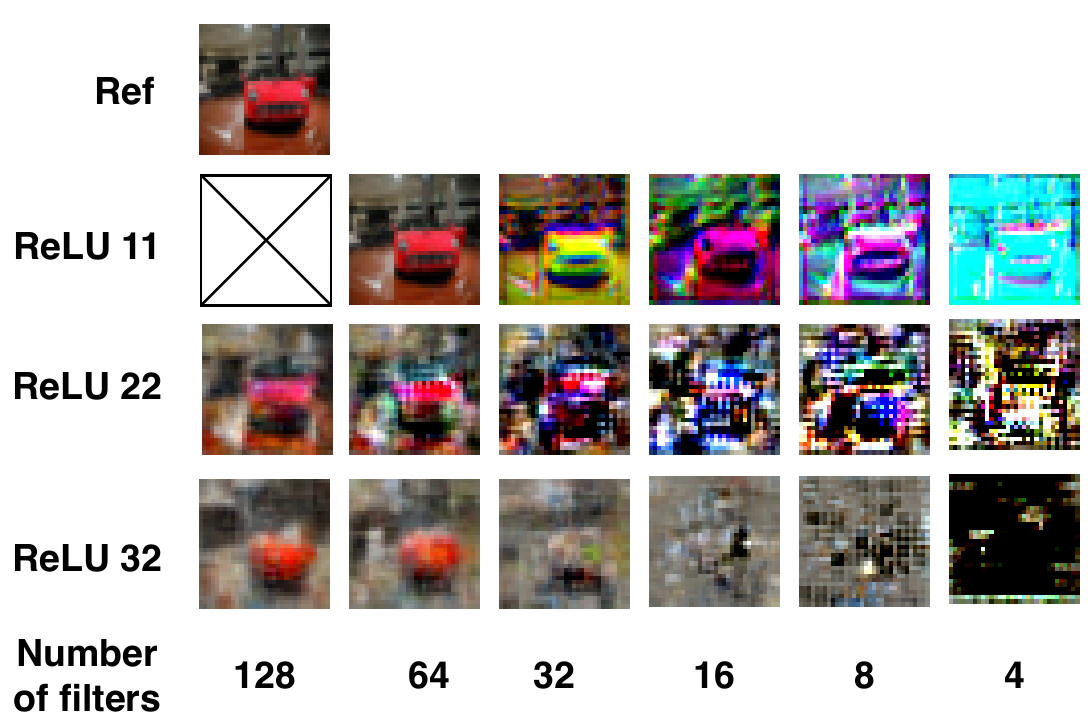}}
}
\hspace{-0.3cm}
	\mbox{
		\subfigure[\label{black-box2}]{\includegraphics[scale=0.51]{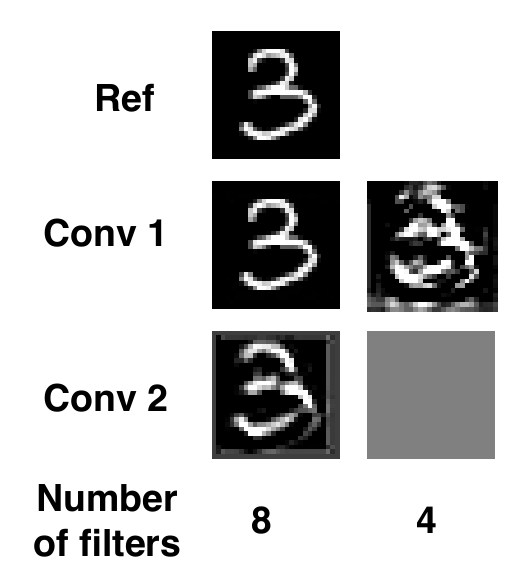}}
}\\
\vspace{-0.2cm}
	\mbox{
		\subfigure[\label{black-box3}]{\includegraphics[scale=0.51]{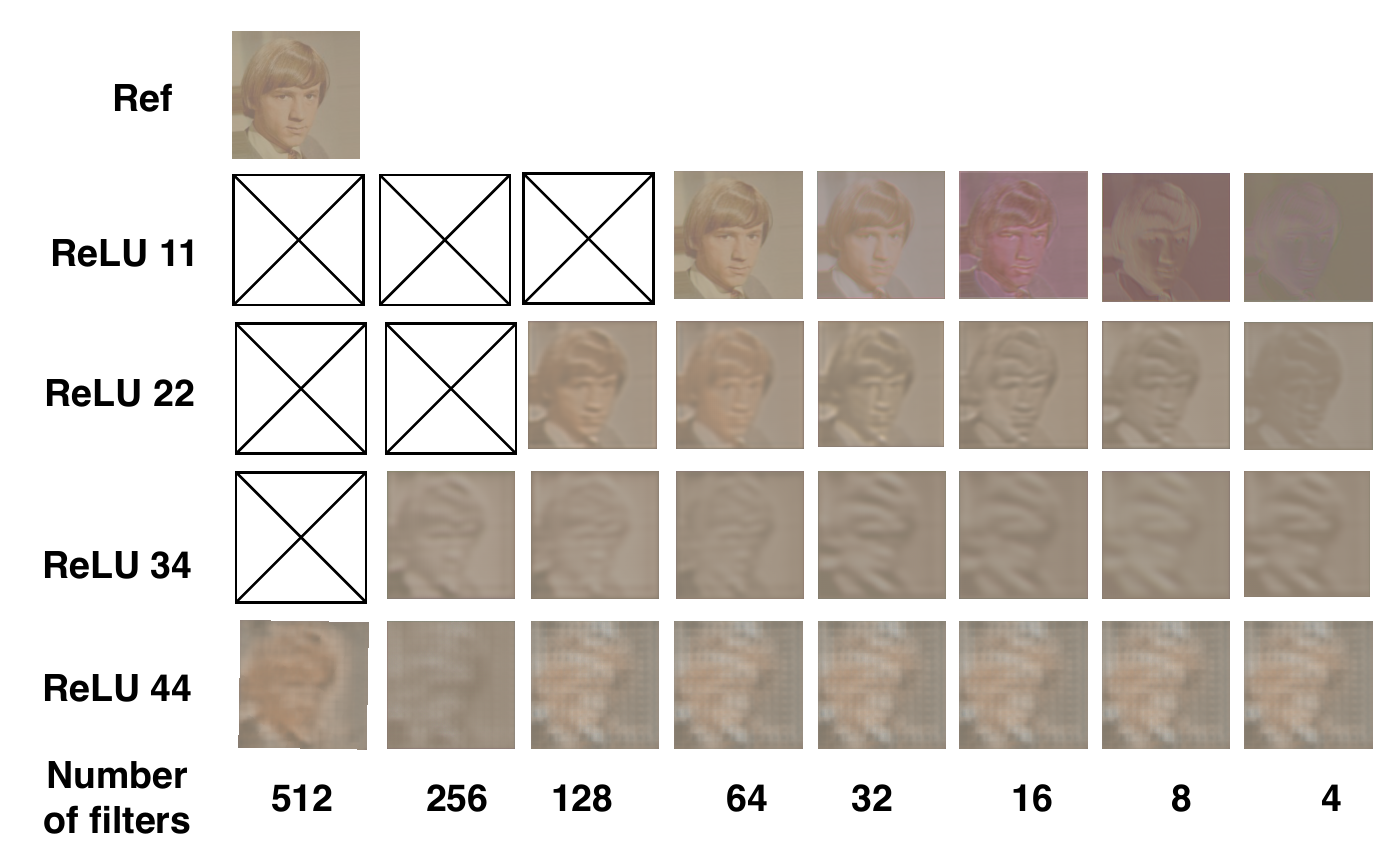}}}\\
		\vspace{-0.2cm}
			\mbox{
	\subfigure[\label{black-box4}]{\includegraphics[scale=0.51]{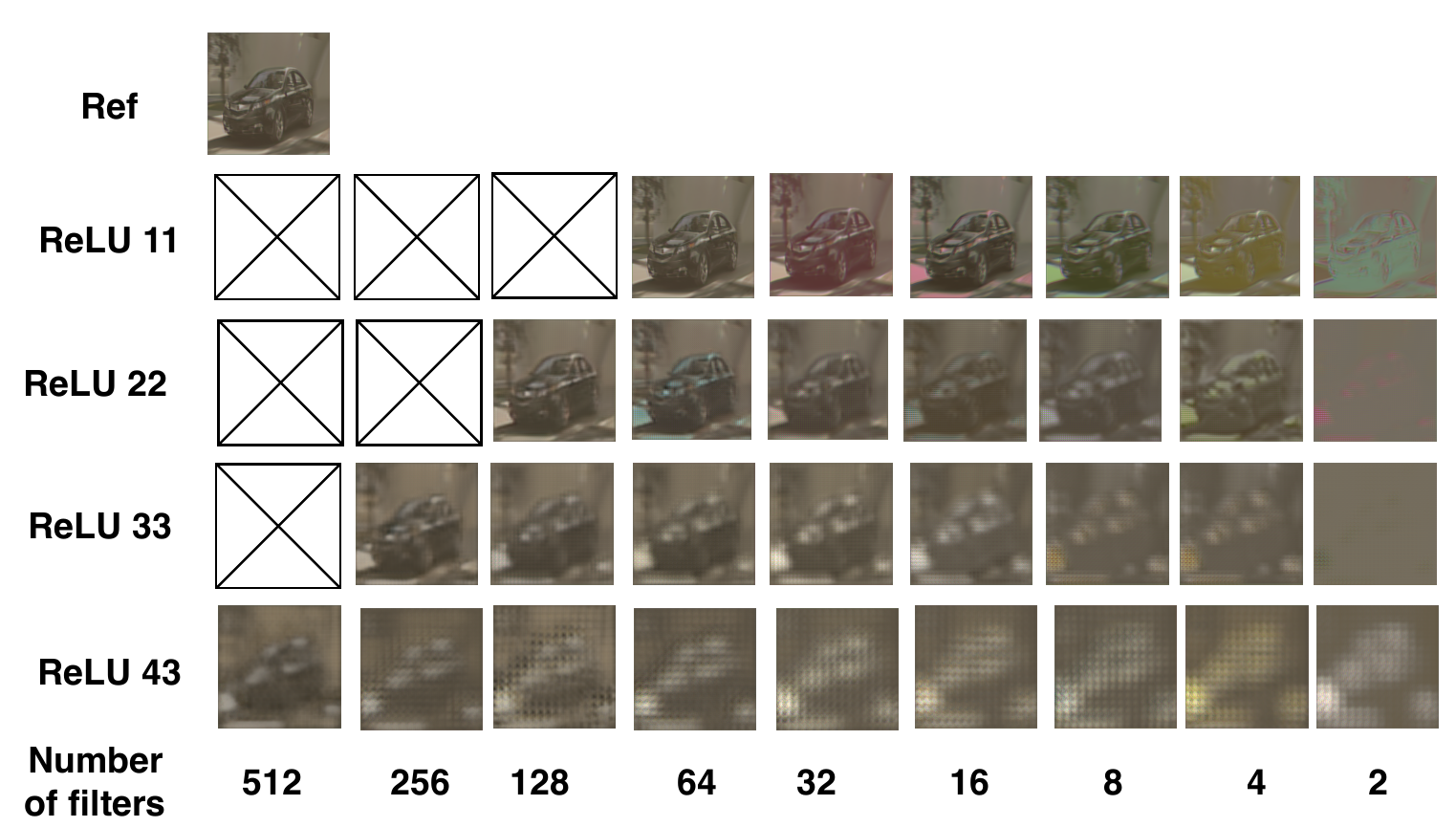}}}
	\caption{Recovered images for different partition scenarios: (a) CIFAR, (b) MNIST, (c) CELEBA, (d) Stanford CARS.}
	\vspace{-0.4cm}
	\label{black-box}
\end{figure}

The black-box attack is applied to recover images of four standard CNN benchmark datasets: CIFAR10, MNIST, CELEBA, and Stanford CAR. More specifically, we trained MNIST using LeNet, which is a small CNN composed of 2 convolutional and 3 fully connected layers. CIFAR10 dataset is trained with a larger CNN (6 convolutional and 2 fully connected layers),  VGG19 (16 convolutional  and 3 fully connected layers) is applied to CELEBA, and  VGG16 (13 convolutional and 3 fully connected layers) to Stanford CAR. In our experiments, we follow the  CIFAR, MNIST, CELEBA, and CAR dataset split standard for training and testing  and we choose ADAM as optimizer, as done in \cite{security}. Figure \ref{black-box} presents the inversion results of different black-box attack scenarios. ReLU $ij$ in the figure  indicates the $j$ ReLU in the $i$ block of the CNN (e.g. ReLU 11 is the first ReLU in the first block.). Without loss of generality, we illustrate the recovered data of layers where the number of filters changes (e.g., conv22/ReLU22 of CIFAR CNN receives 64 feature maps and outputs 128 feature maps). In addition, we present examples of recovered images, if a participant only receives few filters. Accordingly, Table \ref{ssim} presents the similarity level between the original samples and the recovered data, when applying black-box attacks on different datasets and under different splitting scenarios. In this paper, we use the average Structural Similarity Index (SSIM) as a metric that measures the quality degradation of the reversed data compared to the original one.

\begin{table}[h]
\center
\footnotesize
\tabcolsep=0.11cm
\caption{SSIM of inversion black-box attack against different datasets.}
\label{ssim}
\begin{tabular}{|l|l|l|l|l|l|l|l|l|l|}
\hline
\begin{tabular}[c]{@{}l@{}}Nb filters\\ per device/\\Layer\end{tabular} & 512 & 256 & 128 & 64 & 32 & 16 & 8 & 4 & 2 \\ \hline
\begin{tabular}[c]{@{}l@{}}CIFAR\\ ReLU 11\end{tabular} & \cellcolor[HTML]{C0C0C0}{\color[HTML]{EFEFEF} } & \cellcolor[HTML]{C0C0C0}{\color[HTML]{EFEFEF} } & \cellcolor[HTML]{C0C0C0}{\color[HTML]{EFEFEF} } & 0.99 & 0.6 & 0.56 & 0.4 & 0.3 & 0.26 \\ \hline
\begin{tabular}[c]{@{}l@{}}CIFAR\\ ReLU22\end{tabular} & \cellcolor[HTML]{C0C0C0} & \cellcolor[HTML]{C0C0C0} & 0.86 & 0.7 & 0.49 & 0.34 & 0.13 & 0.1 & 0.07 \\ \hline
\begin{tabular}[c]{@{}l@{}}CIFAR\\ ReLU32\end{tabular} & \cellcolor[HTML]{C0C0C0}{\color[HTML]{EFEFEF} } & \cellcolor[HTML]{C0C0C0}{\color[HTML]{EFEFEF} } & 0.6 & 0.51 & 0.41 & 0.18 & 0.08 & 0.07 & 0.01 \\ \hline
\begin{tabular}[c]{@{}l@{}}MNIST\\ Conv1\end{tabular} & \cellcolor[HTML]{C0C0C0}{\color[HTML]{EFEFEF} } & \cellcolor[HTML]{C0C0C0}{\color[HTML]{EFEFEF} } & \cellcolor[HTML]{C0C0C0}{\color[HTML]{EFEFEF} } & \cellcolor[HTML]{C0C0C0}{\color[HTML]{EFEFEF} } & \cellcolor[HTML]{C0C0C0}{\color[HTML]{EFEFEF} } & \cellcolor[HTML]{C0C0C0}{\color[HTML]{EFEFEF} } & 0.99 & 0.28 & \cellcolor[HTML]{C0C0C0}{\color[HTML]{EFEFEF} } \\ \hline
\begin{tabular}[c]{@{}l@{}}MNIST\\ Conv2\end{tabular} & \cellcolor[HTML]{C0C0C0}{\color[HTML]{EFEFEF} } & \cellcolor[HTML]{C0C0C0}{\color[HTML]{EFEFEF} } & \cellcolor[HTML]{C0C0C0}{\color[HTML]{EFEFEF} } & \cellcolor[HTML]{C0C0C0}{\color[HTML]{EFEFEF} } & \cellcolor[HTML]{C0C0C0}{\color[HTML]{EFEFEF} } & \cellcolor[HTML]{C0C0C0}{\color[HTML]{EFEFEF} } & 0.73 & 0 & \cellcolor[HTML]{C0C0C0}{\color[HTML]{FFFFFF} } \\ \hline
\begin{tabular}[c]{@{}l@{}}CELEBA\\ ReLU11\end{tabular} & \cellcolor[HTML]{C0C0C0}{\color[HTML]{EFEFEF} } & \cellcolor[HTML]{C0C0C0}{\color[HTML]{EFEFEF} } & \cellcolor[HTML]{C0C0C0}{\color[HTML]{EFEFEF} } & 0.96 & 0.81 & 0.66 & 0.27 & 0.09 & 0.1 \\ \hline
\begin{tabular}[c]{@{}l@{}}CELEBA\\ ReLU22\end{tabular} & \cellcolor[HTML]{C0C0C0}{\color[HTML]{EFEFEF} } & \cellcolor[HTML]{C0C0C0}{\color[HTML]{EFEFEF} } & 0.76 & 0.69 & 0.71 & 0.59 & 0.59 & 0.4 & 0.4 \\ \hline
\begin{tabular}[c]{@{}l@{}}CELEBA\\ ReLU34\end{tabular} & \cellcolor[HTML]{C0C0C0}{\color[HTML]{EFEFEF} } & 0.56 & 0.51 & 0.47 & 0.49 & 0.46 & 0.45 & 0.45 & 0.45 \\ \hline
\begin{tabular}[c]{@{}l@{}}CELEBA\\ ReLU44\end{tabular} & 0.26 & 0.39 & 0.3 & 0.3 & 0.3 & 0.3 & 0.3 & 0.3 & 0.3 \\ \hline
\begin{tabular}[c]{@{}l@{}}CAR\\ ReLU11\end{tabular} & \cellcolor[HTML]{C0C0C0}{\color[HTML]{EFEFEF} } & \cellcolor[HTML]{C0C0C0}{\color[HTML]{EFEFEF} } & \cellcolor[HTML]{C0C0C0}{\color[HTML]{EFEFEF} } & 0.98 & 0.92 & 0.93 & 0.88 & 0.69 & 0.04 \\ \hline
\begin{tabular}[c]{@{}l@{}}CAR\\ ReLU22\end{tabular} & \cellcolor[HTML]{C0C0C0}{\color[HTML]{EFEFEF} } & \cellcolor[HTML]{C0C0C0}{\color[HTML]{EFEFEF} } & 0.83 & 0.74 & 0.59 & 0.47 & 0.5 & 0.4 & 0.26 \\ \hline
\begin{tabular}[c]{@{}l@{}}CAR\\ ReLU33\end{tabular} & \cellcolor[HTML]{C0C0C0}{\color[HTML]{EFEFEF} } & 0.68 & 0.58 & 0.58 & 0.55 & 0.46 & 0.31 & 0.18 & 0.18 \\ \hline
\begin{tabular}[c]{@{}l@{}}CAR\\ ReLU43\end{tabular} & 0.36 & 0.33 & 0.30 & 0.36 & 0.36 & 0.31 & 0.29 & 0.34 & 0.33 \\ \hline
\end{tabular}
\end{table}

Based on our empirical results depicted in the illustrative Figures and the Table \ref{ssim}, several conclusions can be deducted. First, the malicious participant can perfectly inverse the input data to recover the original image, when attacking the shallow layers, if he/she receives all feature maps generated from the previous layer. As an example, the black-box attack can recover the images with high quality, when applied to layers ReLU11 or ReLU22 of CIFAR, CELEBA or Stanford Car trained CNNs. Similarly, the quality of the inverted images of MNIST dataset is very high, in both layers. Meanwhile, recovering sensitive data at deep layers results in blurred images even when collecting all feature maps. However, in some scenarios, the inverted image can still be distinguished, e.g, ReLU44 layer of CELEBA dataset. To secure the intermediate private data from inversion attacks, we propose that participants do not receive the entire output produced by a previous layer, as done in \cite{dis5}. Instead, each IoT participant obtains only few output feature maps that are not sufficient to recover the original data. In this way, the layer’s tasks (conv, ReLU, etc.) are distributed and executed jointly, while ensuring the privacy of the system. As we can see in Table \ref{ssim}, the output of a shallow layer that is easily inverted if allocated in one device, can be protected by distributing it into multiple devices. For example, if the output feature maps of the ReLU11 are shared with two devices (32 maps each) or four devices (16 maps each), the potential attacker may succeed to get a distinguished image even if it is noisy (see Figures \ref{black-box1},\ref{black-box3},\ref{black-box4}). When this output is distributed to 8 devices (8 maps each), the data recovering becomes unfeasible. In the case of Stanford CARs dataset, the shallow layers should be deployed either on the trusted source devices or distributed to a higher number of participants (2 maps per device). As we go deeper into the network, it becomes more difficult to recover the input data and hence less participants are involved to ensure the data privacy.  Without loss of generality, the level of privacy will be calculated in terms of SSIM. More specifically, when the  SSIM similarity is low, the degradation is high and the original image cannot be distinguished, which means that the attacker could not recover the data. We call this a high privacy level. When the SSIM increases, this means that the malicious participant may succeed to recover the data with higher quality and we call this a low privacy level. From our empirical study, we learned that a low similarity can be accomplished, when a small number of feature maps are assigned to each device. However, this comes at the expense of high dependency and data transmission between participants.
Therefore, in the rest of the paper, we will establish a trade-off between the privacy level, the latency and data dependency while executing the inference.
\subsection{System model}
Our pervasive surveillance system is composed of a set of image-generating IoT devices (e.g. cameras), namely $\mathcal{S}=\{s_1,...,s_t\}$, that are responsible to control an area by capturing images and requesting feedback or classification (e.g. face recognition, accidents detection or cars plates control). In this study, we suppose that we have $N$ CNNs serving as classifiers and that each camera device requests the execution of only one CNN inference type related to its controlling target. Moreover, let $ch$ denote the number of channels corresponding to the captured images ($ch=3$ for RGB images) and we remind that $\mathcal{I}=\{I_1,...,I_D\}$ presents the set of $D$ ubiquitous and heterogeneous IoT devices collaborating to compute different segments of the CNN. Limited resources characterize different IoT participants, particularly, each device has a memory usage denoted by$\Bar{m_i}$, a computation capacity defined as $\Bar{c_i}$, and bandwidth availability equal to $\Bar{b_i}$. Without loss of generality, the image-generating devices will not participate in computing the intermediate segments of the convolutional network to save their resources (e.g., memory, bandwidth, energy, etc.) for image capturing, processing and offloading. We note that involving the source devices in the collaborative inference does not impact the system design. 

In this paper, all IoT devices participating in our system communicate using different transmission rates  equal to $\rho_i$. The analysis later leverages the same model for all wireless links between devices e.g. WiFi. However, such a model can be changed if the devices use different transmission technology e.g. cellular. Furthermore, we assume that the transmission channel between devices is perfect or for imperfect channel, a reliable protocol is used between devices. Finally, we assume that all devices are inter-connected and can reach each other. Else, a constraint on the communication range can be added.

As noted previously, our IoT system uses $N$ CNNs for surveillance purposes. We define $L_{n_j}$ as the number of layers of the CNN $n_j \in \{1....N\}$, where each layer $l^k_{n_j} \in \{1,...,L_{n_j}\}$ results in $P_{l^k_{n_j}}$ feature maps. These output feature maps are considered as the segments to be processed by the   next layer $l^{k+1}_{n_j}$. Accordingly, each segment presents a memory demand $m^{k,p}_j$ and each related task (e.g. conv.) requires a computational load $c^{k,p}_j$. In this work, we define the computational load as the number of multiplications required to accomplish the layer goals. Therefore, the computation demand of layers, where no  multiplication is performed (e.g., ReLU, maxpool), will be neglected \cite{b1}; and the computation requirement of a convolution layer segment $p_{l^k_{n_j}}$ will be calculated as follows:
\begin{equation}\label{eq:1}
    \begin{aligned}
c^{k,p}_j=S_{k+1}  \times P_{l^{k+1}_{n_j}}  \times o_{k+1},
    \end{aligned}
\end{equation}
where $S_{k+1}$ presents the spatial size of the filter defined by the layer $l^{k+1}_{n_j}$ and $o_{k+1}$ denotes the spatial size of the output feature map. The computational requirements of a fully-connected layer can be presented as follows: 
\begin{equation}\label{eq:2}
    \begin{aligned}
 c^k_j=n^*_{k-1}  \times n^*_k,
 \end{aligned}
\end{equation}
where $n^*_k$ and $n^*_{k-1}$ respectively denote the number of neurons forming the layers $l^{k}_{n_j}$ and $l^{k-1}_{n_j}$. As our goal is to maintain the privacy of the distributed system and for simplicity purposes, each fully connected layer will be computed by a single IoT participant as the shared output between such layers cannot be attacked and recovered \cite{security}. We emphasize that this will only be applied to fully connected layers. Accordingly, $\{P_{l^{k}_{n_j}},...,P_{l^{L}_{n_j}}\}$ are equal to 1 if  $l^{k}_{n_j}$ is a fully connected layer. Furthermore, the memory occupation of a segment $p_{l^{k}_{n_j}}$ can be evaluated as the number of stored weights $W^{k,p}_j$ multiplied by the memory word-length (number of bits required to store each weight) \cite{b1}; which is presented as follows:
\begin{equation}\label{eq:3}
    \begin{aligned}
m^{k,p}_j=W^{k,p}_j  \times b,
 \end{aligned}
\end{equation}
where $b$ denotes the memory word-length, which is equal to 4 bits if the single-precision flop data type is used \cite{b3}. Finally, we define $O^k_{i1,i2}$ as the memory occupation of the data generated by the participant $I_{i1}$ while processing segments of the layer  $l^k_{n_j}$ and communicated to $I_{i2}$ to participate in $l^{k+1}_{n_j}$ computation. Table \ref{notations} summarizes different key notations introduced in this paper.
\begin{table}[!h]
\caption {List of notations.}
\label{param}
\centering
\begin{tabular}{l p{6 cm}}
\hline
 \textbf{Notation} &  \textbf{Description}   \\ [0.5ex] 
 \hline
$\mathcal{S}=\{s_1,...,s_t\}$& Set of cameras,  vector. \\ 
$N$ & Number of CNNs,  scalar.  \\ 
$ch$& Number of channels of the samples,  scalar.\\
$\mathcal{I}=\{I_1,...,I_D\}$& Set of IoT participants,  vector.\\
$D$& Number of IoT participants,  scalar.\\
$\Bar{c_i}$& Computation capacity of $ I_i$, scalar.\\
$\Bar{m_i}$& Memory capacity of $ I_i$, scalar.\\
$\Bar{b_i}$& Bandwidth availability of $ I_i$,  scalar.\\
$\rho_i$& Data rate of $ I_i$, scalar.\\
$L_{n_j}$& Number of layer of the CNN $ n_j $, scalar.\\
$l^k_{n_j}$& Layer $k$ of the CNN $ n_j $, scalar.\\
$P_{l^k_{n_j}}$& Number of output feature maps of $ l^k_{n_j}$, scalar.\\
$p_{l^k_{n_j}}$& The feature map index,  scalar.\\
$m^{k,p}_j$& Memory occupation of the segment $p$ of layer $k$ and CNN $n_j$, scalar.\\
$c^{k,p}_j$& Computational demand of the segment $p$ of layer $k$ and CNN $n_j$,  scalar.\\
$O^k_{i1,i2}$& Shared data between $I_{i1}$ and $I_{i2}$, scalar.\\
$S_{k}$& Spatial  size  of  the  filter in layer $l^k_{n_j}$,  scalar.\\
$o_{k+1}$& Spatial size of the output segment,  scalar.\\
$n^*_k$& Neurons' number in a fc layer $l^k_{n_j}$, scalar.\\
$b$& Memory word length, scalar.\\
$W^{k,p}_j$&Stored weights of the segment p, scalar.\\
$r$& Inference request, scalar.\\
$RQ$& Set of requests, vector.\\
$M$& Maximum number of layers in all CNNs, scalar.\\
$A^i_{r^*,l,p}$ & Decision variable, binary. \\
$C^l$ & Equal to 1 if $l$ is a convolutional layer,  binary. \\
$Ac^l$ & Equal to 1 if $l$ is an activation or maxpool layer,  binary. \\
$F^l$ & Equal to 1 if $l$ is a fully connected layer,  binary. \\
$e(i)$ & Number of multiplications $I_i$ can carry,  scalar. \\
$Nf^l(SSIM)$ & Tolerated number of feature maps per device corresponding to the required SSIM, scalar. \\
$SP_{j}$ & Split point, after which the distribution for privacy is not needed, scalar. \\
$S$ & Set of states, vector. \\
$A$ & Set of actions, vector. \\
$P$ & State transition probability, scalar.\\
$R$ & Immediate reward, scalar.\\
$\gamma$ & Discount factor,  scalar.\\
\hline
\label{notations}
\end{tabular}
\vspace{-3mm}
\end{table}
\begin{figure*}[!h]
\centering
	\includegraphics[scale=0.67]{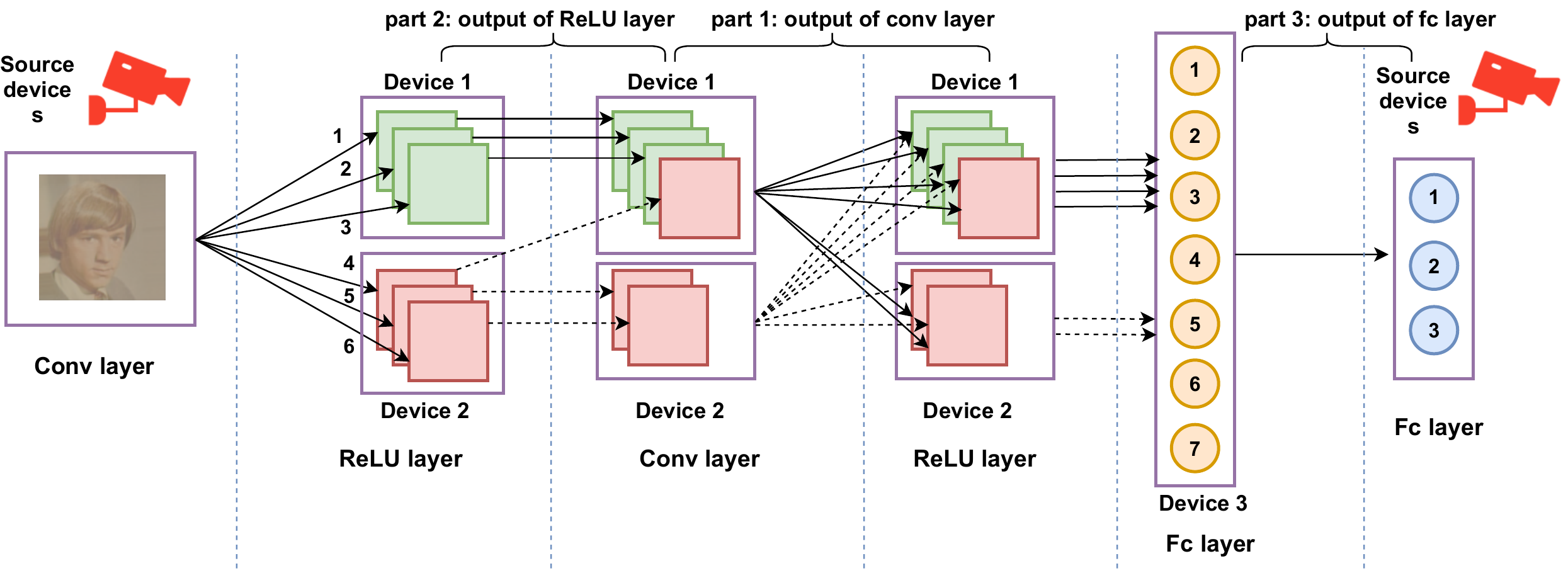}
	\caption{ Descriptive of equation (6): A scenario of CNN distribution among IoT devices, where the structure of the network is simplified for the sake of clarity.}
	\label{scenario1}
	\vspace{- 3 mm}
\end{figure*}
 
Next, the optimal placement of different CNN segments among the IoT devices participating in the collaborative system is formulated, while taking into consideration the available resources and the privacy requirements against black-box attacks.  The optimization is executed periodically to cover the variation of the network, such as the inclusion and removal of participants, or the change of the sources and the type of classifications.
\subsection{Problem formulation}\label{problem_formulation}
Our proposed privacy-aware CNN distribution strategy relies on one decision variable, namely $A^i_{r^*,l,p}$. $A^i_{r^*,l,p}$ is equal to 1, if the participant $I_i$ executes the segment/feature map $p$ produced by the layer $l$ of the request $r$; 0 otherwise.
Let $r \in RQ$ denote the request for image classification. In practice, $r$ represents the index of the source device that requested the inference. We emphasize that each source device is related to a single CNN that we denote by $r^*$. Finally, for simplicity purposes, we suppose that the IoT devices participating in the collaborative inference system are fixed during the optimization to keep the available data rates static. The objective function models the latency of computing the set of requests $RQ$. This latency is defined as the time necessary to transfer the output of layers between participants and to compute different tasks. Hence, the objective is the minimize the following function:
\begin{equation}\label{obj}
\footnotesize
    \begin{aligned}
   L_{IoT}=\sum\limits_{r \in RQ}\sum\limits_{l=2}^{L_{r^*}-2} max( \frac{O^{l-1}_{i,j}}{\rho_i}+t^{r^*,l,j}_c, \forall I_i,I_j \in  \mathcal{I}) +t_s+t_f,\qquad\qquad\qquad\qquad\qquad\qquad\qquad\qquad\qquad
    \end{aligned}
\end{equation}
\begin{equation}\label{O}
\footnotesize
    \begin{aligned}
\text{where: }O^{l}_{i,j}= [( \overbrace{C^l\times o_{l}^2 \times min(1,\sum\limits_{p_1=1}^{P_{l-1}}A^{i}_{r^*,l,p_1}) \times \sum\limits_{p_2=1}^{P_{l}}A^{j}_{r^*,l+1,p_2}}^{\text{part 1}}+\\\overbrace{Ac^l \times o_{l}^2 \times \sum\limits_{p_1=1}^{P_{l-1}}A^{i}_{r^*,l,p_1}A^{j}_{r^*,l+1,p_1}}^{\text{part 2}}+\overbrace{F^l \times n^*_{l} \times A^{i}_{r^*,l,1}A^{j}_{r^*,l+1,1})}^{\text{part 3}}\\**b]*\mathrm{1}_{i\neq j}\qquad\qquad \qquad \qquad \qquad \qquad \qquad \qquad \qquad \qquad \qquad \qquad  
    \end{aligned}
\end{equation}
The objective function in Eq. (\ref{obj}) is composed of four parts:

(1) The transmission latency of intermediate feature maps generated by layer $l-1$ and used to process the next layer $l$. This latency is presented by $\frac{O^{l-1}_{i,j}}{\rho_i}$ $\times$ $\rho_i$ denotes the data rate of the considered transmission technology equipped with the device $I_i$ and $O^{l-1}_{i,j}$ is the size of segments shared between $I_i$ and $I_j$. $O^{l-1}_{i,j}$ is presented by Eq. (\ref{O}), where $C^l$ is a binary variable equal to 1, if the layer $l$ is a conv layer; $F^l$ is equal to 1, if $l$ is a fully connected layer; and $Ac^l$ is equal to 1, if $l$ is an activation layer or a maxpool.
\begin{itemize}
    \item Part 1: if the current layer $l$ is convolutional, each output feature map of $l$ is created by summing all the input segments after applying one of the conv tasks. Hence, if a device $I_i$ participates in the layer $l$ computation, it is responsible to produce $p_{l_{n_j}}$ segments and to offload them to the next layer participants depending on their number of assigned input features.  It means that the size of the data shared between any two devices $I_i$ and $I_j$, is equal to the size of all output feature maps of the layer $l$ (input of the layer $l+1$) assigned to the receiver participant $I_j$. This statement is only valid if the sender $I_i$ computes at least 1 feature map of $l$. The output of the convolutional layer transmitted  between any two devices is illustrated by the first part of Eq. (\ref{O}) and in Figure \ref{scenario1}.
    \item  Part 2: if the current layer is an activation or maxpool layer,  the shared data is equal to the size of the output feature maps of $l$ (input of $l+1$) assigned to $I_j$ and generated by $I_i$. This can be seen in the second part of Eq. (\ref{O}) and in Figure \ref{scenario1}.
    \item  Part 3: regarding the fully connected layer, we assumed in this paper that it is computed by one device and the output is offloaded to only one device, as it is not reversible. Hence,  the shared data between the two participants is equal to the size of the entire output of the layer $l$, which is illustrated by the third part of Eq. (\ref{O}) and in Figure \ref{scenario1}.
\end{itemize}
$O^{l-1}_{i,j}$ is equal to 0, if $I_i = I_j$, including the case of a source device generating the image and processing the first layer. To better understand the distribution of segments between IoT participants, we present some communication scenarios in Figure \ref{scenario1}. We note that the structure of the network in the Figure is simplified for the sake of clarity.

(2) The processing latency of different segments on the IoT participants, which is expressed as follows:
\begin{equation}\label{tc}
\footnotesize
    \begin{aligned}
t_c^{r^*,l,j}=\sum\limits_{p=1}^{P_{l-1}} A^j_{r^*,l,p} \times \frac{c_{r^*}^{l,p}}{e(j)}, \qquad \forall \quad I_j \in \mathcal{I}
    \end{aligned}
\end{equation}
The computation time of the feature map $p$ on the $I_i$-th IoT device is approximated as the ratio between the computational demand $c_{r^*}^{l,p}$ required by the segment, and the number of multiplications $e(j)$ the device $I_j$ is able to carry out in one second \cite{dis5}. In practice, $e(j)$ indirectly includes the available cores per device and the presence of GPU or CPU to parallelize operations. The computation and offloading of different output feature maps of the same layer to next IoT participants are carried out  synchronously. Hence, the delay to handle each layer is calculated as the larger latency among different accomplished tasks (transmissions and computation).

(3) The source latency $t_s$, which is experienced after transmitting the output of the first layer from the source $r$ to the IoT devices executing the second layer of the CNN. We emphasize that the first layer is always computed in the source device to protect the original image. The source latency is expressed as follows:
\begin{equation}\label{ts}
\footnotesize
    \begin{aligned}
t_s=\sum\limits_{r \in RQ}\sum\limits_{p \in ch}  A^r_{r^*,1,p} \times \frac{c_{r^*}^{1,p}}{e(r)} + \sum\limits_{r \in RQ} max(\frac{O^{1}_{r,i}}{\rho_{r}},\forall I_i \in \mathcal{I}) \qquad\qquad\qquad\qquad\qquad\qquad\qquad\qquad
    \end{aligned}
\end{equation}

(4) The final latency $t_f$, which defines the required time to transmit the final segments to the last layer of the CNN and compute it. We note that the last layer should be executed on the source device to protect the privacy of the prediction, as illustrated in Figure \ref{scenario1}. $t_f$ is expressed as follows:
\begin{equation}\label{tf}
\footnotesize
    \begin{aligned}
t_f=\sum\limits_{r \in RQ}\sum\limits_{p}^{P_{L_{r^*}-1}}  A^r_{r^*,L_{r^*},p} \times \frac{c_{r^*}^{L_{r^*},p}}{e(r)}+ \qquad\qquad\qquad\\\sum\limits_{r \in RQ} max (\quad \frac{O^{L_{r^*}-1}_{i,r}}{\rho_i},\quad \forall \quad I_i \in \mathcal{I}) \qquad\qquad\qquad
    \end{aligned}
\end{equation}
Our privacy-aware distributed deep convolutional network can be formulated as follows:
\begin{subequations}
\footnotesize
	\allowdisplaybreaks
		\begin{align}
	\begin{split}
    \underset{ \begin{subarray}{c}
    (A^i_{r^*,l,p}) 
    \end{subarray}}\min \qquad L_{IoT} \qquad \qquad \qquad \qquad 
	\label{eq:optProb} 
	\end{split}\\
	\begin{split}
\text{s.t} \quad \sum\limits_{r \in RQ}\sum\limits_{l=1}^{L_{r^*}}\sum\limits_{p=1}^ {P_{l-1}}A^i_{r^*,l,p}\times m^{l,p}_{r^*}\leq \bar{m_i} \quad \forall  I_i \in \mathcal{I}\cup\{s_1..s_n\},
	\label{eq:constraint1}
	\end{split}\\
			\begin{split}
\qquad \sum\limits_{r \in RQ}\sum\limits_{l=1}^{L_{r^*}}\sum\limits_{p=1}^ {P_{l-1}} A^i_{r^*,l,p}\times c_{r^*}^{l,p}\leq \bar{c_i} \quad \forall  I_i \in \mathcal{I}\cup\{s_1..s_n\},
	\label{eq:constraint2}
	\end{split}\\
		\begin{split}
\qquad \sum\limits_{r \in RQ}\sum\limits_{l=1}^{L_{r^*}}\sum\limits_{p=1}^ {P_{l-1}} \sum\limits_{I_j \in \mathcal{I}} O^{l}_{i,j}\leq \bar{b_i} \quad \forall  I_i \in \mathcal{I}\cup\{s_1..s_n\}
	\label{eq:constraint21}
	\end{split}\\
	\begin{split}
\qquad \sum\limits_{i \in \mathcal{I}\cup\{s_1..s_n\}} A^i_{r^*,l,p}= \Bigg\{ 
    \begin{array}{ll}
        1 & \mbox{if } l \leq L_{r^*}, p \leq P_{l-1} \\
        0 & \mbox{Otherwise} 
    \end{array}
	\label{eq:constraint3}
	\end{split}\\
	\begin{split}
\qquad \sum\limits_{p=1}^ {P_{l-1}} A^i_{r^*,l,p}\leq Nf^l(SSIM)\quad \forall  I_i \in \mathcal{I},   l \leq SP_{r^*}(SSIM)
	\label{eq:constraint5}
	\end{split}\\
		\begin{split}
\qquad \sum\limits_{i \in \mathcal{I}\cup\{s_1..s_n\}}\Pi_{p=1}^ {P_{l-1}}A^{i}_{r^*,l,p} \geq F^l \times \neg F^{l-1}, \quad \forall l \leq L_{r^*}
	\label{eq:constraint10}
		\end{split}\\
		\begin{split}
\qquad A^{r}_{r^*,l,p}=
        F^l \times \neg F^{l-1},\qquad SP_{r^*}(SSIM) \leq l, \quad \forall l \leq L_{r^*}
	\label{eq:constraint31}
		\end{split}\\
	\begin{split}
	\qquad A^i_{r,l,p} \in \{0,1\} 
	\label{eq:constraint9} 
	\end{split}
	\end{align}
\end{subequations}
 \begin{figure*}[!h]
\centering
	\includegraphics[scale=0.67]{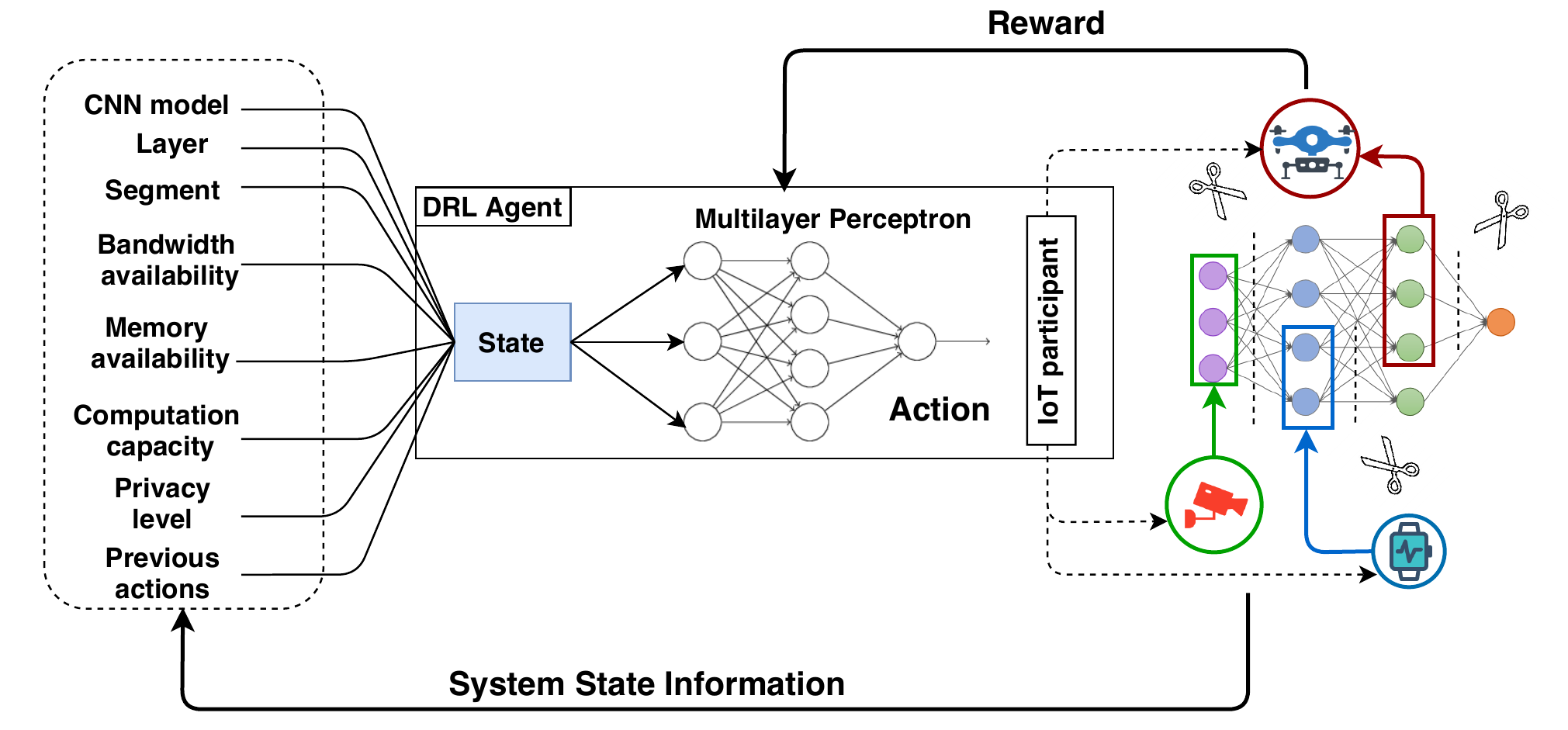}
	\caption{The DRL design for selection of distributed inference participants.}
	\label{scenario}
	\vspace{- 3 mm}
\end{figure*}
Equations (\ref{eq:constraint1}) and (\ref{eq:constraint2}) guarantee that the constraints on memory usage and
computational load are respected for all IoT participants and source devices. The constraint in Eq. (\ref{eq:constraint21}) verifies that the total transmitted output of the computed segments between any two participants respects the available bandwidth. Next, Eq. (\ref{eq:constraint3}) ensures that each feature map $p$ is processed by only one node.  In other words, to avoid computation redundancy, the segment should be assigned to only one participant. Equation (\ref{eq:constraint5}) guarantees that the number of segments assigned to any device $I_i$ should not exceed the threshold $Nf^l$. It means, the number of feature maps received by any IoT device does not import enough information that allows the malicious participant to recover the data with an accuracy larger than the maximum tolerated SSIM. To further illustrate this privacy constraint, we suppose that the tolerated SSIM is equal to 0.4. If a CNN CIFAR classification is requested (Table \ref{ssim}),  ReLU11 (64 filters) and ReLU22 (128 filters) should be distributed on 8 devices ($Nf^{11}(SSIM)=8, Nf^{22}(SSIM)=16$), and ReLU32 (128 filters)  on 4 devices ($Nf^{32}(SSIM)=32$). We only verify this constraint for the layers preceding the split point $SP_{r^*}(SSIM)$, as  following layers cannot be recovered with a higher $SSIM$, even when receiving all feature maps.  Constraint (\ref{eq:constraint10}) ensures that the first fully connected layer $F^l$ after another layer's type $\neg F^{l-1}$ is computed by one device, as chosen in our system model. More specifically, $\sum \Pi_{p=1}^ {P_{l-1}}A^{i}_{r^*,l,p}$ is equal to 1 if one device accomplishes the layer computation and 0 otherwise. In this way, when $F^l.\neg F^{l-1}$ is equal to 1, $\sum \Pi_{p=1}^ {P_{l-1}}A^{i}_{r^*,l,p}$ should be equal to 1. The next fully connected layers will be computed by one participant each, as their output is one segment by design.
If $l$ is lower than $SP_{r^*}$ (case of MNIST), the input of the first $F^l$ layer fed to one device will be exposed to inversion risks. Therefore, in such a scenario, constraint (\ref{eq:constraint31}) ensures that the first fully connected layer is computed on the source device. 

The problem in (\ref{eq:optProb}) is NP-hard, which  makes finding the mathematical solution highly difficult in terms of time. Moreover, the optimization supposes to have a full overview about incoming requests, in the studied period of time, which is not realistic. More specifically, our system environment is extremely dynamic, where the load of requests is highly variable and may follow certain statistical distributions, and the neighboring resources are volatile. Hence, opting for a static environment while solving the optimization is a not a practical assumption. Recently, RL approaches gained a lot of the research attention, particularly when applied to dynamic, large, and complex problems. In fact, the power of RL is its ability to react under unforeseen environments by  making  decisions and learning   policies  from  historical  knowledge  about the system state. Therefore, to approximate the optimal solution and relax the solving complexity, we propose an RL-based approach for dynamic and online CNN distribution.
\subsection{RL-DistPrivacy: Reinforcement learning-based Privacy-Aware Distributed CNN approach}
In this section, we formulate the privacy-aware inference distribution for low-latency applications as an RL process. The RL approaches are known by their power to handle complex problems by interacting with the environment parameters e.g., available resources on different devices, incoming requests, and number of participants, which leads to a near-optimal solution. Particularly, the RL agent learns the statistical distribution of the environment states and produces accordingly the policy that takes the most effective actions leading to the maximum accumulated rewards in addition to adapting to any change of the environment.
\\
At each time step of the distribution process, the predictive agent should select the participant that will compute the CNN segment. The decision is taken based on the CNN network, the current layer, the current segment, the available resources, and the number of segments per device. The agent ultimately aims to meet the required privacy level, while respecting the limited resources among different IoT devices. The RL learning process consists of taking allocation decisions, gaining rewards and experiencing penalties depending on the accurateness of the actions. Such process continues until reaching a convergence to the optimal policy. Accordingly, our RL-DistPrivacy problem can be shaped as a Markov Decision Process (MDP) presented by the five-tuple $(S,A,P,R,\gamma)$. $S$ denotes the system state, $A$ is the set of actions, $P$ defines the probability of transition between states, $R$ presents the immediate reward, and $\gamma$ is defined as the discount factor. In the following sections, we will define these elements in our DistPrivacy context.
\subsubsection{MDP environment design}
The MDP environment represents the surveillance IoT system that the agent interacts with. At each time step $t$, this environment is designed to receive the agent’s action $A_t$, generate the subsequent state $S_{t+1}$, and assign the immediate reward $R_{t}$. The finite sequence of time steps is defined as an episode. In this way, by experiencing the set of generated episodes, the agent learns the performance of past actions based on their assigned rewards. In our context, we define a time-step as one segment allocation in one of the IoT participants and the episode as the segments' distribution of one layer among different devices. We highlight that the training episodes are independent. In other words, each episode is initialized with a total reward equal to 0. Also, it is worth mentioning that the agent has no information about how the environment is designed. Instead, it learns the optimal allocation policy by interacting with the environment, taking actions and accordingly gaining rewards. Meaning, the optimal policy $\Pi$ is a mapping between the environment state $S_t$ and the chosen action $A_t$, i.e., $\Pi: S \rightarrow A$. Figure \ref{scenario} illustrates the design of our DistPrivacy DRL system.
\subsubsection{States and actions}
\begin{figure}[h]
 \hspace{-4mm}
\centering
	\includegraphics[scale=0.62]{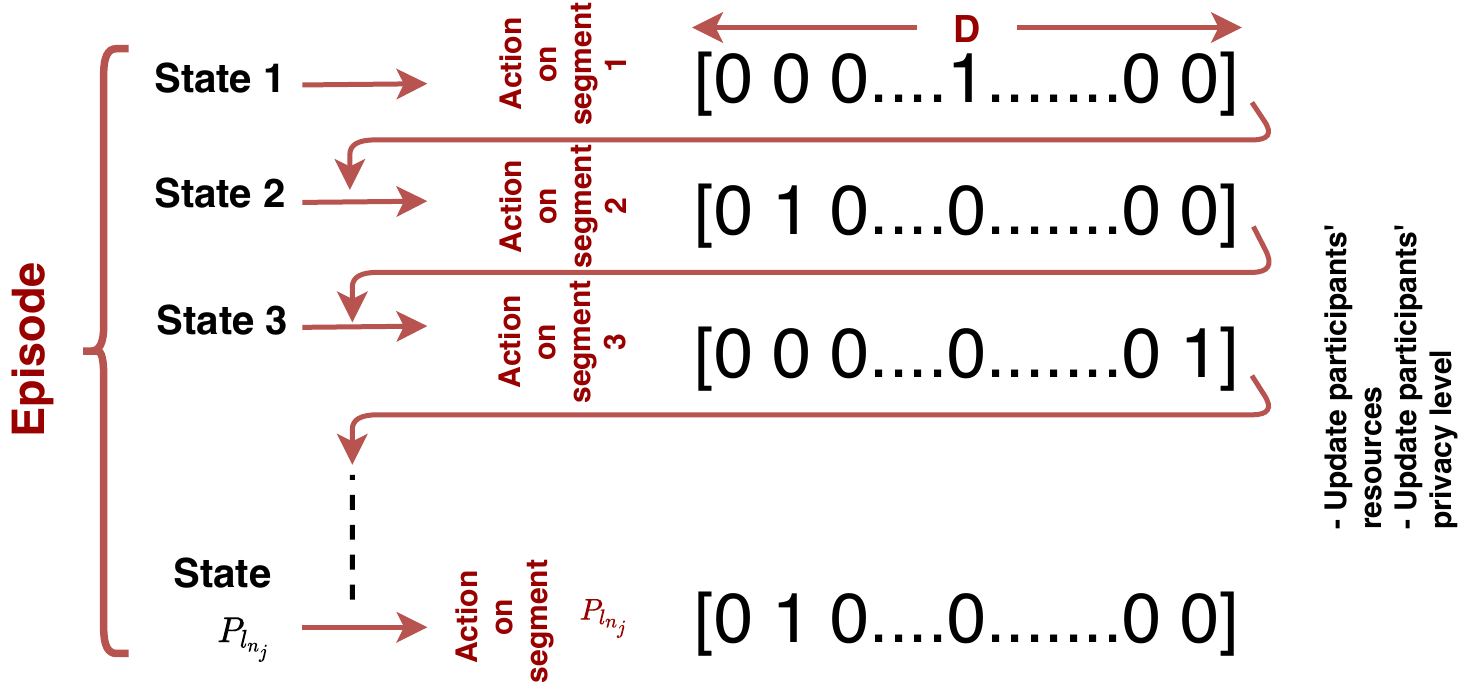}
	\caption{DRL framework.}
	\label{model}
\end{figure}

At each time step, the environment generates the set of states $S$ that comprises the current circumstances and conditions of the system. The set of states $S$ is defined as $S=\{n_i, l^{k}_{n_i}, p_{l^{k}_{n_i}}, \{\bar{c_1},...,\bar{c_D}\}, \{\bar{b_1},...,\bar{b_D}\}, \{\bar{m_1},...,\bar{m_D}\},  \\ \{\bar{L_1},...,\bar{L_D}\}, \{A_1,...,A_{P_{l^{k-1}_{n_i}}}\} \}$,where  $n_i$ is the CNN used for classification, $l^{k}_{n_i}$ is the current layer, $p_{l^{k}_{n_i}}$ is the current segment, $\{\bar{c_1},...,\bar{c_D}\}$ presents the vector of available computational resources within different IoT devices participating in the collaborative system, $\{\bar{b_1},...,\bar{b_D}\}$ defines the vector of available transmission capacities,  $\{\bar{m_1},...,\bar{m_D}\}$ is the set of available memory in each IoT device, and $\{\bar{L_1},...,\bar{L_D}\}$ denotes the number of layers computed by each participant. Additionally, as each episode deals with a single layer, it is important to have an overview about the participants of the previous layer. Hence, the set of the previous episode actions  $\{A_1,...,A_{P_{l^{k-1}_{n_i}}}\}$ should be given to the system. Following this design, the set of states becomes very large and dependable on the number of participating devices. Furthermore, states have different scales and need to be normalized. Hence, the resource availabilities and security levels sets are converted to binaries, where 1 means that the participant has enough resources and it is possible to handle another segment without any privacy risk; 0 otherwise. We note that $S$ consists only of the product of different binary states. The agent’s decision at every time step is a set of $D$ binaries. Particularly, each value in the set indicates if the related IoT device will compute the current segment computation. For example, the action space $A=[0 1 0 0 0 0]$ means that the second device will be responsible for processing the current feature map tasks. We recall that $D$ presents the number of  participants in our pervasive IoT system. After each time step and according to the selected participant, the computational, bandwidth and memory resources of the IoT devices are updated. Additionally, the number of assigned segments to the current participant is increased. This updated state of the system (i.e., resources and allocated segments) is the input to the following time step as shown in Figure \ref{model}. 
\subsubsection{Reward function}
We formulate the reward $R$ to match the objective function of the optimization that aims to minimize the latency of the CNN inference, while respecting the available IoT resources and meeting the privacy constraints required by the surveillance system. Specifically, privacy includes minimizing the number of feature maps exposed to the participants to prevent them from data inversion. Hence, at each time step after receiving the environment state $S_t$, the RL agent tries to maximize its reward by taking an action $A_t$ while respecting the following constraints:
\begin{equation}\label{reward}
\begin{footnotesize}
    \begin{aligned}
\left\{
  \begin{array}{lr}

\text{$C_1$: if } 2<l^k_{n_i}< L_{n_i}-1 \rightarrow \sum A_t=1, \qquad \text{Eq. (\ref{eq:constraint3})} &\\
    \text{$C_2$: if } A_t(j)=1 \rightarrow m^{t,k}_i\leq \bar{m_j}, c^{t,k}_i\leq \bar{c_j}, \\  O^{t-1}_{ii,j} \times 1_{ii\neq j}\leq \bar{b_j} \qquad\qquad\qquad\qquad \qquad \text{Eq. (\ref{eq:constraint1}), (\ref{eq:constraint2}), (\ref{eq:constraint21})} &\\
    \text{$C_3$: }A_t(j)=1 \rightarrow \sum_{l=1}^{t} A_l(j)<=Nf^l(SSIM),  \qquad \text{Eq. (\ref{eq:constraint5})} &
  \end{array}
\right.
    \end{aligned}
    \end{footnotesize}
    \nonumber
\end{equation}
$C_1$ indicates that only one IoT device computes the current segment of the intermediate layer through ensuring that the sum of the action vector is equal to 1, which is equivalent to the constraint (\ref{eq:constraint3}). The first and last layers, in addition to $F^l.\neg F^{l-1}$ layers in previously defined scenarios (see section \ref{problem_formulation}), are directly affected to the source devices as stated in the constraint (\ref{eq:constraint31}) and equations (\ref{ts}) and (\ref{tf}). $C_2$ indicates that the chosen participants should have enough resources to compute the assigned task, which matches the constraints (\ref{eq:constraint1}), (\ref{eq:constraint2}) and (\ref{eq:constraint21}). $C_3$ ensures that each selected participant respects the privacy requirements of the surveillance system. Particularly, a maximum number of $Nf^l(SSIM)$ feature maps can be exposed to any untrusted device, which matches the constraint (\ref{eq:constraint5}). Accordingly, the immediate reward function is defined as follows:
\begin{equation}\label{reward3}
\begin{footnotesize}
    \begin{aligned}
R_t=C_1 \times C_2 \times C_3 \times max(1,\sigma \times \sum\limits_{j=1}^{D}A_t(j) \times \sum\limits_{l=1}^{t}A_l(j))
    \end{aligned}
    \end{footnotesize}
\end{equation}
If constraint $C_1$ is not respected ($C_1=0$), the distributed system will show poor performance because the related segment either could not be handled by any IoT device or multiple participants compute the same task. To avoid such invalid situations, the reward is set to be 0. If one of the resource or privacy constraints  $C_2$ or $C_3$ is not respected, the reward will be also equal to 0. In this way, we attribute a maximum reward only when all requirements are fulfilled. Furthermore, to encourage the system to select the minimum number of inference contributors, we increment the reward for each redundant participant through: $\sigma \times \sum\limits_{j=1}^{D}A_t(j) \times \sum\limits_{l=1}^{t}A_l(j)$. In addition to the rewards gained for respecting different constraints, the agent is penalized for assigning segments to devices with low capacities. These penalties are also added to the reward function. Thus, in order to maximize the cumulative rewards, the RL-DistPrivacy agent have to minimize the penalties by selecting the optimal allocations. On these bases, the immediate reward $R_t$ is initialized to ($\frac{O^{l-1}_{i,j}}{\rho_i}+\frac{c^{l,p}_{r^*}}{e(j)}$), which is the delay of receiving the previous layer output and computing the current layer segment $p$ as indicated by the objective function (\ref{eq:optProb}). As described in section \ref{problem_formulation}, the inference latency is composed of two delays, transmission and computation. The transmission delay $\frac{O^{l-1}_{i,j}}{\rho_i}$ depends on the previous layer decisions, i.e. previous episode actions. Meaning, the capacity ${\rho_i}$ of previously chosen devices impacts the given penalty, even though it is not related to the current actions. Hence, to solve this issue, we added an immediate penalty $\beta$ of choosing less performant IoT devices, as it will affect the transmissions to next layers/episodes.
\subsubsection{Agent design}
The main objective of an RL agent is to learn the optimal policy $\Pi$: $S$ $\times$ $A$ $\rightarrow$ $[0,1]$, which is the strategy that results in maximum reward throughout different episodes. The optimal strategy is established by finding an approximation of the optimal action-value function $Q(S_t,A_t)$ known also as the future expected reward. Particularly, $Q(S_t,A_t)$ depicts  the goodness of selecting an action $A_t$  for a given state $S_t$ and it is depicted as follows:
\begin{equation}\label{eq:Q}
\begin{aligned}
Q(s,a)=E[\sum_{k=1}^{N}\gamma^k R_{t+k}|S_{t}=s,A_{t}=a],
\end{aligned}
\end{equation}
To remind, $\gamma \in [0,1]$ denotes the discount factor that serves to trade-off the weight of the immediate reward gained at each time step compared to the long-term reward obtained at the end of the episode. If $\gamma$ is close to 0, the agent will only consider the time step reward. Else, the agent will give the future rewards higher weights. To evaluate $Q(s,a)$, a temporal difference method is used:
\begin{equation}\label{eq:Q}
\begin{aligned}
Q(s,r) \leftarrow Q(s,r)+\alpha (R_{t}+\gamma \max_{a'} Q(s',a')-Q(s,a)),
\end{aligned}
\end{equation}
where $\alpha \in (0,1]$ presents the learning rate. Next, for each state, the action-value function $Q(s,r)$ is evaluated and stored in a Q-table. When $Q(s,r)$ is learned, the agent will act as follows:  
\begin{equation}\label{eq:argmax}
\begin{aligned}
\Pi(s)=arg max_a Q(s,r)
\end{aligned}
\end{equation}
\subsubsection{Deep Q-learning algorithm}
Our privacy-aware system for distributed inference is highly dynamic because of the varying number of neighboring IoT devices and the volatile requests' distribution. Moreover, the action space can be highly dimensional depending on the number of helpers joining the collaborative system. Therefore, it is challenging to use the traditional RL techniques based on storing all experienced Q-values in a Q-table. Thus, we opt to use DQN approach \cite{dqn} to approximate the optimal policy.
DQN is an algorithm that is able to learn the parametric representation of the action-value function $Q(s,a)$. The advantage of DQN approaches is that they are free of state transition knowledge. Meaning, instead of learning from the states model, DQN learns the optimal state–action values by interacting with the environment. The training of the DQN is done through minimizing the following loss function:
\begin{equation}\label{eq:Q}
\begin{aligned}
L(\theta)=E[(R{t}+\gamma \max_{a'}Q(s',a',\theta')-Q(s,a,\theta))^2].
\end{aligned}
\end{equation}
In this equation, $Q(s,r)$ is approximated to $Q(s,r,\theta)$, where $\theta$ presents the weights of deep neural network used for the DQN training.
To stabilize this training and enable the system to converge to the optimal learning, different steps should be followed, which are detailed in algorithm \ref{DQN}. The algorithm is initiated by creating two identical Q-Networks using the same NN weights (lines 2-3). Then, during the learning process, the system generates repeatedly the episodes of experiences ($S_t$,$A_t$,$R_t$,$S_{t+1}$) (lines 5-18) that are stored constantly in a buffer $D$ (lines 19-21). The decision taking process follows an $\epsilon-$ greedy algorithm (line 10). This algorithm allows the agent to take actions either randomly with a probability $\epsilon$ or by selecting the highest Q-value ($A_t=$arg max$Q(S_t,A_t)$) experienced until the current time step. More specifically, the RL agent starts by exploring the environment with a probability $\epsilon$ equal to 1, and then this probability decays over time to enable the system to use the past learned knowledge. At the end of the training process, although $\epsilon$ becomes small, it should not be equal to 0 to allow the system to take some random actions in order to discover environment changes and adapt to it. To improve the Q-values, the agent randomly samples at each time step a minibatch from the replay buffer D (line 22), and calculates the target values for each sample in the batch using the Q-network (line 24). This mechanism, called experience replay, contributes to reach the convergence of the system by learning from past experience. Moreover, to stabilize the learning, the agent keeps the target network fixed, while improving the main Q-network. Then, after each $G$ steps, this target network is updated towards the most recent learning presented by the main network (line 27) \cite{dqn}.
\begin{algorithm}[h]
\caption{RL-DistPrivacy}
\label{DQN}
\begin{algorithmic}[1]
\State \textbf{DQN initialization:}
\State Initialize the Q-network and the target network with the  
\State same random parameters $\theta$ and $\theta'$, where $\theta' \leftarrow \theta$.
\State \textbf{DQN Learning:}
\FOR{each $r\in RQ$}
\FOR{each episode $l^{k}_{r^*}\in L_{r^*}$}
\FOR{each time-step $t=1..p_{l^{k}_{r^*}}$}
\State $S_t=\{r^*, l^{k}_{r^*}, p_{l^{k}_{r^*}}, \{\bar{c_1},...,\bar{c_D}\}, , \{\bar{b_1},...,\bar{b_D}\},$
\State $\{\bar{m_1},...,\bar{m_D}\}, \{\bar{L_1},...,\bar{L_D}\},\{A_1,...,A_{P_{l^{k-1}_{r^*}}}\} \}$
\State Select $A_t$ based on $\epsilon-$greedy policy
\IF{$\sum A_t=1$}
\FOR{$j=1:D$}
\IF {$t>1$ and $t<L_{r^*}$}
\State $R_t=R_t -A_t(j)\times(\frac{O^{k-1}_{i,j}}{\rho_i}\times \mathrm{1}_{I_i\neq I_j}+$
\State $c_{p,k}^k/e(j))+\beta_j$
\IF {$C_1 \times C_2 \times C_3=1$}
\State $R_t=R_t+max(1,\sigma \times \sum\limits_{j=1}^{D}A_t(j)$
\State  $\times\sum\limits_{l=1}^{t}A_l(j))$
\ENDIF
\ENDIF
\ENDFOR
\ENDIF
\State (1) Receive $R_t$ and observe the next state $S_{t+1}$.
\State (2) Store $(S_{t},A_t,R_t,S_{t+1})$  in the replay
\State  memory $D$.
\State  (3) Sample a random minibatch of 
\State $(S(j),A(j),R(j),S(j+1))$ from  $D$.
\State (4) Calculate the target Q-value $Q_{target}(j)$ using   
\State the target Q-network: $Q_{target}(j)=R(j)+$
\State $\gamma max_{a'} Q(s',a',\theta')$.
\State (5) Every $G$ steps, use the loss function:  
\State  $L(\theta)=[Q_{target}(j)-Q(s,a,\theta)]^2$ 
\State to update the target $Q$-network.
\ENDFOR
\ENDFOR
\ENDFOR
			\end{algorithmic}
\end{algorithm}
\section{Performance evaluation}\label{simulation}
In this section, we present the performance of our privacy-aware system for low-latency and distributed CNN, under different networking configurations. Specifically, we study the impact of the requests' type, the privacy level, and the number of devices and their capacities on the total latency and the shred data between participants. Then, to prove the performance of our novel RL-DistPrivacy system, we compared it to the optimal framework, our heuristic-based solution \cite{Emna_globecom} and a state-of-art CNN distribution strategy \cite{dis5}. These simulation results are presented after introducing the framework settings.
\subsection{Framework settings}
Our DistPrivacy approach has been validated on four state-of-the-art benchmark CNNs, as described in section \ref{emperical_results}, in a surveillance scenario for image classification to control a critical area. Following the characteristics of the datasets used in our empirical study, we suppose that the area is surveilled by 10 devices (e.g., cameras) capturing 36x36 RGB  images (CIFAR), 28x28 gray images (MNIST), and 128x128 RGB images (CELEBA and Stanford CAR). This area is also composed of resource-limited IoT devices, where two technological families are evaluated in this paper. The first one is Raspberry Pi B+ (RPi3) equipped with 1.4 GHz 64-bit quad-core processor and 1 GB RAM while the second is LG Nexus supplied with a more powerful 2.28 GHz processor and a higher memory availability equal to 2 GB RAM. The number of multiplications per second $e$, defined as the tenth of the clock cycles per number of cores \cite{dis5}, is equal to 560 for RPi3, and 800 for LG Nexus. Finally, we suppose that both technologies are endowed with an IEEE 802.11n standard having an average data rate $\rho$ equal to 72.2 Mb/s. We note also that the 10 image-capturing devices belong to RPi3 technology. Moreover, the cooperative system is composed of 70 IoT devices, where 20 of them are LG Nexus and 50 are RPi3.

The   proposed   system   is   evaluated,   first,   on  decision taking  latency, which is  defined  as  the  delay  between acquiring  the  image  and obtaining the classification.  Second,  we evaluate  the  data  shared  between all  participants  to  accomplish  all  CNN tasks. The privacy robustness of our approach is applied on 4 types of classifications. i.e., all requests are Mnist-LeNet, all requests are CIFAR-CNN, all requests are VGG, and heterogeneous requests, and  3 privacy levels equal to 0.8, 0.6, and 0.4. As discussed previously, the privacy level is presented  by the SSIM metric (see table I). In other words, a level equal to 0.8 means a SSIM similarity equal to 0.8. Such similarity level implies low data blurring; hence, we call it low  privacy level. When the SSIM is equal to 0.4, we consider the privacy level high. We remind that to respect these privacy requirements, the maximum number of feature maps per device set for each SSIM, is added as a constraint to both optimization and RL system. Furthermore, if we fix the SSIM level to be 0.4, the similarity of the recovered image should be equal or less than 0.4 compared to the original sample, and this constraint should be respected throughout the simulation to guarantee the privacy of the inference.
The RL-DistPrivacy algorithm is validated according to the parameters shown in Tables \ref{param}  and \ref{param2}. These parameters are empirically chosen and we expect that similar architectures perform identically.

\begin{table}[!h]
\caption {Parameters of the RL simulation.}
\label{param}
\centering
\footnotesize
\begin{tabular}{|l|l|l|}
\hline
Parameter & Description & Value \\ \hline
$\gamma$ & discount factor & 0.95 \\ \hline
$\epsilon$ & exploration rate & 1 \\ \hline
$bz$ & buffer size & 50000 \\ \hline
$M$ & batch size & 64 \\ \hline
\end{tabular}
\end{table}

\begin{table}[!h]
\caption {Parameters of the RL simulation for each network type.}
\label{param2}
\centering
\footnotesize
\tabcolsep=0.05cm
\begin{tabular}{|c|l|c|c|c|c|c|c|c|c|c|c|l|l|}
\hline
Parameter & Description & \multicolumn{3}{c|}{LeNet} & \multicolumn{3}{c|}{CIFAR-CNN} & \multicolumn{3}{c|}{VGG } & \multicolumn{3}{c|}{All} \\ \hline
SSIM & \begin{tabular}[c]{@{}l@{}}Privacy \\ level\end{tabular} & 0.8 & 0.6 & 0.4 & 0.8 & 0.6 & 0.4 & 0.8 & 0.6 & 0.4 & 0.8 & 0.6 & 0.4 \\ \hline
$\epsilon_{decay}$ & decay & \multicolumn{3}{c|}{0.995} & \multicolumn{3}{c|}{0.995} & \multicolumn{3}{c|}{0.995} & \multicolumn{3}{c|}{0.9995} \\ \hline
G & \begin{tabular}[c]{@{}l@{}}update \\ frequency\end{tabular} & \multicolumn{3}{c|}{100} & \multicolumn{3}{c|}{3000} & \multicolumn{3}{c|}{3000} & \multicolumn{3}{c|}{3000} \\ \hline
$\alpha$ & \begin{tabular}[c]{@{}l@{}}learning \\ rate\end{tabular} & \multicolumn{3}{c|}{0.0001} & \multicolumn{2}{c|}{0.001} & 0.0001 & \multicolumn{3}{c|}{0.0001} & \multicolumn{3}{c|}{0.00001} \\ \hline
$\sigma$ & \begin{tabular}[c]{@{}l@{}}Reward \\ parameter\end{tabular} & \multicolumn{3}{c|}{1} & \multicolumn{2}{c|}{1} & 0 & \multicolumn{3}{c|}{0} & \multicolumn{3}{c|}{0} \\ \hline
$\beta$ & penalty & \multicolumn{3}{c|}{0.5} & \multicolumn{3}{c|}{0.5} & \multicolumn{3}{c|}{0} & \multicolumn{3}{c|}{0} \\ \hline
\end{tabular}
\end{table}
 All the simulations are conducted on a computer, having the following characteristics:  core i7 and 16 GB RAM. More specifically, the RL framework is developed using python language and stable baseline library\footnote{https://stable-baselines.readthedocs.io/en/master/} and the optimization is tested on Matlab.
\subsection{Simulation results}
\subsubsection{Performance of the RL convergence}
\begin{figure}[h]
	\mbox{
		\subfigure[\label{lenet1}]{\includegraphics[scale=0.38]{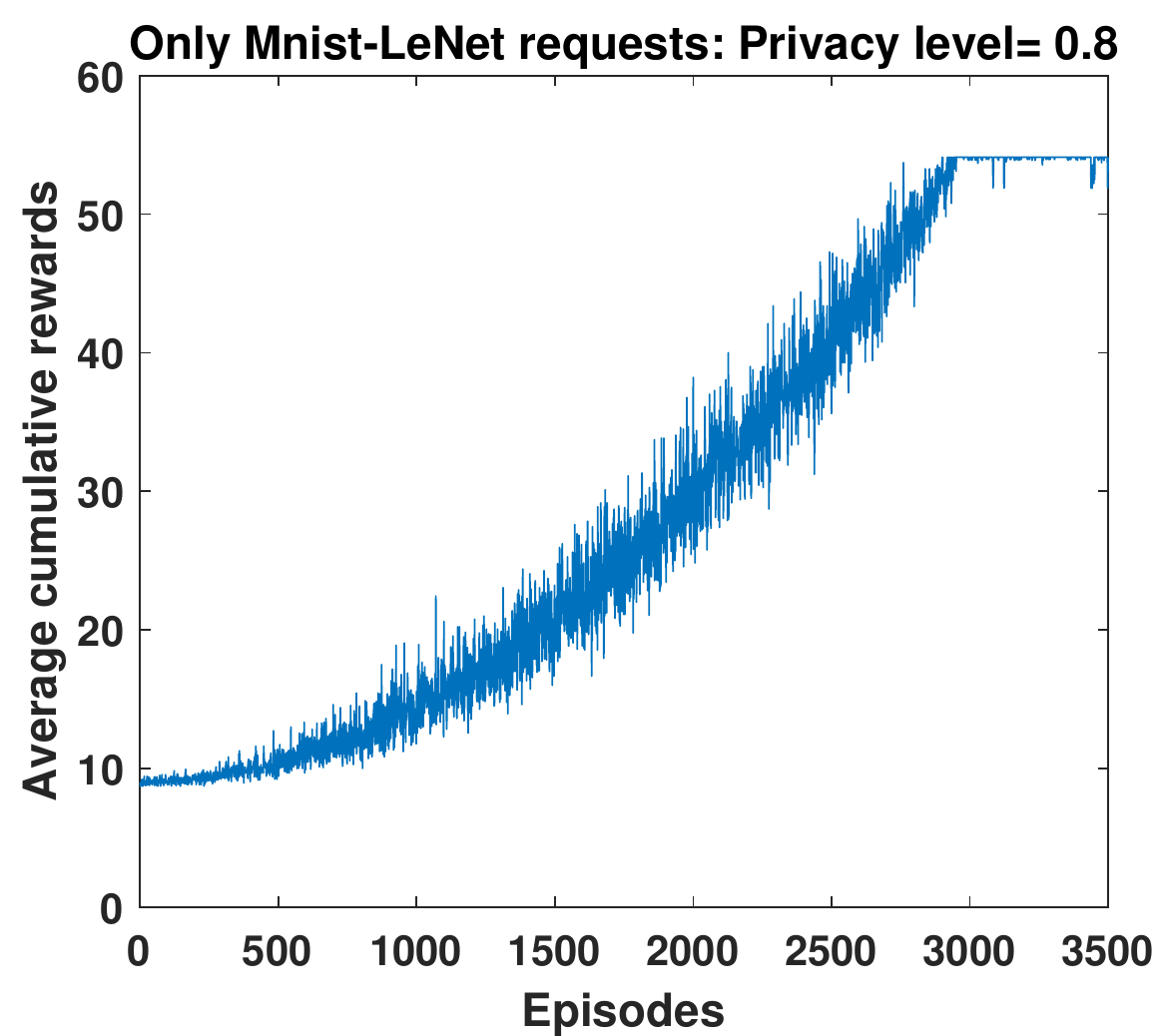}}
		\subfigure[\label{lenet2}]{\includegraphics[scale=0.38]{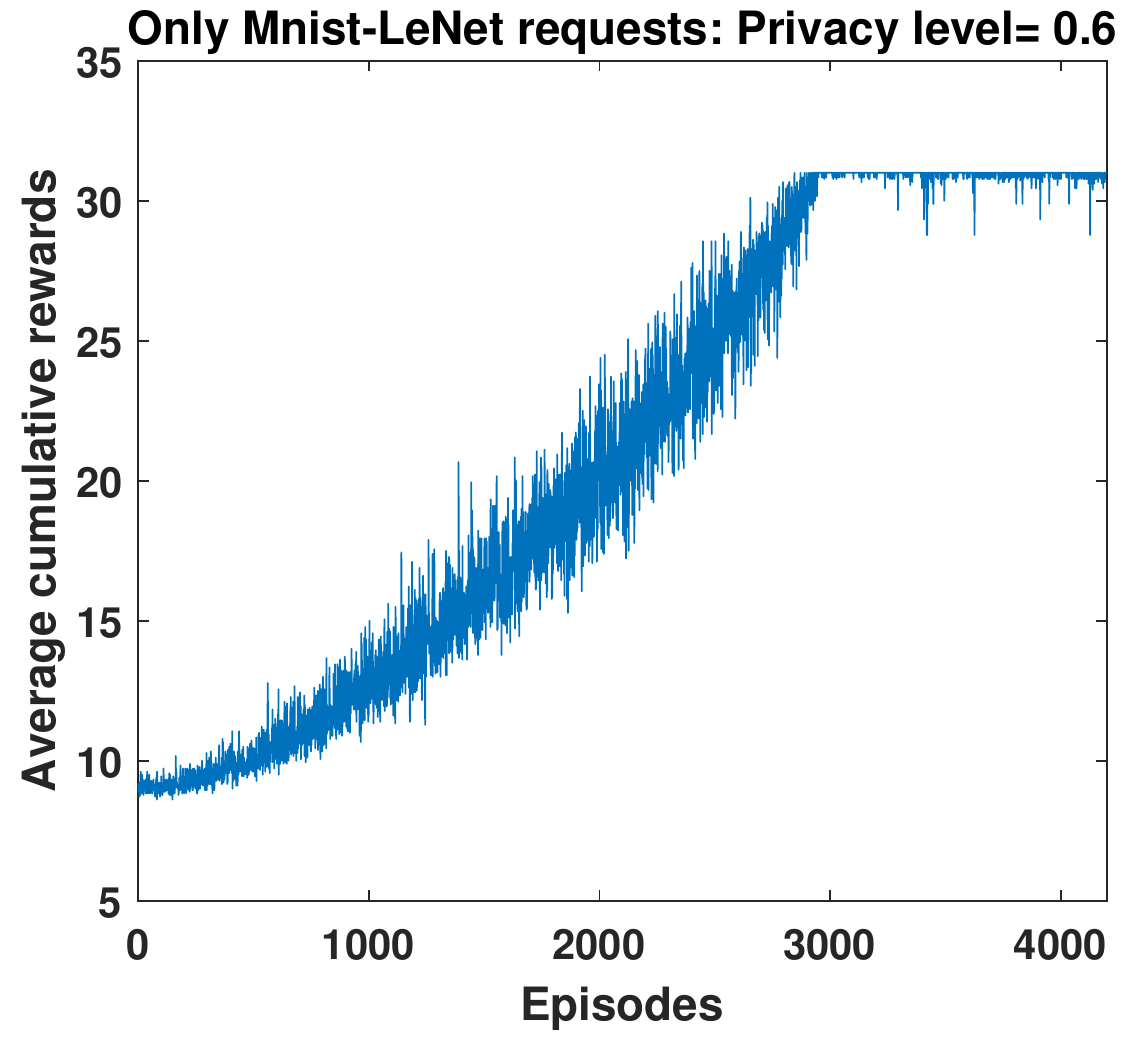}}}\\
	\mbox{	\subfigure[\label{cifar1}]{\includegraphics[scale=0.38]{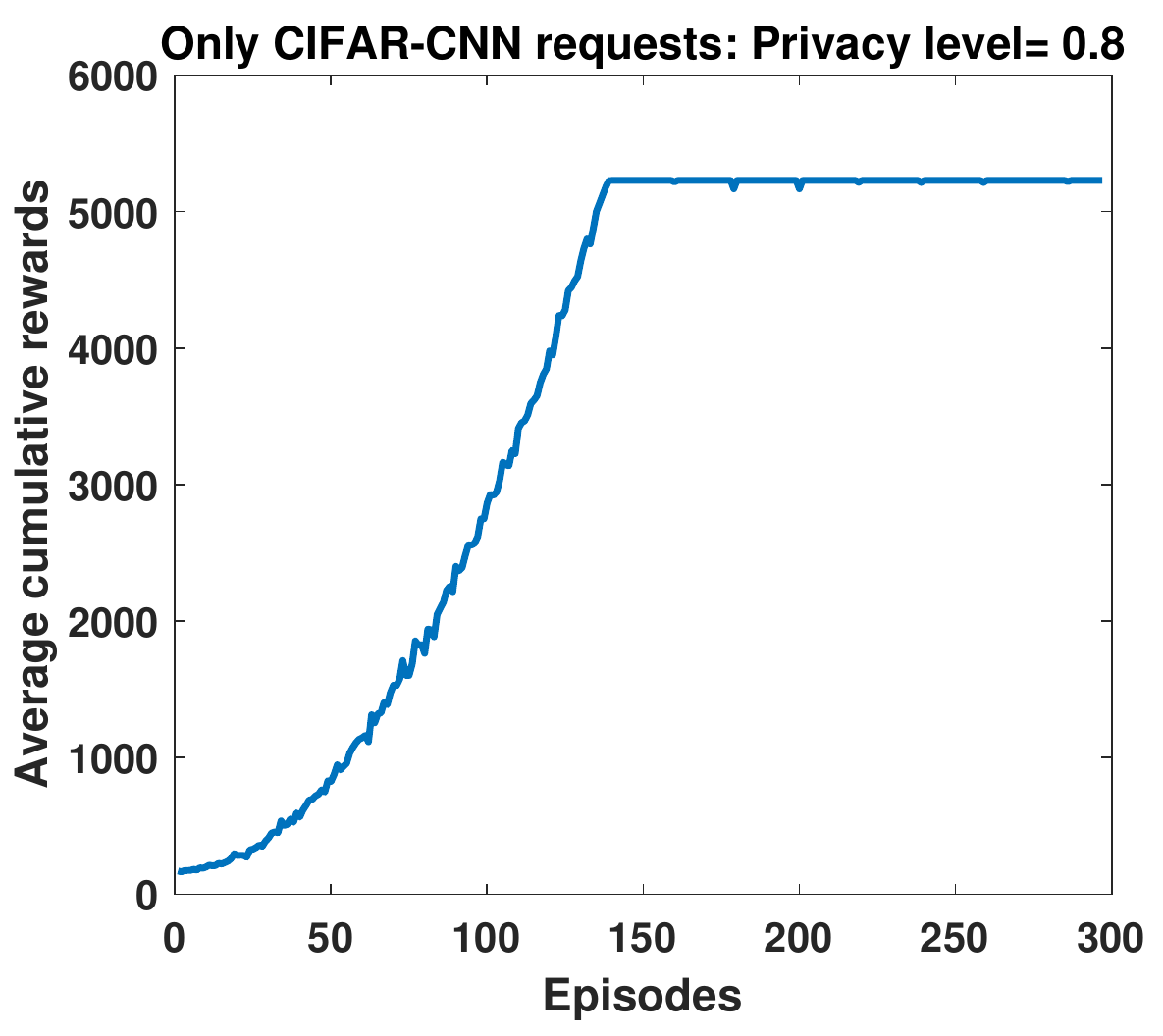}}
		\subfigure[\label{cifar2}]{\includegraphics[scale=0.38]{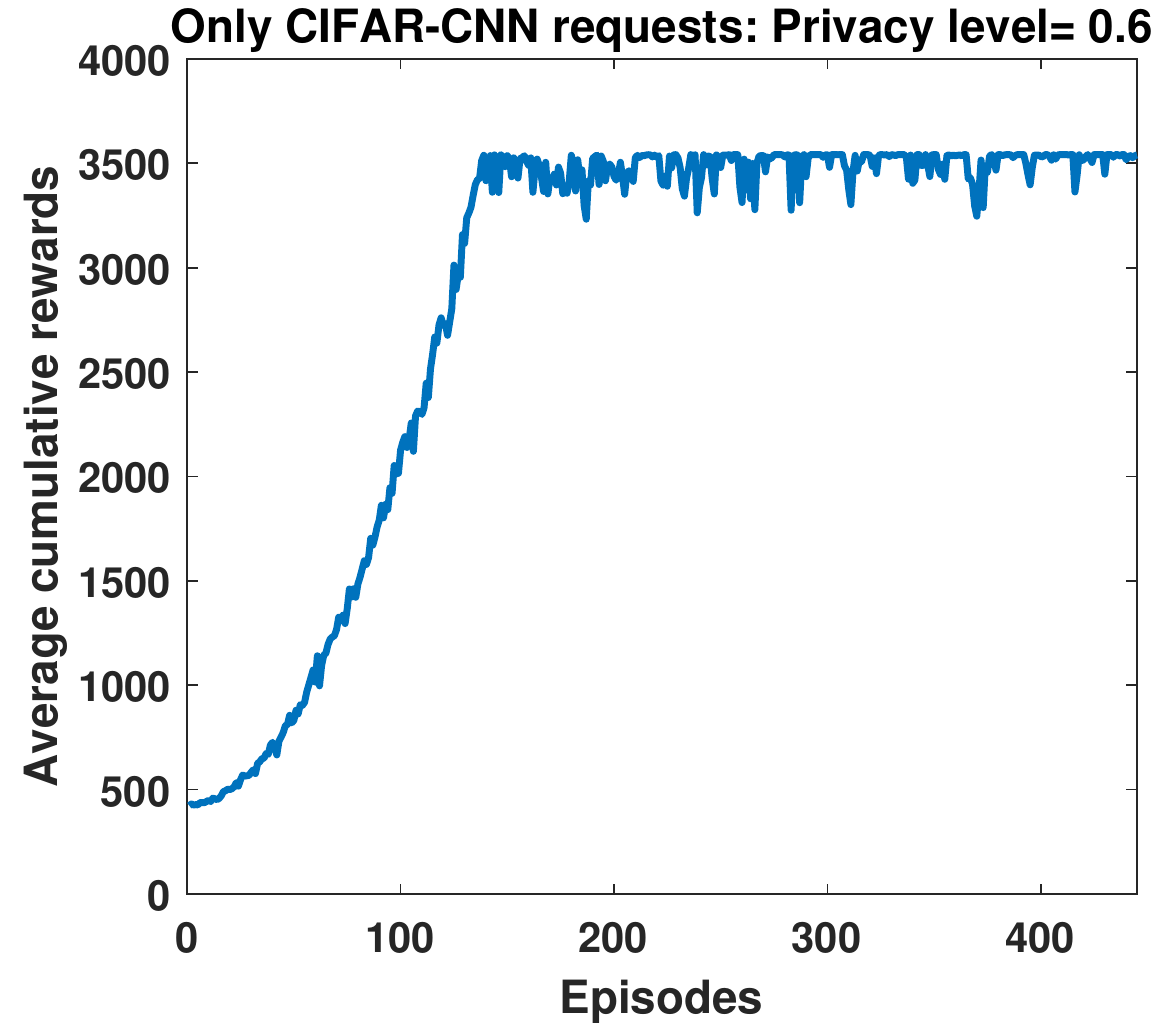}}}\\
	\mbox{	\subfigure[\label{VGG0}]{\includegraphics[scale=0.38]{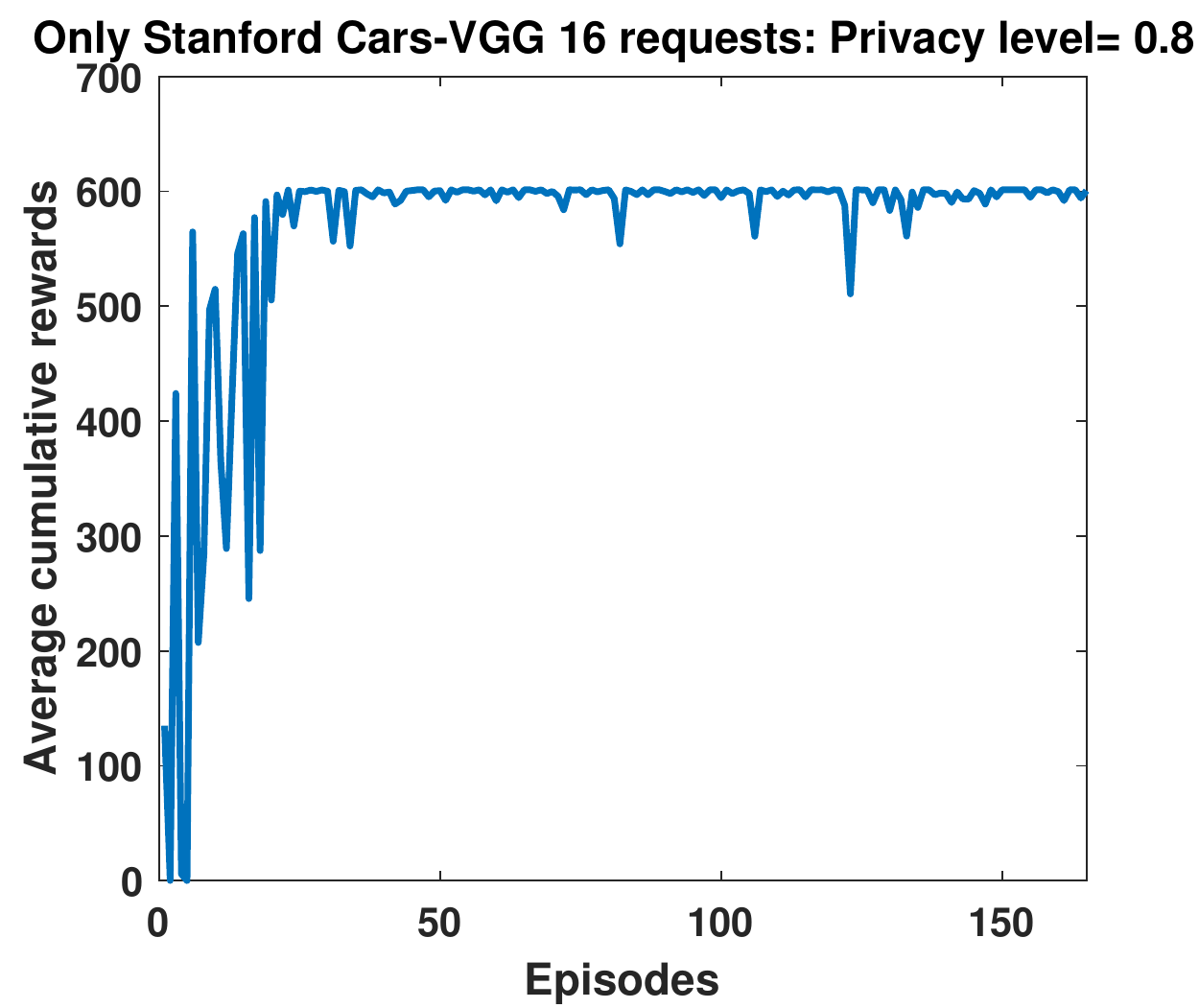}}
        \subfigure[\label{VGG1}]{\includegraphics[scale=0.38]{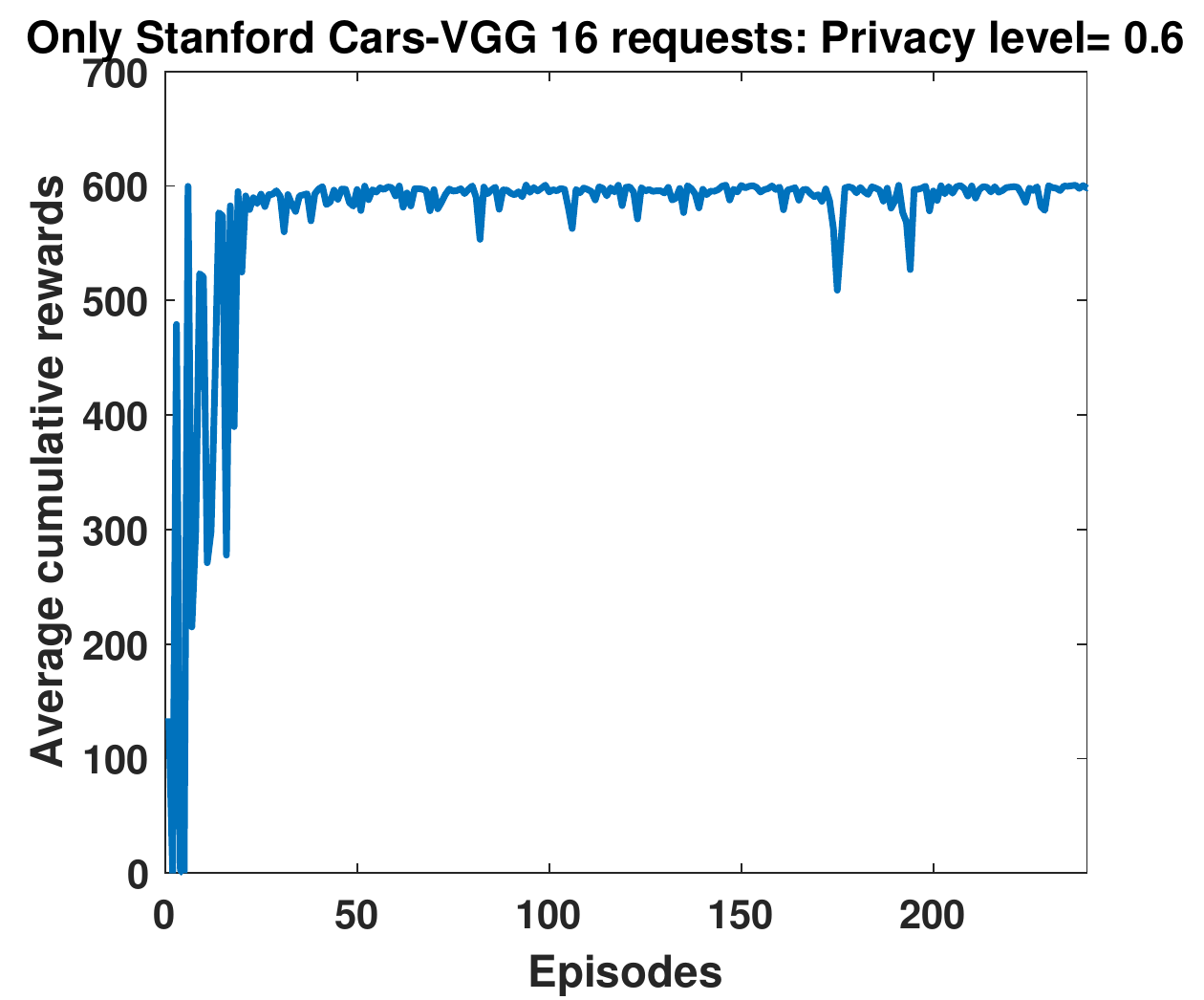}}
        }
	\caption{Average cumulative rewards vs. training episodes: LeNet, CIFAR, VGG requests.}
	\label{convergence1}
\end{figure}

\begin{figure*}[h]
	\centering
	\mbox{
		\subfigure[\label{all0}]{\includegraphics[scale=0.36]{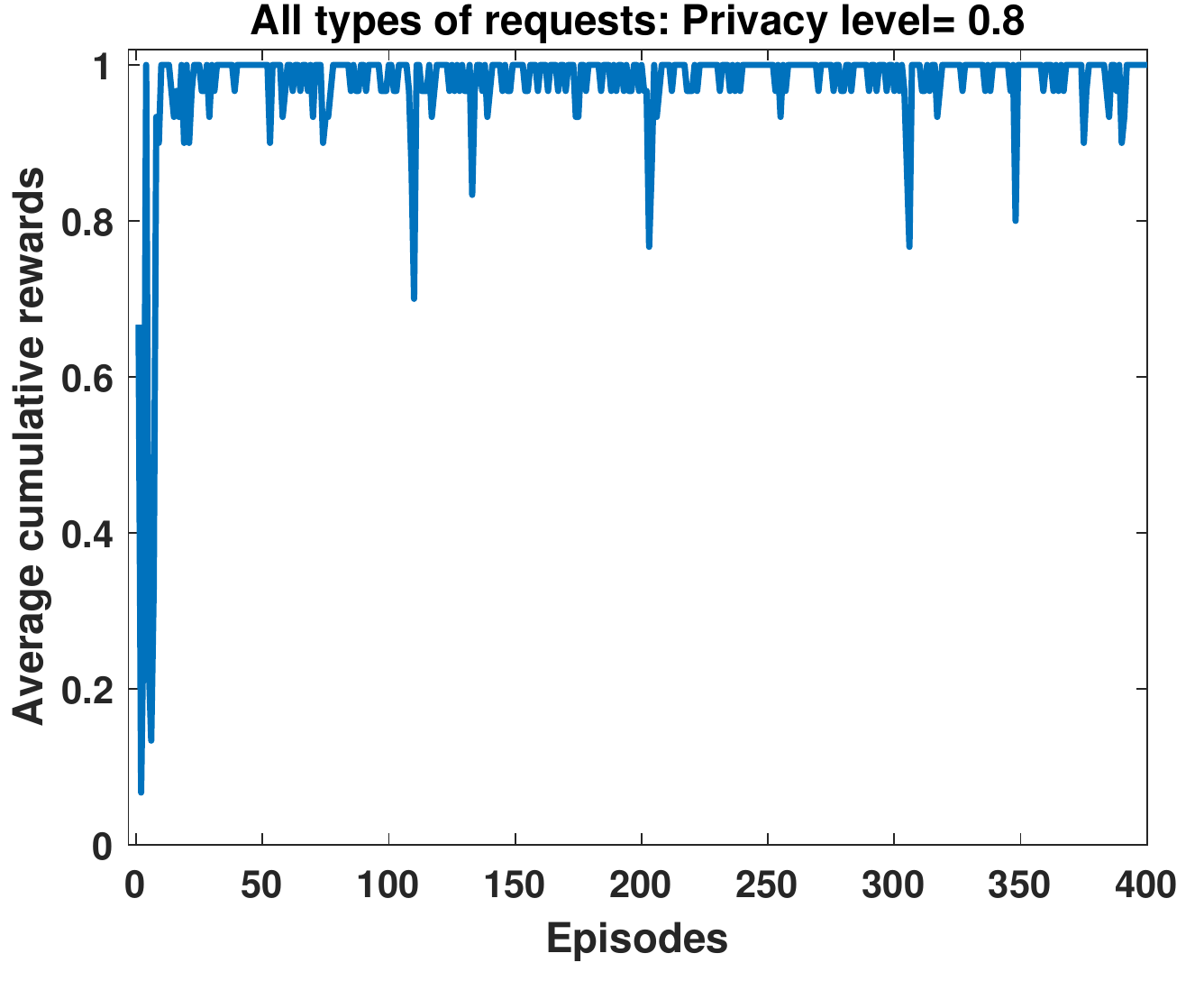}}
        \subfigure[\label{all1}]{\includegraphics[scale=0.37]{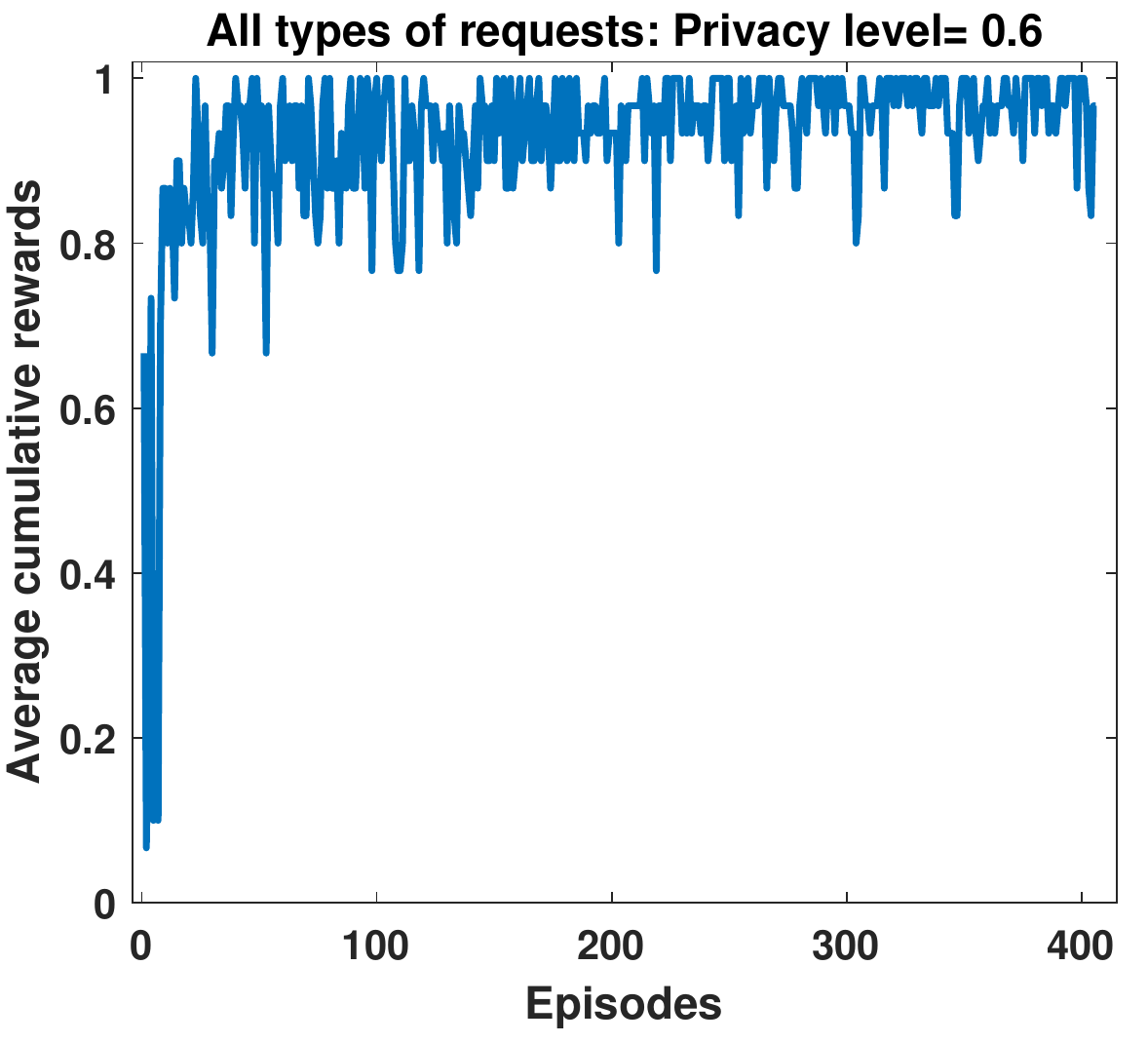}}
		\subfigure[\label{all2}]{\includegraphics[scale=0.4]{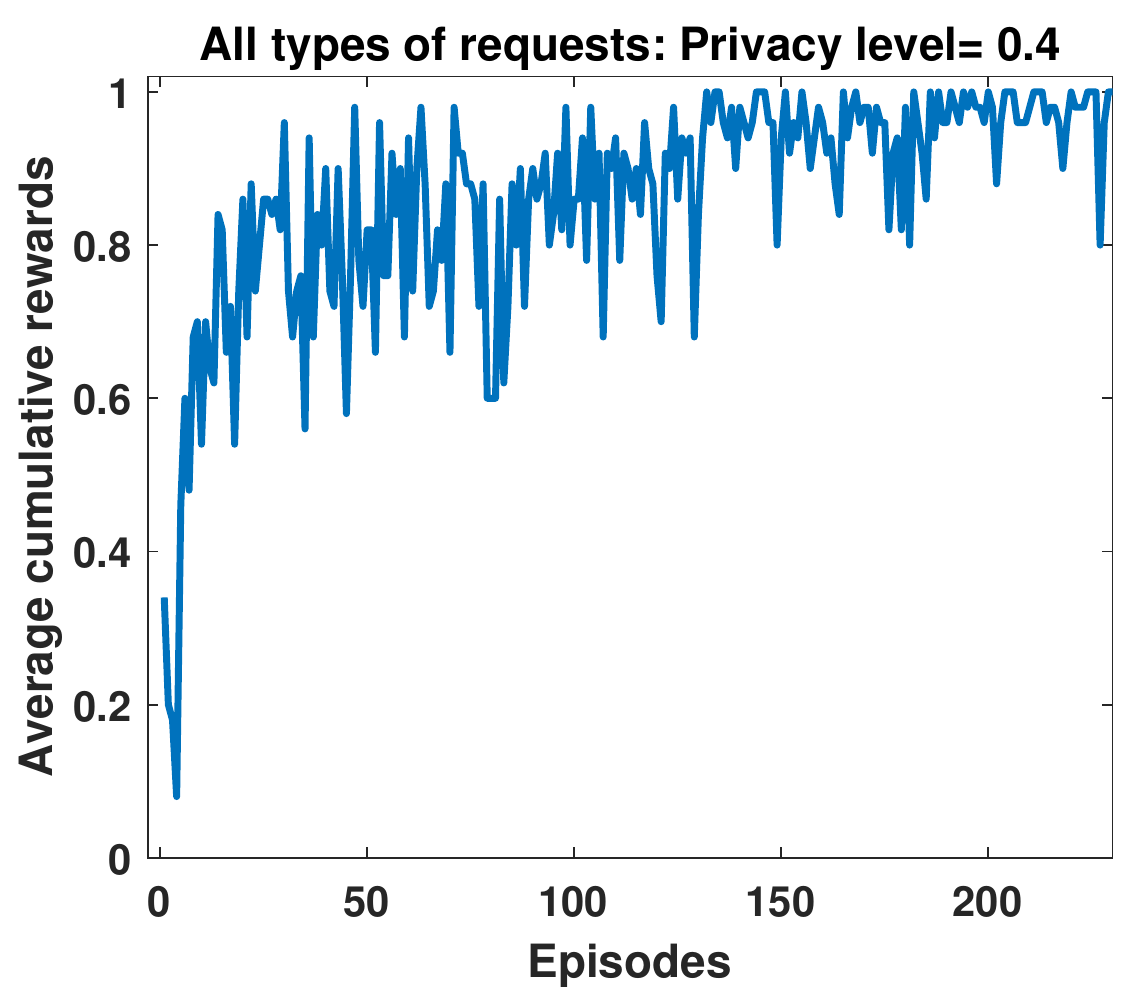}}}
	\caption{ Average cumulative rewards vs. training episodes: heterogeneous requests.}
	\label{convergence2}
\end{figure*}

\begin{figure*}[h]
	\centering
	\mbox{
		\subfigure[\label{cost0}]{\includegraphics[scale=0.5]{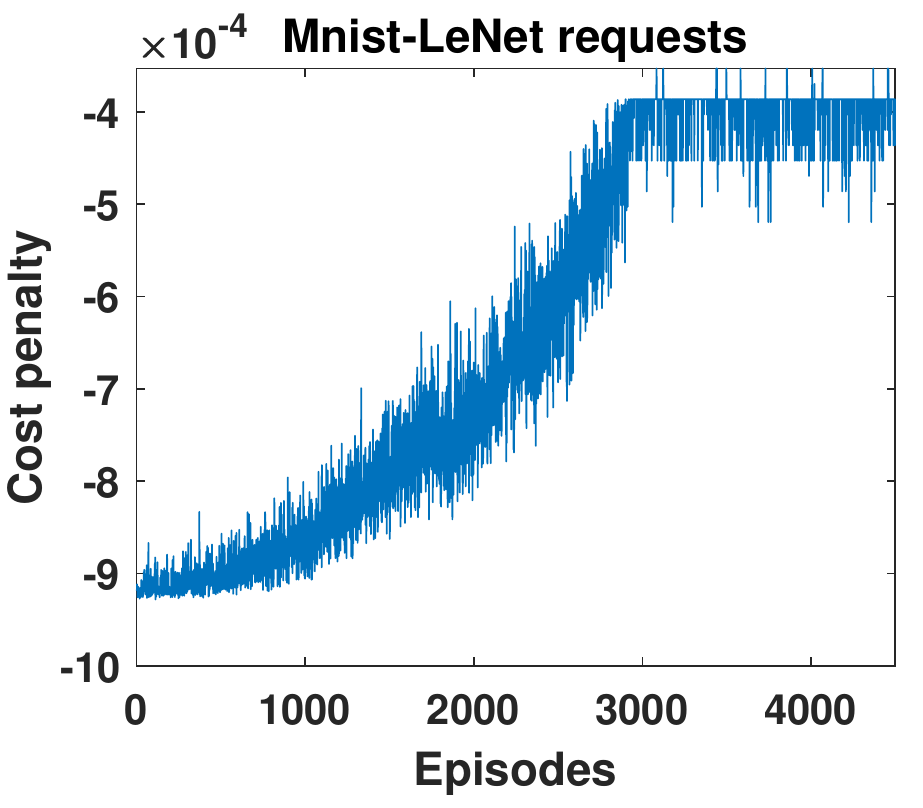}}
        \subfigure[\label{cost1}]{\includegraphics[scale=0.5]{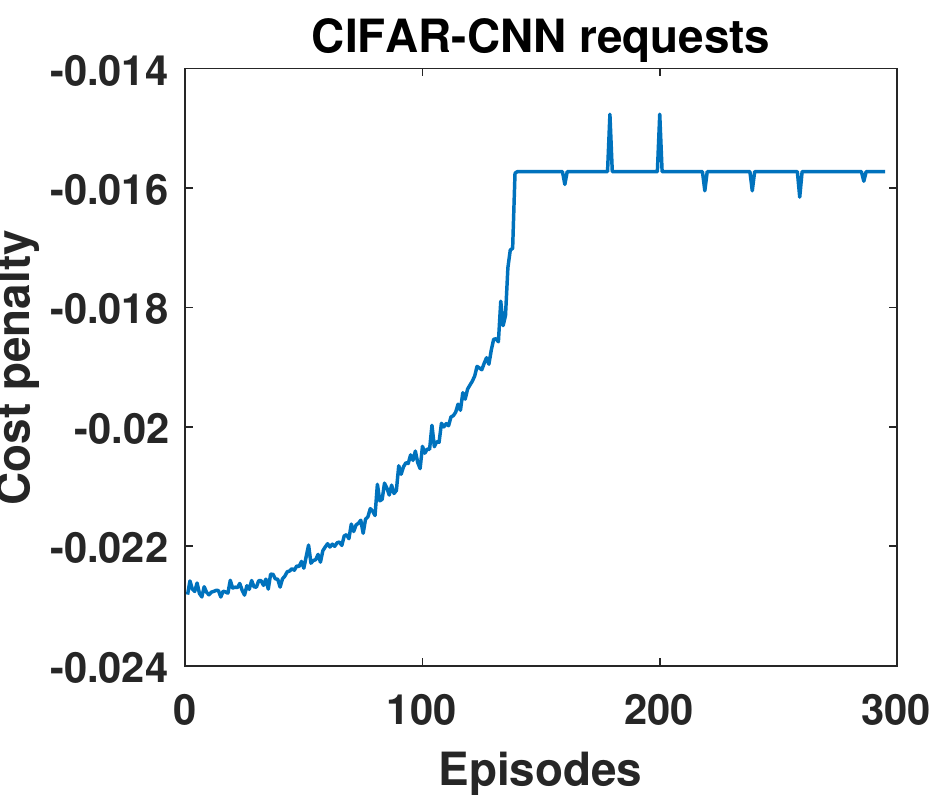}}
		\subfigure[\label{cost2}]{\includegraphics[scale=0.5]{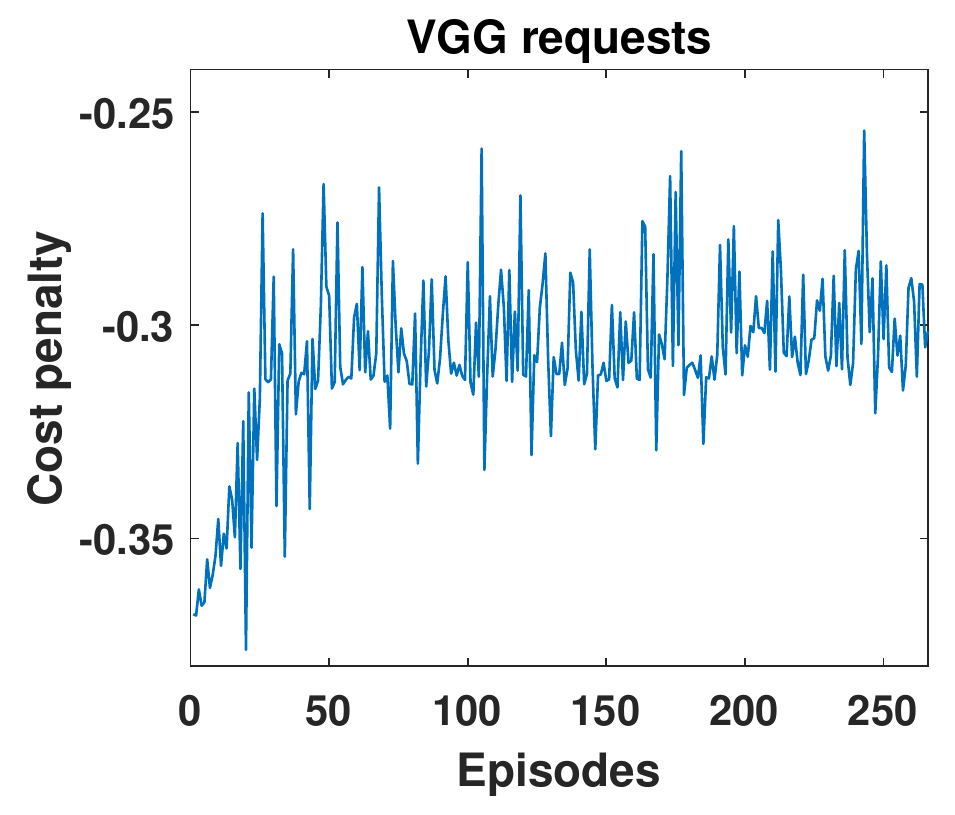}}}
	\caption{ Average cumulative cost penalty vs. training episodes: LeNet, CIFAR-CNN and VGG requests.}
	\label{convergence3}
\end{figure*}

Figures \ref{convergence1} and \ref{convergence2} illustrate the variation of cumulative rewards among training episodes. The obtained rewards are smoothed over a window of 34 (CIFAR CNN), 18 (LeNet), 66 (VGG 16), and 100 (heterogeneous requests) depending on the trained network. To have an initial estimate of the policy, we set $\epsilon$ to be equal to 1 in the first 1000 episodes, which means all actions are taken randomly. Next, a decay parameter, namely $\epsilon_{decay}$ is added throughout the episodes to slowly reduce the number of random actions and enable the system to smoothly pass to an enhance policy. On these bases, we can see that, at the beginning of the learning process, not all constraints are respected. However, as the system experiences new episodes and new scenarios, it starts to learn how to respect constraints, until reaching a smooth stability, which confirms the convergence performance of our RL model for different types of networks. Without loss of generality, we only present the convergence of the networks (LeNet, CIFAR-CNN, and VGG) respecting the two first privacy levels and we illustrate only VGG 16 network as VGG19 demonstrates the same performance.
Furthermore, we can see that the convergence of the LeNet network  (Figures \ref{lenet1} and \ref{lenet2}) takes longer compared to the other networks' types. This slow convergence is due to the small number of layers and segments, which requires a higher number of episodes to learn how to respect security constraints while involving the minimum number of participants. Similarly, VGG  network  (Figures \ref{VGG0} and \ref{VGG1}) converges faster than the CIFAR-CNN, as it has ten times the number of segments compared to the latter one. Regarding the CIFAR-CNN  (Figures \ref{cifar1} and \ref{cifar2}), we can observe that the convergence of the lower privacy level is smoother than the others. This can be explained by the fact that we added a penalty $\beta$ on choosing the less performant devices. Hence, since lower security requirements can be satisfied using only two participants, the incoming requests are handled by high-performing devices, without resorting to less performant ones. When the privacy level is higher, a larger number of IoT participants is needed to satisfy the requirements. Hence, devices with low capacities should contribute in the CNN  distribution, which implies some penalties. Finally, when the network is heterogeneous  (Figures \ref{all0}, \ref{all1} and \ref{all2}), we can see that the convergence is not too smooth, which is related to the rejections that can occur due to the non-availability of IoT resources.  In Figure \ref{all0}, the convergence is rapid and the performance is high due to the low privacy requirements that do not involve much participants and resources. As the privacy level increases (Figures \ref{all1} and \ref{all2}) and the number of required participants and resources is larger, the  convergence to reach the optimal policy becomes slower and the rejections become higher. We note that, in Figure \ref{convergence2}, we did not present the actual cumulative rewards, as the distribution of requests' types is random and the episodes' lengths are not equal. Instead, we attribute 1 if the episode fulfills the maximum cumulative rewards and 0 otherwise. Figure \ref{convergence3} illustrates the convergence performance of the cost penalty that is added to the reward function in order to learn how to minimize the latency of the inference. We can see that, after converging, the latency of VGG  (Figures \ref{cost2}) is highly oscillating, which is explained by the fact that this model is resource-consuming and the processing delay highly depends on the available devices and the requests' load. This is not the case of LeNet  (Figures \ref{cost0}) and CIFAR-CNN  (Figures \ref{cost1}) that can assign the computation to devices with the highest capacities, at most of the time.  Hence, it can be seen that the convergence is smooth and the penalty is always akin to the same range.
\begin{figure}[!h]
\centering
	\includegraphics[scale=0.47]{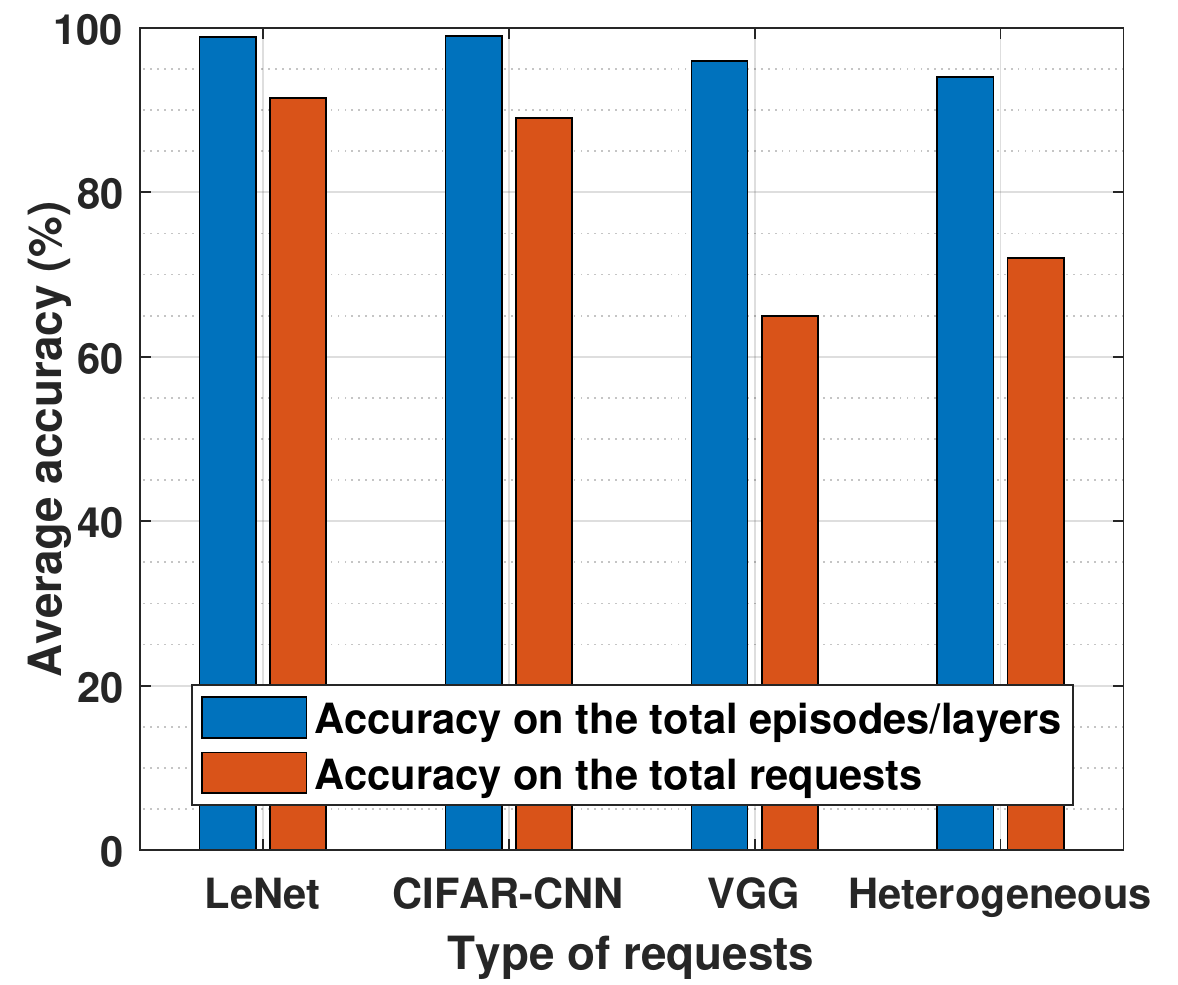}
	\caption{Accuracy of the system in terms of meeting constraints.}
	\label{accuracy}
\end{figure}

Figure \ref{accuracy} depicts the average accuracy (over different privacy levels) of different types of networks. We define the accuracy of our RL-DistPrivacy system as the ability to respect different constraints after convergence. It can be seen that our system achieves a high accuracy reaching more than 90\% for all types of CNNs, which means that all resources and privacy constraints are respected in more than 90\% of the episodes. In our design, an episode presents one layer of an inference. Hence, in terms of RL-accuracy of the complete classification requests, the studied CNN networks show different performances. More specifically, Mnist-LeNet and CIFAR-CNN requests are classified with an accuracy equal to 90\%, where all privacy requirements and available IoT resources are considered. When the requests are heterogeneous, the system achieves on average among different privacy levels, an accuracy equal to 72 \%. When the classification of the captured data is done using VGG CNN, only 65\% of the inferences are distributed while meeting all the requirements. Such low RL-accuracy is caused by the complexity and resource requirements of VGG requests that cannot be always allocated by the available IoT participants due to their limited capacities. In this case, the system either decides to exploit all the existing resources without considering the privacy constraints or assign the CNN tasks to unavailable devices while respecting the privacy level.  
\begin{figure}[h]
	\centering
	\mbox{
	\hspace{-2.7mm}
		\subfigure[\label{RL_300}]{\includegraphics[scale=0.4]{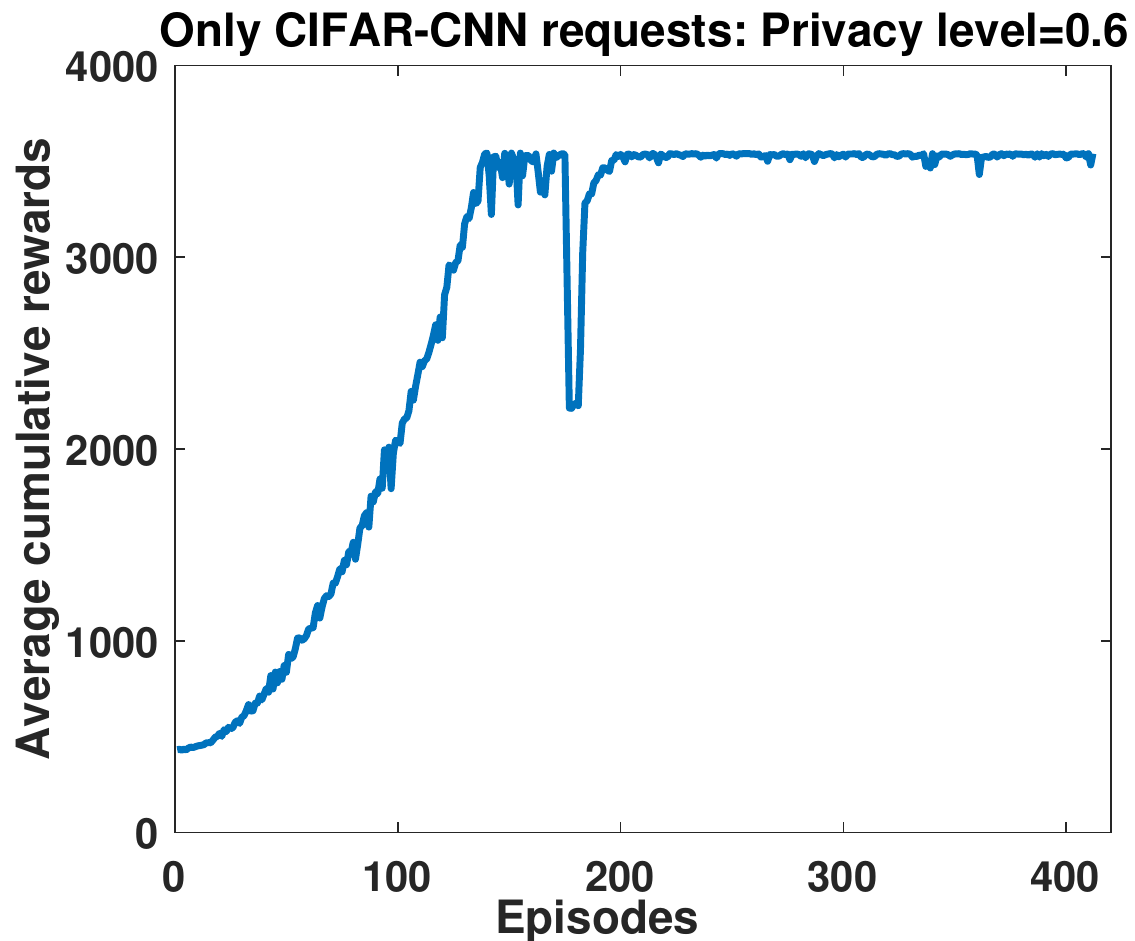}}
		\hspace{-2.3mm}
        \subfigure[\label{RL_500}]{\includegraphics[scale=0.4]{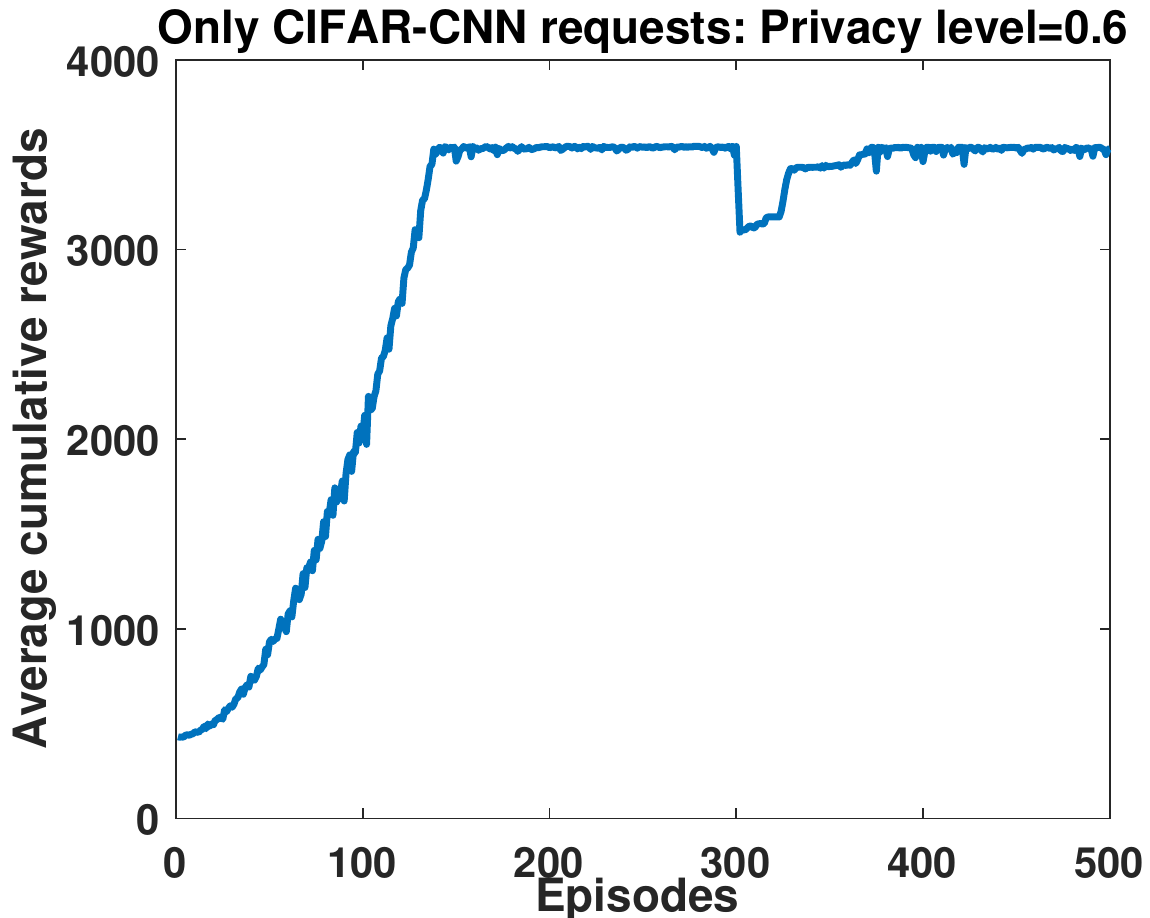}}
}
	\caption{ Cumulative rewards averaged  over 34 episodes: CIFAR-CNN requests distributed to 20 participants where 10 of them leave after (a) 5000 episodes, (b) 10000 episodes.}
	\label{RL_review}
\end{figure}

Figure \ref{RL_review} presents the average cumulative rewards of a network composed of 20 participants computing CIFAR-CNN requests. After an interval of training episodes, 10 devices leave the system. We can see that, when the participants abandon the collaborative system after 5000 episodes (Figure \ref{RL_300}), the rewards drop drastically. On the other hand, when the system changes after 10000 training episodes, the reward's decrease is less noticed. However, in both cases, the RL-agent re-gains rapidly the convergence and learns again to respect the constraints relying only on the remaining devices. This prompt re-convergence proves that the RL-DistPrivacy manages the environment dynamics and changes easily using its past learning. Moreover, the RL performs better, when the system changes after a longer learning and stability phase.

\subsubsection{Comparison to state-of-the-art approaches}
\begin{figure}[!h]
\centering
\hspace{-6mm}
	\includegraphics[scale=0.55]{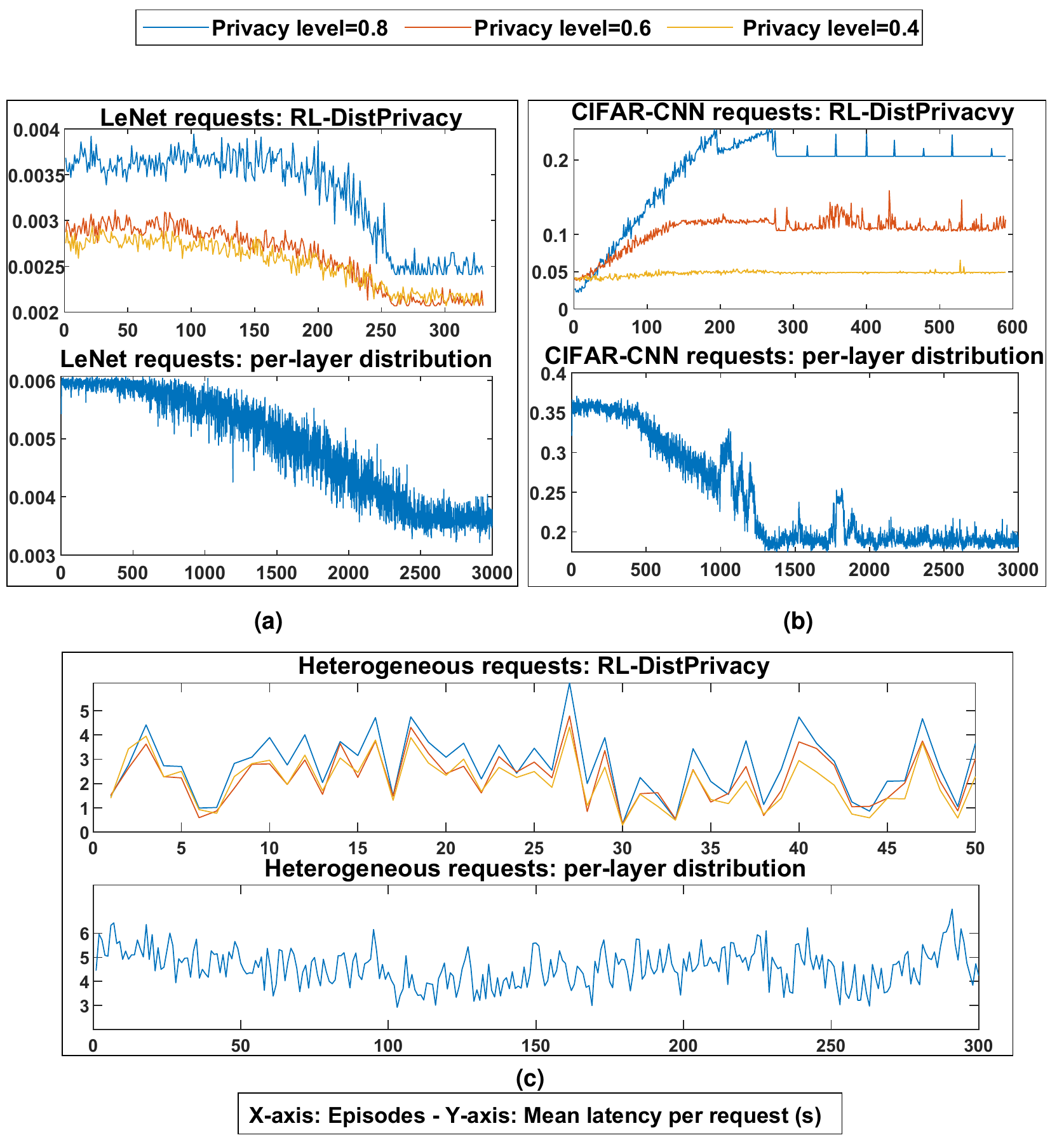}
	\caption{Latency per request: RL-DistPrivacy vs per-layer distribution.}
	\label{heuristic_latency}
\end{figure}
\begin{figure}[!h]
\centering
	\includegraphics[scale=0.55]{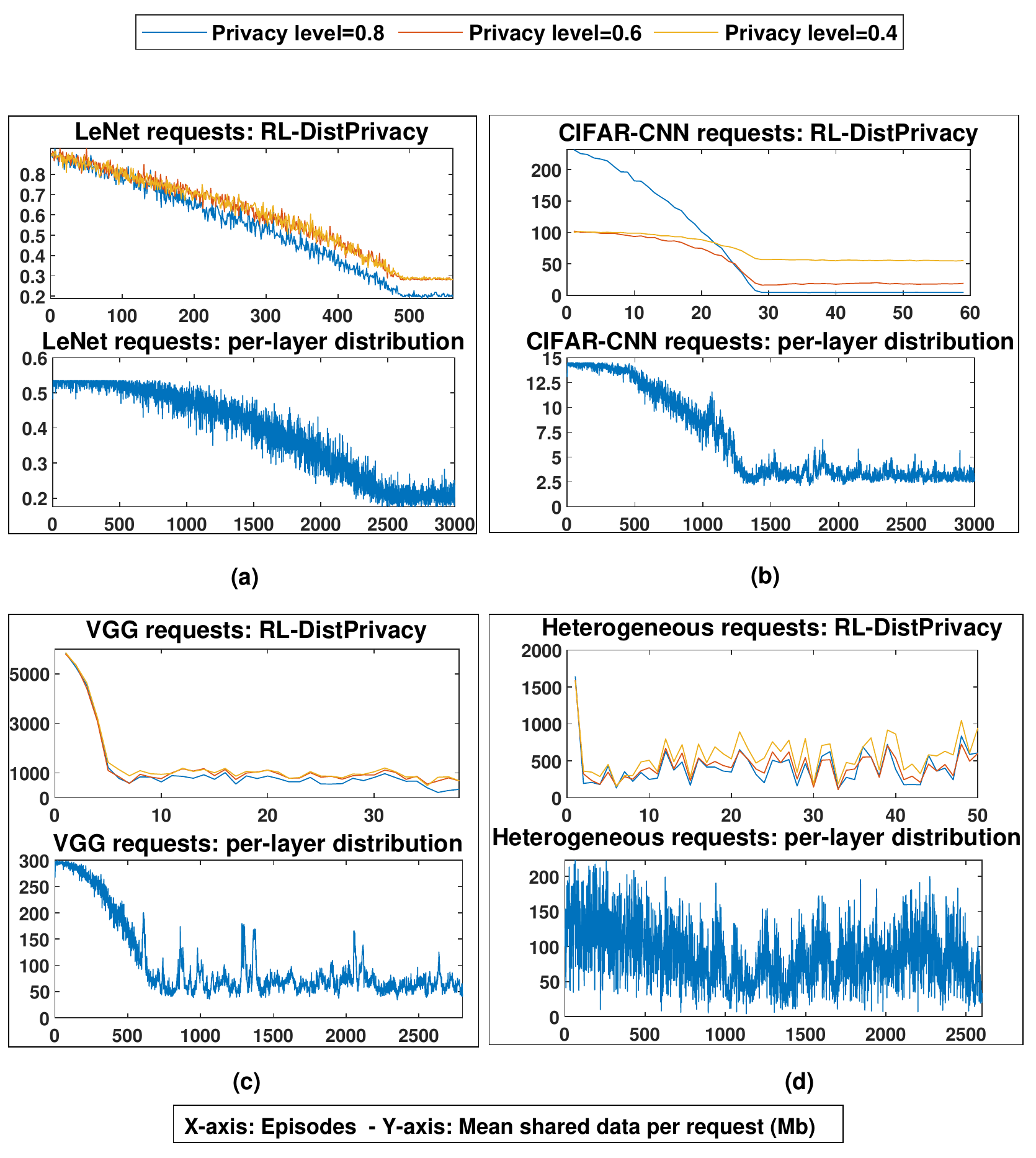}
	\caption{Shared data per request: RL-DistPrivacy vs per-layer distribution.}
	\label{heuristic_data}
\end{figure}
Figures \ref{heuristic_latency} and \ref{heuristic_data} present the performance of our CNN distribution approach against per-layer distribution systems, including \cite{dis5}.  We note that existing per-layer distribution approaches do not handle any privacy concerns and focus only on CNN partitioning and that authors in \cite{dis5} presented their work as an optimization problem giving optimal results. Hence, we developed their approach as an RL design for a fair comparison. Furthermore, the goal from this comparison is not to show the efficiency of our framework from privacy perspective. Instead, we aim first to show the capability of our approach to minimize the inference latency with some compromises. In these Figures, we can notice that when the privacy level is equal to 0.8 (lower level), the performances of our distribution approach and per-layer strategy are similar in terms of latency and shared data, for small CNNs, i.e., LeNet in Figures \ref{heuristic_latency}(a) and \ref{heuristic_data}(a), and CIFAR-CNN in Figures \ref{heuristic_latency}(b) and \ref{heuristic_data}(b). This can be explained by the fact that LeNet and CIFAR-CNN are light weight networks and 2 devices are enough to handle an incoming request. Consequently, the per-layer approach assigns the tasks to a maximum of 2 participants. In our approach, at the lower privacy level, the security requirements are also met when involving only 2 contributors. When the security level is tight, more devices should participate in the classification to ensure the privacy of the sensitive data. Hence, CNN tasks (conv, ReLU, etc.) are parallelized, multiple channels are used for transmission, and latency is reduced compared to per-layer distribution that transmits the whole output data on one communication link. Moreover, we can notice that when the privacy level is higher, the latency is further minimized, as segments are highly distributed and tasks are more parallelized. Regarding the shared data, our approach presents a larger data transmission compared to per-layer distribution, for high privacy levels (SSIM = 0.6 or 0.4) and large networks (e.g. VGG in  Figure \ref{heuristic_data}(c)).  The same data sharing behavior can also be seen for heterogeneous requests in Figure \ref{heuristic_data}(d), where the communication between participant is much higher in per-segment distribution. This large data sharing is mainly related to the distribution of each convolutional task among multiple participants. In fact, this is not the case of the per-layer approach, which shares only the resultant reduced data after executing the task. However, this is acceptable as our system enhances both the privacy of sensitive data and the latency of the classification.  More specifically, for the higher privacy level, our system gains 30\%, 70\%, and 40\% in terms of classification latency for LeNet (Figure \ref{heuristic_latency}(a)), CIFAR (Figure \ref{heuristic_latency}(b)) and heterogeneous networks (Figure \ref{heuristic_latency}(c)) respectively, compared to per-layer distribution.  Figure \ref{heuristic_data} shows also that a higher level of privacy, results in a larger data sharing. This can be justified by the high distribution of segments incurring a larger dependency between devices. To summarize, a trade-off can be established, where sacrificing on privacy results in higher latency and less data sharing, which means less cost and energy consumption.

\begin{figure}[!h]
\centering
	\includegraphics[scale=0.5]{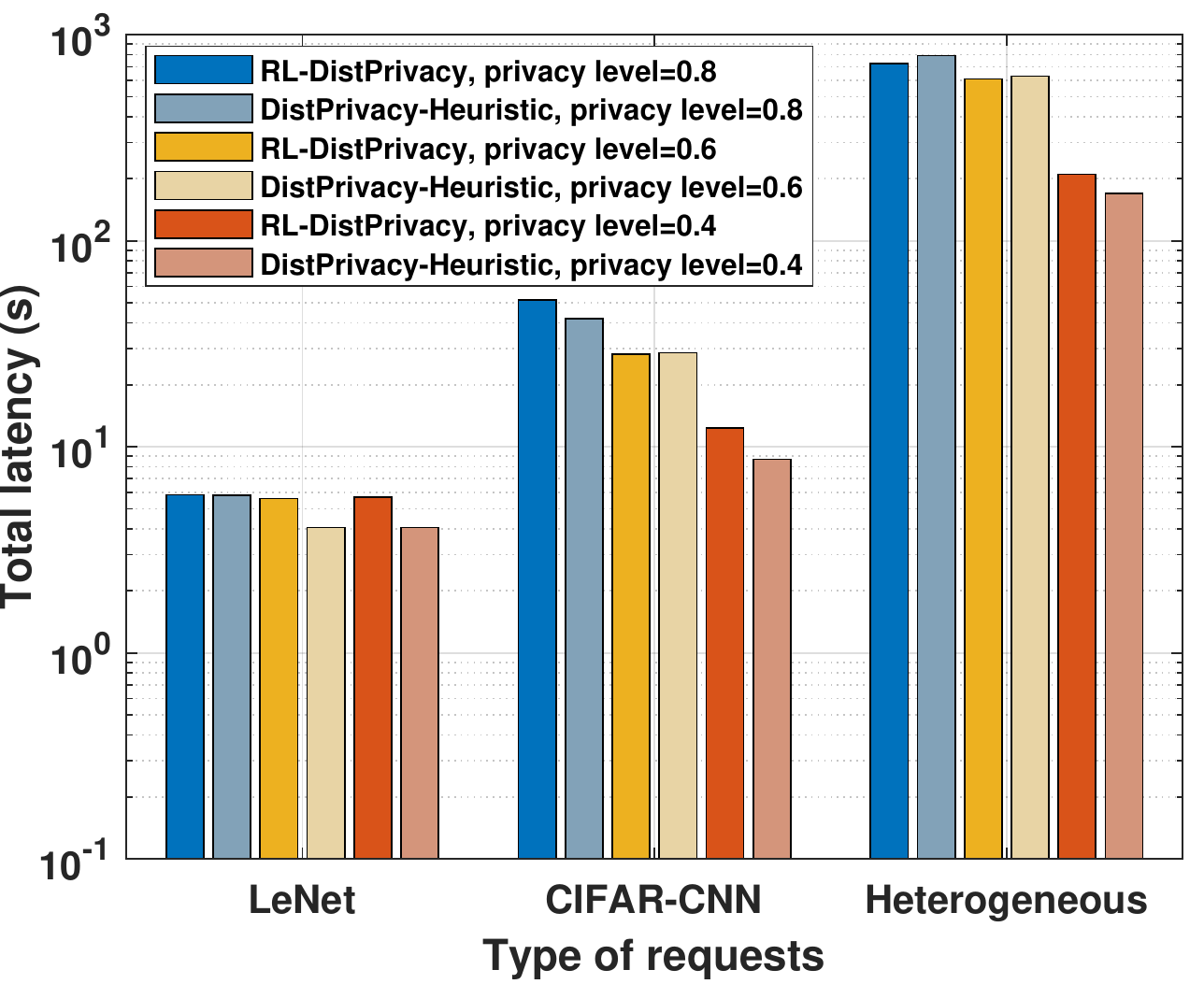}
	\caption{Total latency for different types of requests: RL-DistPrivacy vs. DistPrivacy-Heuristic.}
	\label{latency}
\end{figure}
\begin{figure}[!h]
\centering
	\includegraphics[scale=0.5]{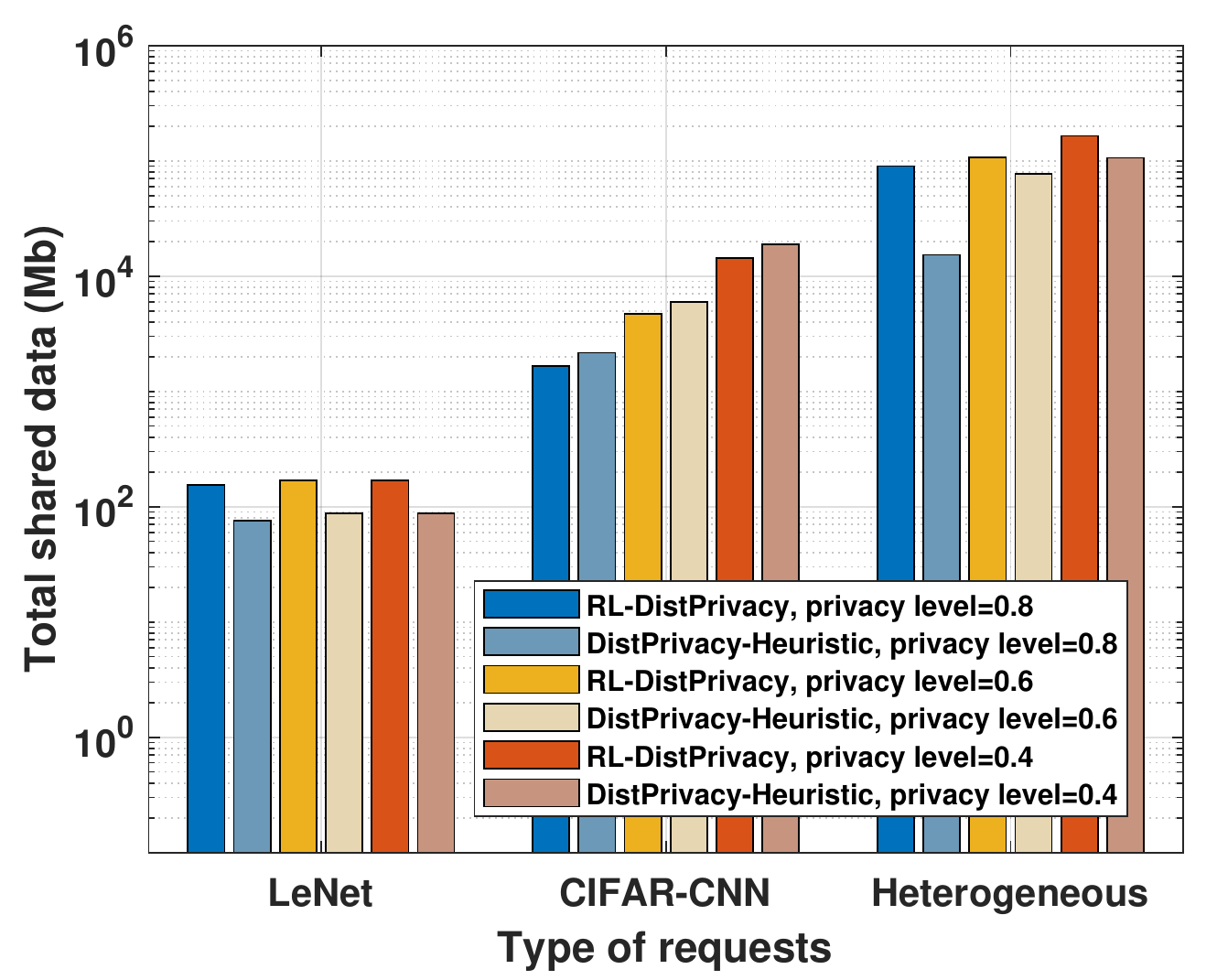}
	\caption{Total shared data for different types of requests: RL-DistPrivacy vs. DistPrivacy-Heuristic.}
	\label{shared_data}
\end{figure}

Figures \ref{latency} and  \ref{shared_data}  depict  the  total  latency  and  the  total shared data, when generating 250 requests to be classified by the same type of CNN or by heterogeneous CNNs. Furthermore, the figures show the performance of our RL-DistPrivacy approach compared to our heuristic based approach designed with the same network parameters and latency minimization objective \cite{Emna_globecom}. This heuristic is based on a greedy resource allocation strategy, where segments are assigned to devices with available resources and contributing to minimize the latency.  When the surveillance system uses only LeNet network (9 layers and 28x28 images) for image classification, the inference latency (Figure \ref{latency}) and the shared data (Figures \ref{shared_data}) are very low. This is justified by the shallow structure of LeNet that is composed of 8 feature maps and requires only 2 participants to guarantee the privacy of the data. When the classifier is CIFAR CNN, the inference latency is still small compared to the time needed to compute VGG requests, as the number of layers is also limited (17 layers), the size of captured images is small (32x32), and the number of intermediate feature maps does exceed 128. Finally, VGG16 and VGG19 are characterized by a deep network enabling them to present outstanding performance for image classification. This complex structure requiring high computation and memory load restrains them from being deployed in resource-constrained devices. Distributing the deep VGG networks into IoT devices allows to parallelize the computation of different tasks to achieve smaller decision-taking latency, in addition to reducing the memory occupation per participant. 

Our system presents a slightly less performance compared to heuristic based approach, when dealing with the LeNet and CIFAR. This can be explained by the fact that the aforementioned models do not need a strategy learning to assign the computational tasks and any greedy heuristic can manage to exploit the available resources to allocate the CNN tasks. Meanwhile, the RL-systems exhibit sometimes random tasks to detect any occurring changes in the network, which justifies the small difference between the two approaches (see Figure \ref{latency}). When the network is heterogeneous, the RL-DistPrivacy presents a better latency and lower rejections (28\% of rejected classifications compared to 40\% presented by the heuristic, when SSIM=0.4), owing to the ability of RL to approximate the near-optimal allocation policy and estimate the required transmission and computation resources for each layer, and manage the existing participants accordingly. 

\subsubsection{Impact of the network configuration}
\begin{figure}[!h]
\centering
	\includegraphics[scale=0.46]{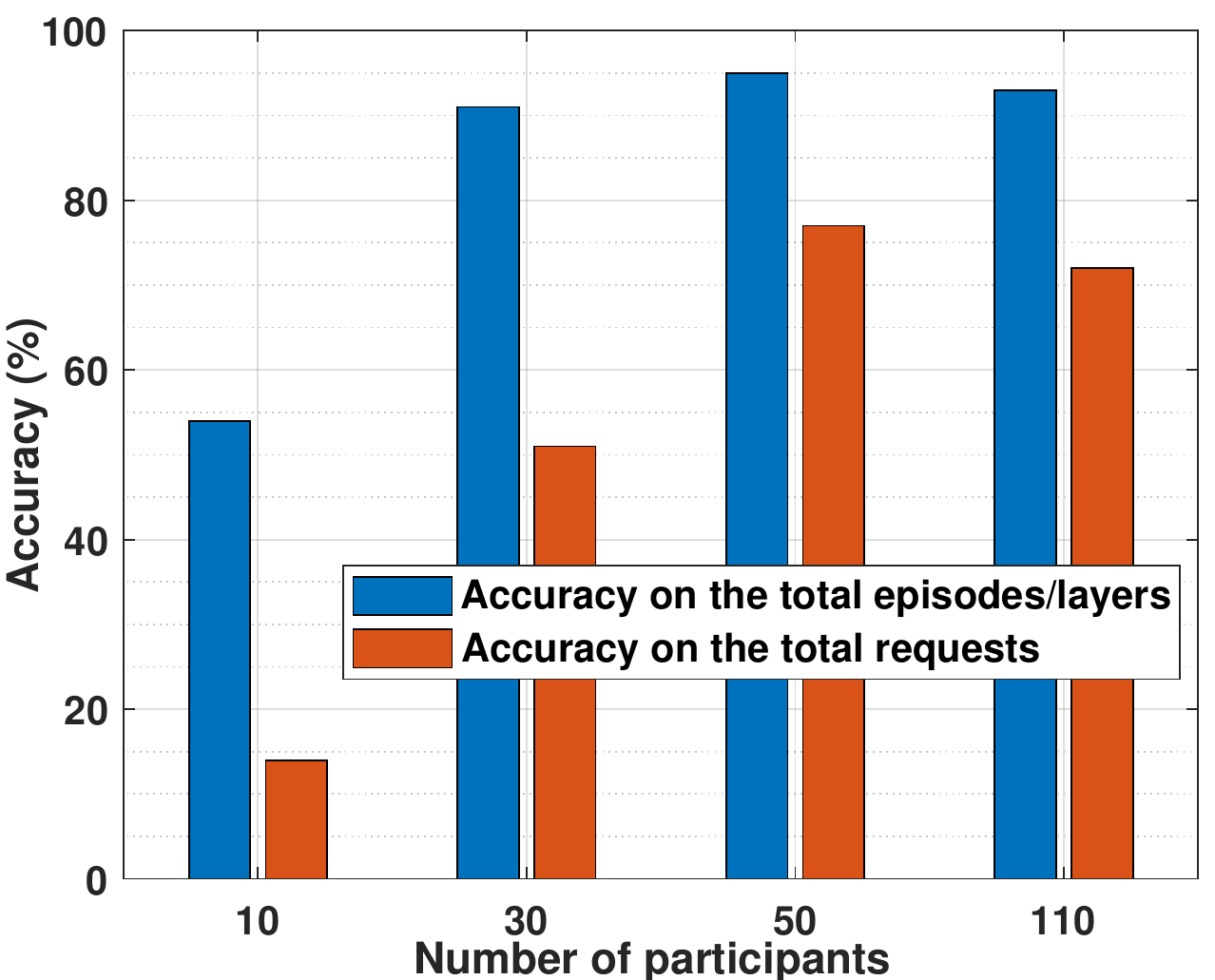}
	\caption{ Accuracy of the system in terms of meeting constraints with heterogeneous requests and while varying the number of devices (70\% RPi3 - 30\% LG Nexus).}
	\label{number}
\end{figure}
Figure \ref{number} shows the performance of a network computing heterogeneous requests, while ensuring a privacy level equal to 0.4 and varying the number of participants. We note that participants are composed of 70\% of RPi3 and 30\% of LG Nexus devices. We can see that 50 IoT devices are sufficient to satisfy the requirements of the surveillance system in terms of resources and privacy. A lower number of participants may not be enough to infer all the requests, particularly VGG classifications. We notice also that a higher number of devices does not apport any benefit to the performance of the system. Instead, increasing the number of devices means enlarging the actions space, which impacts the accuracy of the decisions as seen in Figure \ref{number}.
\begin{figure}[!h]
\centering
	\includegraphics[scale=0.46]{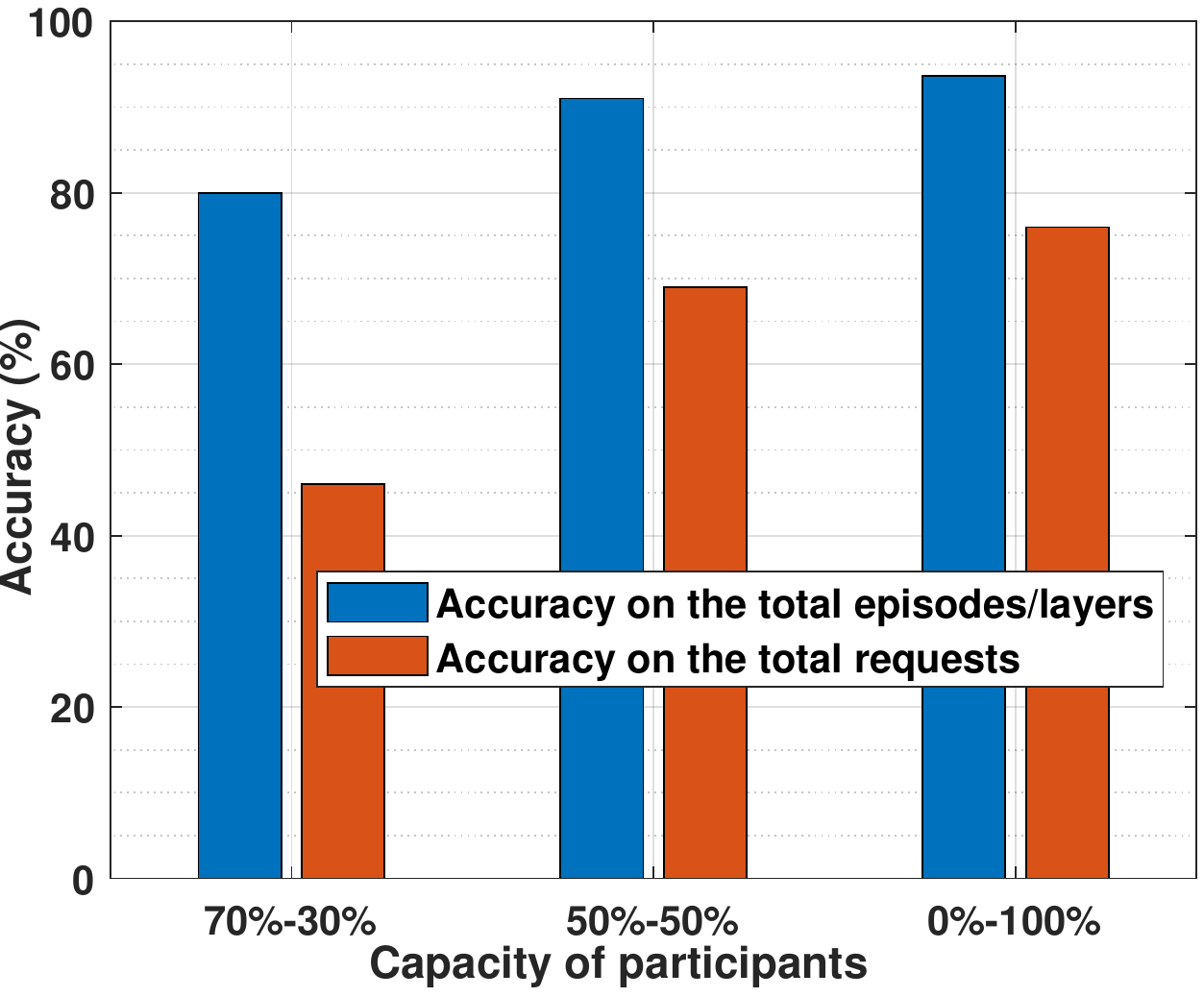}
	\caption{ Accuracy of the system in terms of meeting constraints with heterogeneous requests and 70 IoT devices (STM32H7 \% - LG Nexus \%).}
	\label{capacity}
\end{figure}

Figure \ref{capacity} illustrates the performance of the system computing heterogeneous requests, when varying the capacities of IoT devices. The simulated network is composed of 70 participants of high performance (LG Nexus with 2.28 GHz processor and 2 GB RAM) and low performance (STM32H7 with 400 MHz-cortex and 1 MB RAM) devices. We can conclude that a collaborative network composed of resources-limited devices is not adequate for distributed CNN, particularly complex networks such as VGG. However, when 50\% of devices or more are powerful, the system performs very well.
\subsubsection{Comparison to the optimal framework}
\begin{figure}[h]
	\hspace{-3mm}
	\mbox{
		\subfigure[\label{latency_optimal1}]{\includegraphics[scale=0.43]{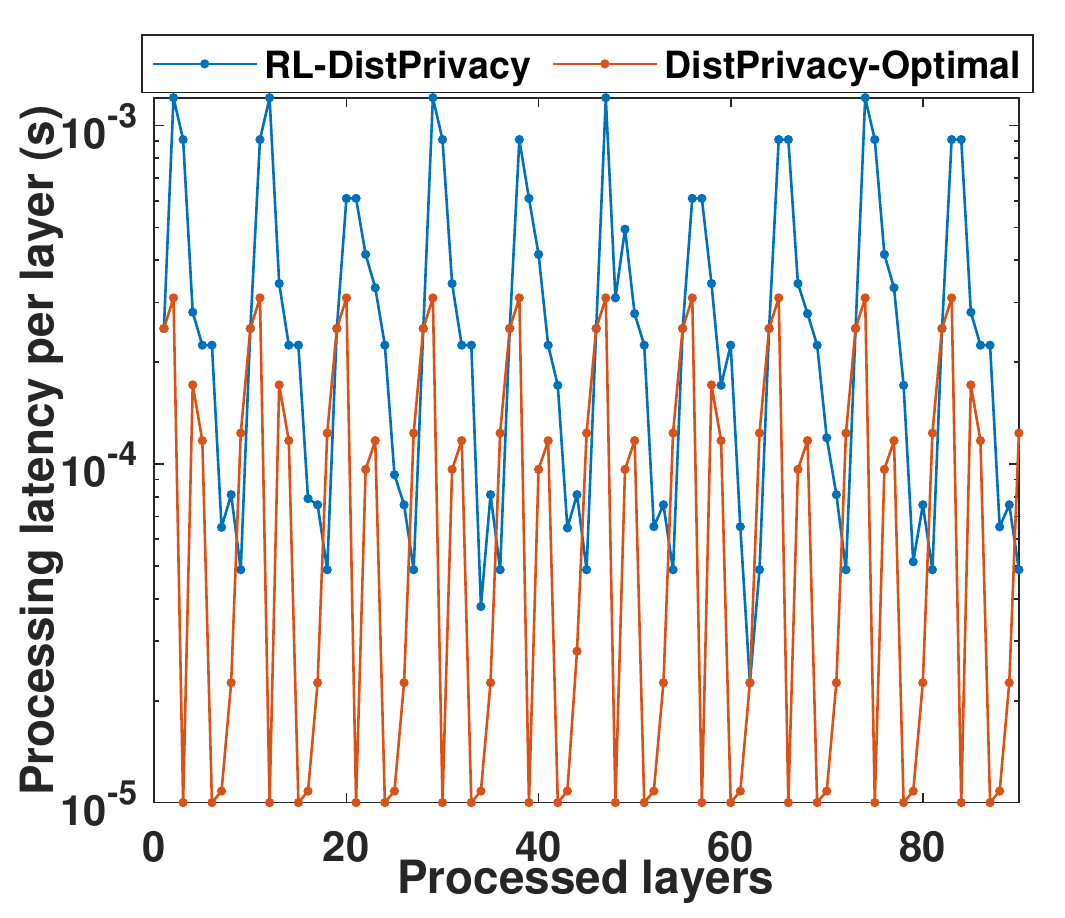}}
		\hspace{-3.7mm}
        \subfigure[\label{data_optimal1}]{\includegraphics[scale=0.43]{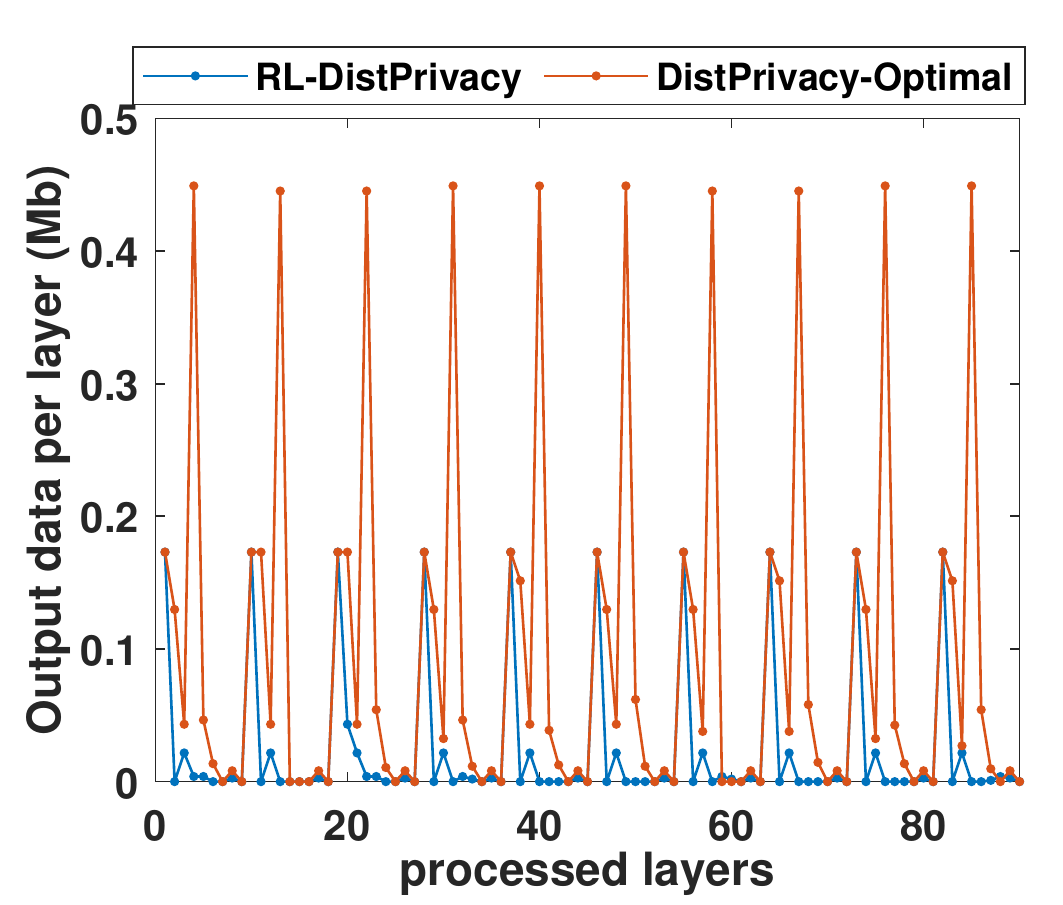}}
}\\

	\mbox{
	\hspace{-3mm}
		\subfigure[\label{latency_optimal2}]{\includegraphics[scale=0.43]{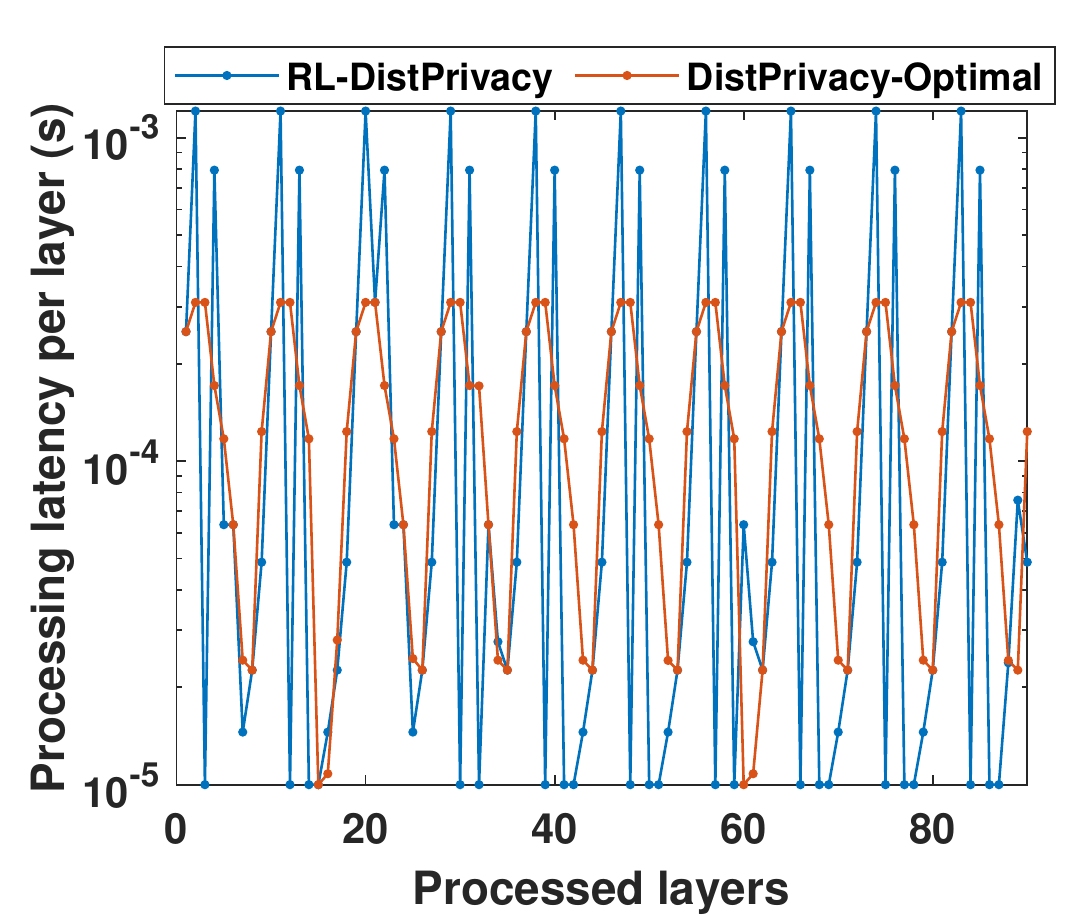}}
		\hspace{-3.7mm}
        \subfigure[\label{data_optimal2}]{\includegraphics[scale=0.43]{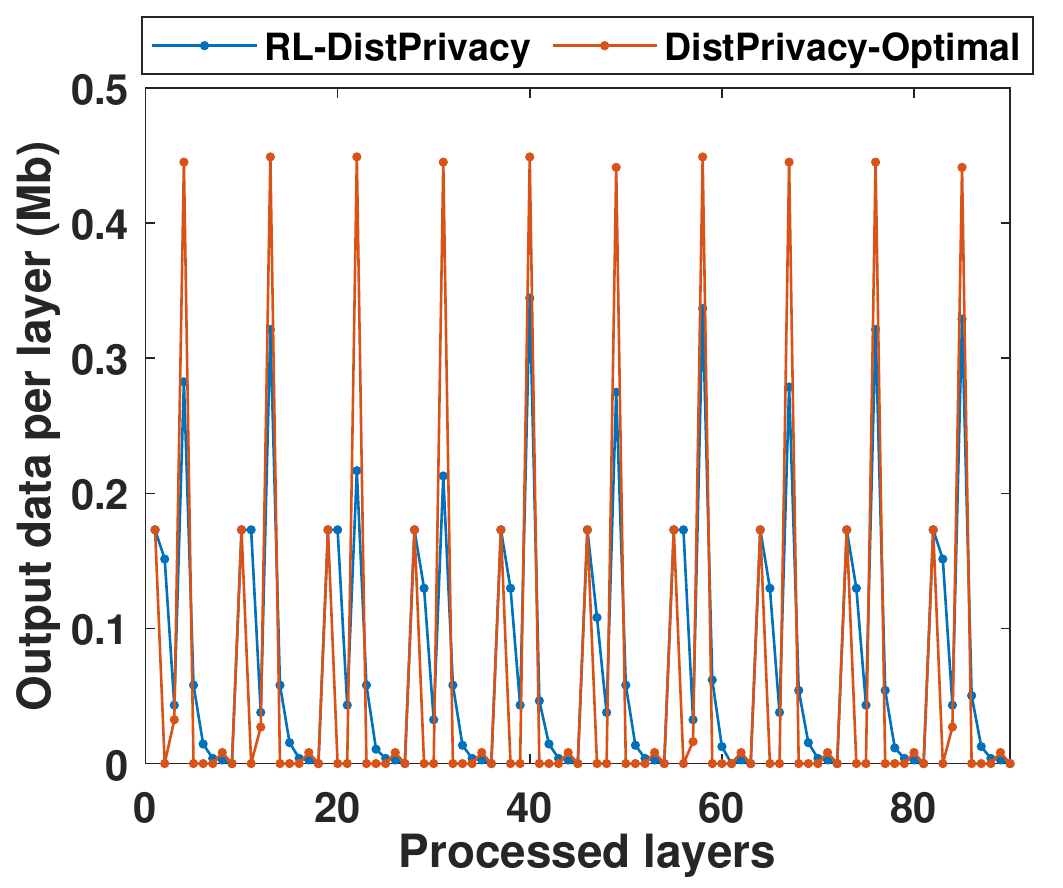}}
}
	\caption{RL-DistPrivacy vs. DistPrivacy-Optimal: (a) and (b): results of LeNet classification requests with SSIM=0.8; (c) and (d): results of LeNet classification requests with SSIM=0.6.}
	\label{optimal}
\end{figure}

\begin{figure}[!h]
\centering
	\includegraphics[scale=0.57]{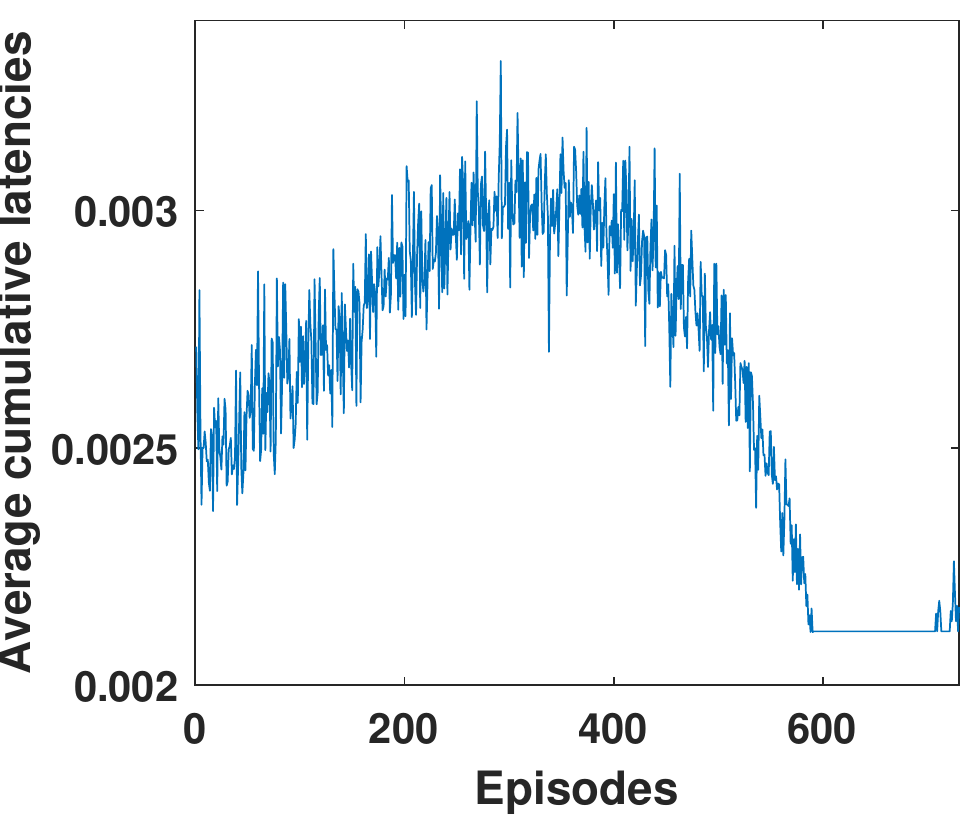}
	\caption{Average cumulative latencies vs. training rewards.}
	\label{optimal_conv}
\end{figure}

In Figure \ref{optimal}, we compare the performance of our RL-DistPrivacy system to the optimal results. Due to the combinatorial complexity of the problem, the optimal results are conducted on 10 LeNet classification requests (90 layers and 750 segments) and 10 IoT participants, while respecting two SSIM levels, namely 0.8 and 0.6. One optimal LeNet request allocation is executed in 15 minutes, while 3 and 10 requests cost the system, 1 and 5 hours of computation respectively. The design of our online solution to handle real-time CNN distribution and allocation is justified by the time complexity to solve the optimization. We remind that we run our simulation on a core i7 computer with 16 GB RAM. In Figure \ref{latency_optimal1}, we can see that the optimal solution achieves a much better latency compared to our RL approach, when the required privacy level is equal to 0.8. In fact, the objective of our optimization is to minimize the latency while respecting different constraints. As we explained previously, when more tasks are paralellized, the latency is further reduced. Thus, the optimization opts for including the maximum number of participants to synchronize tasks execution. However, this strategy is not beneficial for data sharing, as  depicted in Figure \ref{data_optimal1}. Indeed, the low latency accomplished by the optimization is accompanied by a higher data transmission compared to the online RL-approach. The optimal strategy will not be the same, when dealing with VGG models having critical requirements in terms of transmission resources. In this context, we encouraged our RL system to use the minimum number of devices by being rewarded following the equation in (\ref{reward3}). Our reward function impacted the shape of the latency convergence. In fact, the first phase of the RL learning is the random acting process, where devices are chosen randomly, leading to involving a big number of participants. This process results in starting by a sub-optimal latency as depicted in Figure \ref{optimal_conv}. Then, the system starts to minimize the number of participants, until reaching again a reduced latency. Figures \ref{latency_optimal2} and \ref{data_optimal2} illustrate the performance of the DistPrivacy, when SSIM is equal to 0.6. We can see that both online and optimal solutions become closer, which is caused by the privacy requirements that impose to distribute the segments on a higher number of devices.
\section{Conclusion}\label{conclusion}
In this work, we examined the vulnerability of  CNN distribution to black-box attacks and the possibility of recovering sensitive intermediate data, under different scenarios. We proved that distributing the output of layers into multiple IoT participants and exposing only few feature maps can strengthen the privacy of the original data. Hence, we re-designed the deep learning distribution solutions to reduce black-box risks and match capacities constraints characterizing the IoT devices. This approach has been formulated as an optimization problem, where inference latency is minimized. Next, because of the problem complexity, we proposed a reinforcement learning solution, namely RL-DistPrivacy. This solution is characterized by its online decisions and its satisfactory performance compared to the optimal results, in addition to its  capacity to manage the dynamic of the system through continuous learning. Our extensive simulations illustrated the performance of our approach compared to a recent per-layer distribution strategy and a heuristic-based approach. Future works will encompass the management of failures or re-transmissions in the communication. Moreover,
the methodology could be extended to deal with new participants, such as edge and fog servers.  Another possible direction is to conduct an extensive study to generalize our defense strategy and prove its ability to mitigate different types of attacks, independently from the ability of the attacker.
\section*{Acknowledgement}
This publication was supported by Qatar Foundation and partially made possible by NPRP-Standard (NPRP-S) Thirteen (13th) Cycle grant \# NPRP13S-0205-200265 from the Qatar National Research Fund (a member of Qatar Foundation). The findings herein reflect the work, and are solely the responsibility, of the authors.

\vspace{-2cm}
\begin{IEEEbiography}[{\includegraphics[width=1in,height=1.3in,clip]{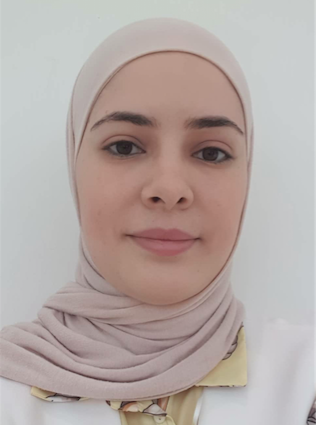}}]{Emna Baccour}
received the Ph.D. degree in computer Science from the University of Burgundy, France, in 2017. She was a postdoctoral fellow at Qatar University on a project covering the interconnection networks for massive data centers and then on a project covering video caching and processing in mobile edge computing networks. She currently holds a postdoctoral position at Hamad Ben Khalifa University. Her research interests include data center networks, cloud computing, green computing and software defined networks as well as distributed systems. She is also interested in edge networks and mobile edge caching and computing.
\end{IEEEbiography}
\vskip -1\baselineskip plus -1fil
\begin{IEEEbiography}[{\includegraphics[width=1.4in,height=1.4in,clip,keepaspectratio]{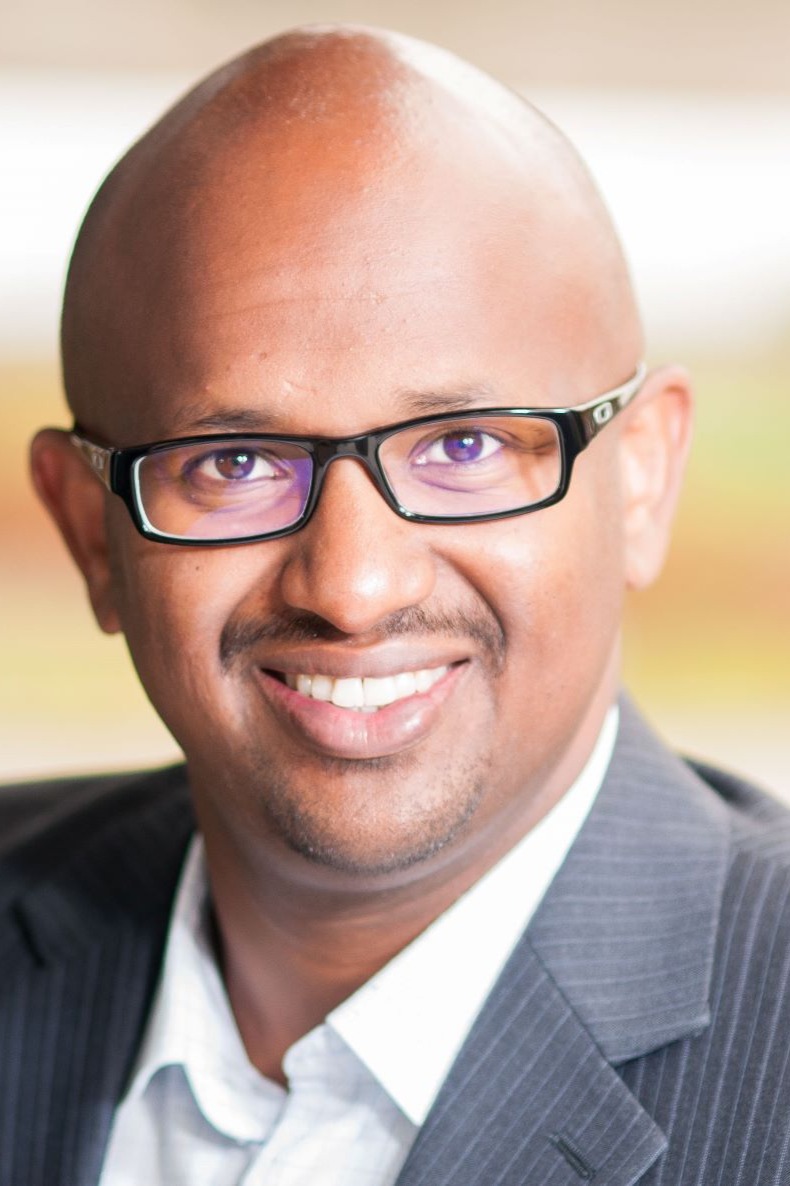}}]{Aiman Erbad}
is an Associate Professor and Head of Information and Computing Technology Division in the College of Science and Engineering, Hamad Bin Khalifa University (HBKU). Prior to this, he was an Associate Professor at the Computer Science and Engineering (CSE) Department and the Director of Research Planning and Development at Qatar University until May 2020. He also served as the Director of Research Support responsible for all grants and contracts (2016-2018) and as the Computer Engineering Program Coordinator (2014-2016). Dr. Erbad obtained a Ph.D. in Computer Science from the University of British Columbia (Canada) in 2012, a Master of Computer Science in embedded systems and robotics from the University of Essex (UK) in 2005, and a BSc in Computer Engineering from the University of Washington, Seattle in 2004. He received the Platinum award from H.H. The Emir Sheikh Tamim bin Hamad Al Thani at the Education Excellence Day 2013 (Ph.D. category). He also received the 2020 Best Research Paper Award from Computer Communications, the IWCMC 2019 Best Paper Award, and the IEEE CCWC 2017 Best Paper Award. His research interests span cloud computing, edge intelligence, Internet of Things (IoT), private and secure networks, and multimedia systems. He is a senior member of IEEE and ACM.
\end{IEEEbiography}
\vskip -1\baselineskip plus -1fil
\begin{IEEEbiography}[{\includegraphics[width=1.05in,height=1.3in,clip]{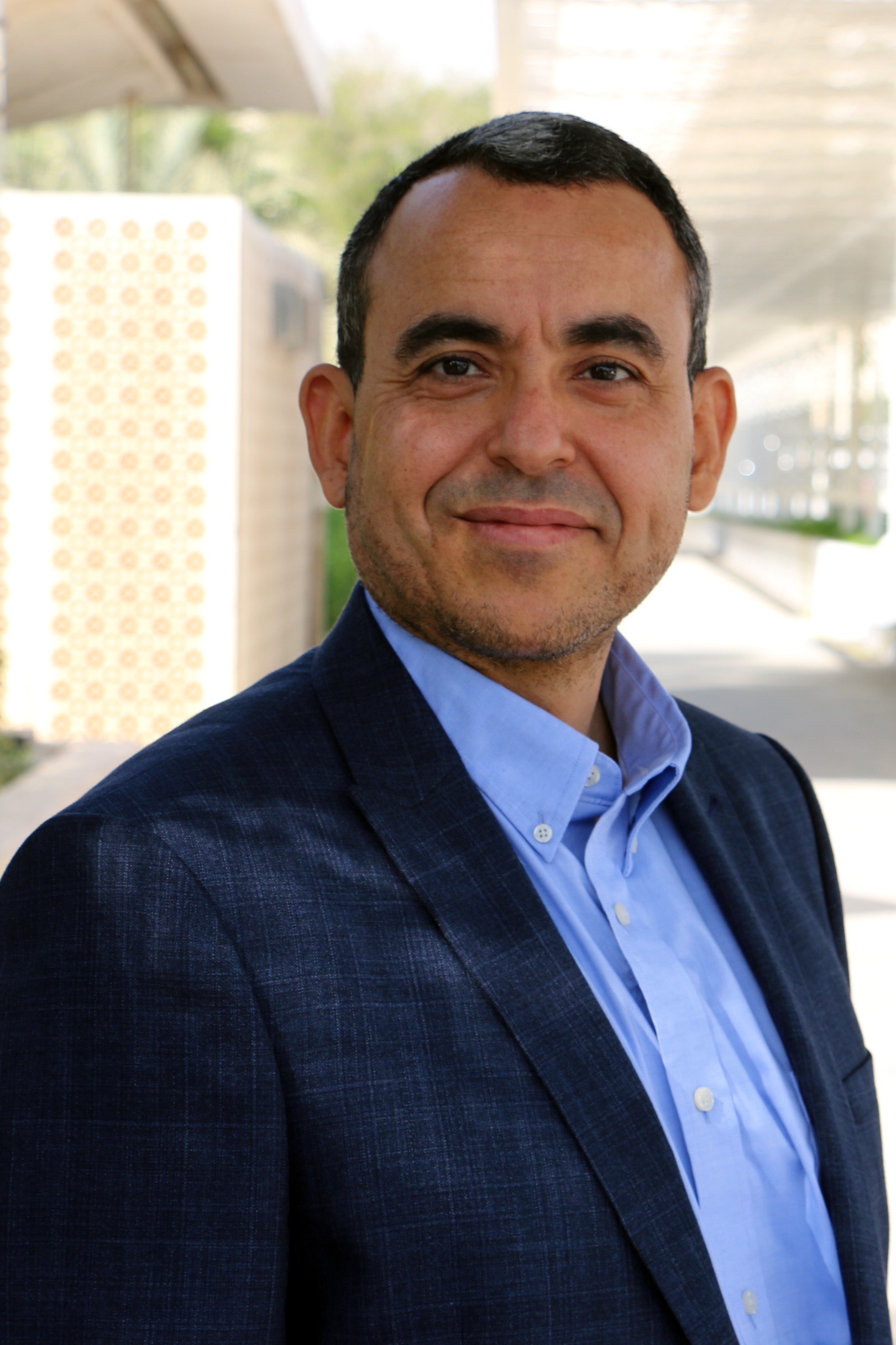}}]{Amr Mohammed}
(S’ 00, M’ 06, SM’ 14) received his M.S. and Ph.D. in electrical and computer engineering from the University of British Columbia, Vancouver, Canada, in 2001, and 2006 respectively. He has worked as an advisory IT specialist in IBM Innovation Centre in Vancouver from 1998 to 2007, taking a leadership role in systems development for vertical industries.  He is currently a professor in the college of engineering at Qatar University and the director of the Cisco Regional Academy. He has over 25 years of experience in wireless networking research and industrial systems development. He holds 3 awards from IBM Canada for his achievements and leadership, and 4 best paper awards from IEEE conferences. His research interests include wireless networking, and edge computing for IoT applications. Dr. Amr Mohamed has authored or co-authored over 160 refereed journal and conference papers, textbook, and book chapters in reputable international journals, and conferences. He is serving as a technical editor for the journal of internet technology and the international journal of sensor networks. He has served as a technical program committee (TPC) co-chair for workshops in IEEE WCNC’16. He has served as a co-chair for technical symposia of international conferences, including Globecom’16, Crowncom’15, AICCSA’14, IEEE WLN’11, and IEEE ICT’10. He has served on the organization committee of many other international conferences as a TPC member, including the IEEE ICC, GLOBECOM, WCNC, LCN and PIMRC, and a technical reviewer for many international IEEE, ACM, Elsevier, Springer, and Wiley journals.
\end{IEEEbiography}
\vskip -1\baselineskip plus -1fil
\begin{IEEEbiography}[{\includegraphics[width=1in,height=1.3in,clip]{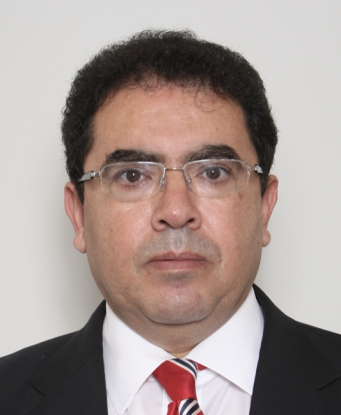}}]{Mounir Hamdi}
received the B.S. degree (Hons.) in electrical engineering (computer engineering) from the University of Louisiana, in 1985, and the M.S. and Ph.D. degrees in electrical engineering from the University of Pittsburgh, in 1987 and 1991, respectively. He was a Chair Professor and a Founding Member of The Hong Kong University of Science and Technology (HKUST), where he was the Head of the Department of Computer Science and Engineering. From 1999 to 2000, he held visiting professor positions at Stanford University and the Swiss Federal Institute of Technology. He is currently the Founding Dean of the College of Science and Engineering, Hamad Bin Khalifa University (HBKU). His area of research is in high-speed wired/wireless networking, in which he has published more than 360 publications, graduated more 50 M.S./Ph.D. students, and awarded numerous research grants. In addition, he has frequently consulted for companies and governmental organizations in the USA, Europe, and Asia. He is a Fellow of the IEEE for his contributions to design and analysis of high-speed packet switching, which is the highest research distinction bestowed by IEEE. He is also a frequent keynote speaker in international conferences and forums. He is/was on the editorial board of more than ten prestigious journals and magazines. He has chaired more than 20 international conferences and workshops. In addition to his commitment to research and academic/professional service, he is also a dedicated teacher and a quality assurance educator. He received the Best 10 Lecturer Award and the Distinguished Engineering Teaching Appreciation Award from HKUST. He is frequently involved in higher education quality assurance activities as well as engineering programs accreditation all over the world.
\end{IEEEbiography}
\vskip -1\baselineskip plus -1fil
\begin{IEEEbiography}[{\includegraphics[width=1in,height=1.3in,clip]{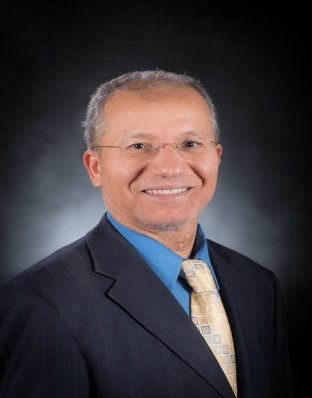}}]{Mohsen Guizani}
(M’89–SM’99–F’09) received the BS (with distinction), MS and PhD degrees in Electrical and Computer engineering from Syracuse University, Syracuse, NY, USA in 1985, 1987 and 1990, respectively. He is currently a Professor of Machine Learning and the Associate Provost at Mohamed Bin Zayed University of Artificial Intelligence (MBZUAI), Abu Dhabi, UAE. Previously, he worked in different institutions in the USA. His research interests include applied machine learning and artificial intelligence, Internet of Things (IoT), intelligent autonomous systems, smart city, and cybersecurity. He was elevated to the IEEE Fellow in 2009 and was listed as a Clarivate Analytics Highly Cited Researcher in Computer Science in 2019, 2020 and 2021. Dr. Guizani has won several research awards including the “2015 IEEE Communications Society Best Survey Paper Award”, the Best ComSoc Journal Paper Award in 2021 as well five Best Paper Awards from ICC and Globecom Conferences. He is the author of ten books and more than 800 publications. He is also the recipient of the 2017 IEEE Communications Society Wireless Technical Committee (WTC) Recognition Award, the 2018 AdHoc Technical Committee Recognition Award, and the 2019 IEEE Communications and Information Security Technical Recognition (CISTC) Award. He served as the Editor-in-Chief of IEEE Network and is currently serving on the Editorial Boards of many IEEE Transactions and Magazines. He was the Chair of the IEEE Communications Society Wireless Technical Committee and the Chair of the TAOS Technical Committee. He served as the IEEE Computer Society Distinguished Speaker and is currently the IEEE ComSoc Distinguished Lecturer. 
\end{IEEEbiography}
\end{document}